\documentclass[twoside,11pt]{article}
\usepackage{jair,apacite,rawfonts}

\usepackage[dvipsnames]{xcolor}

\usepackage{amsmath}
\usepackage{amsfonts}
\usepackage{numprint}

\usepackage{graphicx}
\usepackage{tikz}

\usepackage{mathdots}
\usepackage{cancel}

\usepackage{array}
\usepackage{multirow}
\usepackage{amssymb}
\usepackage{gensymb}
\usepackage{tabularx}
\usepackage{booktabs}
\usepackage{caption}
\usepackage{subcaption}
\usepackage{makecell}

\usepackage{algorithm}
\usepackage{algpseudocode}

\usepackage{physics}
\usepackage{amsmath}
\usepackage{yhmath}
\usepackage{siunitx}
\usepackage{extarrows}
\usepackage{csquotes}

\usetikzlibrary{fadings}
\usetikzlibrary{patterns}
\usetikzlibrary{shadows.blur}
\usetikzlibrary{shapes}

\usepackage{mathtools}

\mathtoolsset{showonlyrefs,showmanualtags}

\DeclareMathOperator*{\argmax}{argmax}
\DeclareMathOperator*{\mean}{average}

\jairheading{80}{2024}{273-326}{07/2023}{06/2024}
\ShortHeadings{On the Trade-off between Redundancy and Cohesion in Extractive Summarization}
{Cardenas, Galle, \& Cohen}
\firstpageno{273}

\begin{document}

%
%
\title{On the Trade-off between Redundancy and Cohesion in Extractive Summarization}

\author{\name Ronald Cardenas \email ronald.cardenas@ed.ac.uk \\
       \addr Institute for Language, Cognition and Computation \\
       School of Informatics, University of Edinburgh \\
       10 Crichton Street, Edinburgh, UK
       \AND
       \name Matthias Gall\'e \email matthias@cohere.com \\
       \addr Cohere \\
       51 Great Marlborough St, London, UK
       \AND
       \name Shay B. Cohen \email scohen@inf.ed.ac.uk \\
       \addr Institute for Language, Cognition and Computation \\
       School of Informatics, University of Edinburgh \\
       10 Crichton Street, Edinburgh, UK
       }

\maketitle

\begin{abstract}

Extractive summaries are usually presented as lists of sentences with no expected cohesion between them
and with plenty of redundant information if not accounted for.
In this paper, we investigate the trade-offs incurred when aiming to control for inter-sentential cohesion and redundancy in extracted summaries,
and their impact on their informativeness.
As case study, we focus on the summarization of long, highly redundant documents and consider 
two optimization scenarios, reward-guided and with no supervision.
In the reward-guided scenario, we compare systems that control for redundancy and cohesion during sentence scoring.
In the unsupervised scenario, we introduce two systems that aim to control all three properties --informativeness, redundancy, and cohesion-- in a principled way.
Both systems implement a psycholinguistic theory that simulates how humans keep track of relevant content units 
and how cohesion and non-redundancy constraints are applied in short-term memory during reading.
Extensive automatic and human evaluations reveal that systems optimizing for --among other properties-- cohesion are 
capable of better organizing content in summaries compared to systems that optimize only for redundancy, while maintaining comparable informativeness.
We find that the proposed unsupervised systems manage to extract highly cohesive summaries across varying levels of document redundancy, although
sacrificing informativeness in the process.
Finally, we lay evidence as to how simulated cognitive processes impact the trade-off between the analyzed summary properties.

\end{abstract}


\section{Introduction}

Automatic single-document summarization is the task of reading a text document and presenting an end-user (be it a human user or a module down a processing pipeline)
with a shorter text, the \textit{summary}, that retains the gist of the information consumed in the document.
Such a complex task can be divided into the following three general steps: 
(i) discretization of the information in the source document into semantic content units and building a 
representation of these units,
(ii) selection of content units such that they are relevant with respect to the source document,  non-redundant among themselves, and informative to the end-user; and finally,
(iii) production of a summary text that is coherent and cohesive.
From the many variations of the summarization task investigated in recent years \shortcite{litvak2017query,shapira2017interactive,narayan2019article,xiao2019extractive,amplayo2021aspect},
most extractive summarization approaches choose sentences as the indivisible content unit,
assign a numerical score to each sentence, select a subset of them, and finally concatenate them into a single text to be presented as the summary.

Even though recent advances in machine learning brought promising results --mostly involving increasingly larger neural networks-- in all stages of the summarization pipeline,
core challenges such as redundancy \shortcite{xiao2020systematically,jia2021deep,gu2022memsum}
remain critically open.
Notably, \citeA{xiao2020systematically} reported that modern extractive summarization systems are prone to produce highly redundant excerpts when redundancy is not explicitly accounted for.
The problem becomes particularly acute when the source document is highly redundant, i.e.\ information is repeated in many parts of the document.
Some examples of highly redundant documents include scientific articles, and in general, long-structured documents.
Consider the example in Figure~\ref{fig:doc-red} showcasing how information is repeated across sections in a scientific article.
Information redundancy is characteristic of the writing style in scientific literature: 
the `Introduction' section is expected to lay down the research questions addressed in the paper, each of which will be elaborated upon in the following sections,
and the `Conclusion' section (or equivalent) gathers insights and summarizes the answers to each research question.

\begin{figure}[t]
\centering
\scriptsize
\begin{tabular}{p{14cm}}
\toprule
\textbf{Introduction}   \\
\underline{Wolf Rayet (WR)} stars are evolved, massive stars that are losing their mass rapidly through strong \underline{stellar winds} (Conti, 1976).     \\
In this scenario, hot, massive OB stars are considered to be the \underline{WR} precursors that lose their external layers via \underline{stellar winds}, leaving exposed their He-burning nuclei and H-rich surfaces …    \\
\textcolor{orange}{[At radio frequencies, the excess of emission is associated with the contribution of the free thermal emission coming from the ionized and expanding envelope formed by the stellar wind]$_	\circ$} …      \\

In this paper, we present \textcolor{blue}{[simultaneous, multi-frequency observations of a sample of 13 WR stars using the VLA at $4.8$, $8.4$, and $22.5$ GHz]$_\diamond$}, aimed at \textcolor{teal}{[disentangling the origin of their stellar wind radio emission through the analysis of their spectral index and time variability by comparison with previous observations.]$_\bigtriangleup$} \\ \midrule
\textbf{Observations}                                    \\
We performed \textcolor{blue}{[radio observations of a sample of 13 WR stars]$_\diamond$}, listed in Table 1, \textcolor{blue}{[with the Very Large Array ( VLA )]$_\diamond$} of the National Radio Astronomy Observatory (NRAO) …                              \\ \midrule
\textbf{Results} \\
We observed a total of \textcolor{blue}{[13 WR stars]$_\diamond$} and \textcolor{red}{[detected 12 of them at least at one frequency]$_\bullet$} …  \\
Summarizing, \textcolor{purple}{[we have found four T (...) , one NT (...) , and seven T/NT sources (...)]$_\bigtriangledown$} …  \\
as we mentioned in Section 1, \textcolor{orange}{[it is possible to estimate the free radiation emitted from ionized extended envelopes]$_	\circ$} …    \\ \midrule
\textbf{Discussion}      \\
\textcolor{teal}{[The results of our observations presented in Section 3 provide relevant information about the nature of the radio emission of the 12 detected WR stars]$_\bigtriangleup$} .      \\
\textcolor{purple}{[The detected flux densities and spectral indices displayed by the sources of our sample indicate the existence of thermal, non-thermal dominant, and composite spectrum sources]$_\bigtriangledown$} …     \\ \midrule
\textbf{Conclusions}                                                                                               \\
We have presented \textcolor{blue}{[simultaneous, multi-frequency observations of 13 WR stars at $4.8$, $8.4$, and $23$ GHz.]$_\diamond$}  \\
We have \textcolor{red}{[detected 12 of the observed sources at least at one frequency]$_\bullet$} …     \\
\textcolor{teal}{[From the observed flux densities, spectral index determinations, and the comparison of our results with previous ones, we have disentangled the nature of the emission in these WR stars]$_\bigtriangleup$} …    \\ \bottomrule
\end{tabular}
\caption{Sections of a scientific article taken from the \textsc{arXiv} dataset showcasing 
information redundancy and cohesion.
Repeated content is marked by text chunks with the same color and symbol, whilst consecutive sentences present 
cohesive phrases underlined. }
\label{fig:doc-red}
\end{figure}

Another open challenge in summarization --and in open text generation in general-- is the production of coherent text \shortcite{sharma2019bigpatent,hua2021dyploc,steen-markert-2022-find,goyal-etal-2022-snac}.
In particular, local coherence --the property by which a text connects semantically similar content units between neighbouring sentences-- has proven challenging to capture computationally \shortcite{moon2019unified,jeon2020centering,jeon2022entity} and to incorporate into the summarization task 
without sacrificing performance in other aspects such as informativeness \cite{wu2018learning,xu2020discourse}.
When the connection between adjacent sentences is not explicitly clued by linguistic units, 
humans resort to \textit{inference}, the cognitive process by which prior knowledge is incorporated in order to force a connection and make sense of a text.
A special case of local coherence, \textit{cohesion}, makes the connection between adjacent sentences explicit by means of cohesive ties \cite{halliday1976cohesion} such as word repetitions,
pronouns, anaphoric expressions, and conjunctions \cite{garrod1977interpreting}.
Psycholinguistic research has found that cohesion improves text comprehension --the building of a mental representation of content-- especially when the subjects' background knowledge is insufficient to perform
inference successfully \cite{kintsch1990macroprocesses,garrod1994resolving}.
Critically, when human subjects were asked to read a document and write a summary immediately after, higher cognitive demand during comprehension was found to severely impact the cohesion and redundancy in the produced summaries \shortcite{lehto1996working,kintsch1998comprehension,ushiro2013measures,spirgel2016does}.

In this work, we investigate the trade-offs automatic summarization systems incur when aiming to control for redundancy and cohesion in produced summaries, and the impact on their informativeness.
We focus on control strategies performed during sentence scoring, resorting to greedy selection of the top-scoring sentences
until a predefined budget is met.
We study the case of long, highly redundant documents from complex knowledge domains --scientific articles collected from \textsc{arXiv} and \textsc{PubMed} \shortcite{cohan2018discourse}.
Two optimization scenarios are investigated, (i) when a specific summary property is optimized for under a reinforcement learning (RL) setup,
and (ii) when the summary property is modeled through proxies in an unsupervised setup.
In the RL setup, we compare systems that aim to balance informativeness and redundancy, against those which balance informativeness and cohesion.
We model this trade-off as a linear combination of property-specific rewards,
e.g.\ by combining a reward that encourages high ROUGE scores with 
a reward that encourages high local coherence.

In the unsupervised setup, we introduce two novel models that aim to control all three properties --informativeness, redundancy, and lexical cohesion.
These models implement the Micro-Macro Structure theory of text comprehension \cite{kintsch1978toward}, henceforth called KvD,
which provides a principled way of discretizing content into semantic units and organizing them in short and long-term memory.
Reading is performed one sentence at a time in \textit{memory cycles}, applying constraints 
to a representation of working memory --a type of short-term memory-- that explicitly model 
relevancy, non-redundancy, and cohesion among content units.
In each memory cycle, relevancy is modeled by pruning working memory down to a fixed number of content units,
keeping only the most relevant units read so far; cohesion, by ensuring lexical overlap between units in memory;
and non-redundancy, by discarding redundant units from memory.
Note that these models do not employ any reward signal and instead are completely unsupervised.

In the reward-guided scenario, extensive automatic --both quantitative and qualitative-- evaluation revealed that systems optimizing for cohesion 
are better at organizing content in the produced summaries, compared to systems only optimizing for informativeness or redundancy.
Moreover, cohesion-optimized models are able to obtain comparable --if not better-- informativeness and coverage levels.
In the unsupervised scenario, we found that simulated KvD reading is effective at
balancing cohesion and redundancy during sentence scoring, however at the expense of reduced informativeness.
Most notably, the proposed KvD systems manage to extract highly cohesive summaries across increasing levels of document redundancy.
We corroborated our findings with two human evaluation campaigns comparing our KvD systems against 
a strong unsupervised baseline that optimizes for cohesion.
In the first study, we found that participants find KvD summaries more informative,
indicating the effectiveness of constraining working memory to keep only 
the most relevant units, compared to modeling relevancy through sentence centrality as done by the analyzed baseline.
In the second study, we found that explicitly enforcing lexical cohesive links during reading allows 
the proposed KvD systems to extract summaries that exhibit a smooth topic transition between
adjacent or near-adjacent sentences,
with cohesive links connecting most sentences in the extracted summary.
Finally, we lay extensive evidence as to how the simulated cognitive processes impact the trade-off between informativeness, redundancy, and lexical cohesion in final summaries.\footnote{Code available at \url{\textbf{https://github.com/ronaldahmed/trade-off-kvd/}}}

The rest of the paper is organized as follows.
An overview of previous related work is presented in \S~\ref{sec-related},
followed by the problem formulation of the reward-guided control scenario in \S~\ref{section:rl-models}.
Then, \S~\ref{section:unsup-scenario} elaborates on control strategies in the unsupervised scenario,
providing a detailed description of the KvD theory(\S~\ref{section:kvd-model}) and the proposed systems (\S~\ref{section:kvd-sys}).
Lastly, \S~\ref{section:exp-setup} and \S\ref{section:res} describe our experimental setup and discuss our results, respectively.

\section{Related Work}
\label{sec-related}
In this section we discuss previous efforts related to automatic summarization, both traditional and modern (neural based), 
how the problems of redundancy and cohesion are being tackled,
and how cognitive science has influenced automatic summarization.

\subsection{Summarization Approaches}
Early approaches represented and organized content in a document using semantic and discourse methods such as lexical chains \shortcite{barzilay-elhadad-1997-using,silber2002efficiently},
latent semantic analysis \shortcite{gong2001generic,hachey2006dimensionality},
coreference information \shortcite{baldwin1998dynamic,steinberger2007two},
and rhetorical structure theory \shortcite{ono1994abstract,marcu1998build}.
In particular, graph representations proved effective in encoding relations between content units such as discourse relations \shortcite{wolf2004paragraph,louis2010discourse} and word co-occurrence statistics \shortcite{mihalcea2004textrank,erkan2004lexrank}.
After obtaining a representation of a document, the selection of content units (usually sentences) is posed as a unit ranking problem or a sequence labeling problem in which each unit is labeled as `select' or `not select'.
For this selection stage, machine learning approaches have proven effective at identifying summary-worthy units (i.e.\ relevant and informative) 
by leveraging manually-crafted features such as word frequency \shortcite{vanderwende2007beyond,nenkova2006compositional}, sentence length \shortcite{radev-etal-2004-mead}, and the presence of keywords of proper nouns \shortcite{kupiec1995trainable,jones2007automatic}.

More recently, summarization approaches rely instead on neural networks to obtain deep representations of content units by means of 
convolutional neural networks \shortcite{perez2019generating,narayan2019article}, recurrent neural networks \shortcite{narayan2018document,narayan2018ranking,cheng2016neural}, Transformers \shortcite{song2019mass,dong2019unified}
and lately by leveraging large pretrained language models \shortcite{zheng2019pacsum,liu2019text,zhang2020pegasus}.
Building upon traditional methods, neural summarization models leverage discourse \shortcite{clarke2010discourse,cohan2018discourse}, topical \cite{narayan2019article}, and graph representations \shortcite{graphsumsurvey21,qiu-cohen-2022-abstractive}.
Even though most research concentrates on summarization of middle-sized documents like news articles and Reddit posts \cite{volske2017tl}, recent work has shifted attention to long document summarization and its challenges \cite{cohan2018discourse,sharma2019bigpatent,xiao2019extractive,fonseca2022factorizing}.
Among recent efforts, it is worth mentioning architectures tailored to consume longer inputs by reducing the time complexity of the attention mechanism \shortcite{Beltagy2020Longformer,wang2020linformer,huang2021efficient}
or leveraging the structure of the input document \shortcite{cohan2019pretrained,narayan2020stepwise}.
The present work follows this line of research by introducing summarization systems capable of consuming long documents and extracting a summary in linear time w.r.t.\ the number of sentences.
Note that the proposed systems do not employ neural networks during content representation or selection but instead operate over cognitively inspired data structures of propositions representing human memory.

Finally, of special interest to this work are unsupervised approaches to summarization, an area not explored as much as its supervised counterpart given the availability of large summarization datasets nowadays \shortcite{hermann2015teaching,cohan2018discourse,narayan2019article}.
Central to most extractive approaches is a weighted graph representation of the source document \cite{graphsumsurvey21} followed by sentence ranking based on node centrality, where edge weights are calculated by TF-IDF \cite{mihalcea2004textrank} or by finetuned, dedicated architectures \cite{zheng2019pacsum}.
Our work differs from this line of research in two aspects.
First, content is organized in tree and graph structures where nodes are modeled as propositions instead of sentences or words.
However, content selection is still performed at the sentence level.
Second, the proposed node scoring strategy exploits cognitively-grounded properties of human memory structures.
We demonstrate through extensive experiments that this scoring strategy outperforms previously proposed systems that model sentence relevancy through centrality.

\subsection{Informativeness, Redundancy, and Cohesion}
Traditional summarization approaches 
sought to provide more control over properties of generated summaries such as their informativeness \shortcite{jones1993might,carbonell1998use,nenkova2011automatic,lloret2012text,teufel2016deeper}, non-redundancy \shortcite{carbonell1998use}, or discourse organization \shortcite{marcu1998build,christensen2013towards}.
However, more modern approaches mostly employ neural end-to-end models \shortcite{cheng2016neural,lewis2020bart,zhang2020pegasus}, meaning that crucial intermediate steps such as content planning or selection are not explicitly modeled.
Recent efforts have demonstrated that accounting for planning helps dealing with discourse organization of final summaries \shortcite{goldfarb2020content,sharma2019bigpatent,hua2021dyploc}, whereas explicit content selection modules can be tailored to tackle problems such as factuality \shortcite{cao2018faithful,maynez-etal-2020-faithfulness,zhao-etal-2020-reducing}, coverage \shortcite{kedzie2018content,puduppully2019data,wiseman2017challenges}, 
and redundancy \shortcite{liu2019text,jia2021deep,bi2021aredsum}.
Specifically, production of low-redundant summaries has proven to be challenging, especially when the source document is highly redundant, such as scientific articles \shortcite{xiao2020systematically,gu2022memsum}.

Regarding cohesion, \citeA{wu2018learning} combined an informativeness reward with a cohesion reward in a reinforcement learning setup,
reporting heavy trade-offs between the two properties.
In contrast, \citeA{xu2020discourse} reported an improvement in informativeness when incorporating information about the global discourse organization (RST trees) and co-reference chains in a supervised setup.
However, discourse organization of final summaries experimented only a marginal improvement.

In this work, we demonstrate that it is possible to improve lexical cohesion --a special case of local coherence-- while
maintaining a high level of informativeness and without selecting overly redundant content, under a reinforcement learning setup.
In the unsupervised scenario, our proposed models are able to successfully balance cohesion and redundancy, although sacrificing 
informativeness in the process.

\subsection{Cognitive Models for Summarization}
In psycholinguistics, summarization as a task is often used as a method to investigate cognitive processes involved in text comprehension and production \shortcite{kintsch1978toward,kintsch1990macroprocesses,lehto1996working,kintsch1998comprehension,ushiro2013measures,spirgel2016does}.
Such processes are in charge of generalizing, synthesizing, and coherently organizing content units.
Comprehension, in turn, is modeled after psycholinguistic models of human reading comprehension \cite{kintsch1978toward,kintsch1988role} which provide a rich and robust theoretical foundation on how content units are discretized and manipulated by cognitive processes.
For this reason, comprehension models such as the Micro-Macro Structure (KvD; Kintsch and van Dijk, 1978\nocite{kintsch1978toward}) and Construction-Integration theory (CI; Kintsch, 1988\nocite{kintsch1988role}), have drawn the attention of researchers in automatic summarization in recent years \shortcite{fang2014summariser,zhang2016coherent,fang2019proposition}.
These theories outline procedures to discretize content into semantic propositions and build text representations that account for
local and global coherence.
However, computational implementations proposed so far \shortcite{fang2014summariser,zhang2016coherent,fang2019proposition}
show a heavy reliance on NLP tools such as entity extractors and coreference resolution systems, as well as external resources like WordNet \cite{miller1995wordnet}.
These requirements greatly limit their application in highly technical domains such as scientific literature.
Additionally, many design choices prevented these systems from exploiting properties of memory structures, modeling retrieval processes, or manipulating information at the right granularity level, e.g.\ ranking words or sentences instead of semantic propositions.

We address these limitations by introducing two computational implementations of the KvD theory \cite{kintsch1978toward}
that require only a dependency parser and no external resources, making it possible to test these systems on other languages and domains.
Moreover, our proposed systems better exploit memory structure properties and retrieval processes during reading simulation, 
which makes them capable of producing notoriously less redundant and more cohesive summaries than strong baselines.

\section{Reward-guided Control Scenario}
\label{section:rl-models}
In this section, we formulate the first scenario in which sentence scoring is guided by explicit rewards that encourage informativeness, non-redundancy, and local coherence in candidate summaries, in a reinforcement learning training setup.
We posit the task of extractive summarization as the task of scoring the sentences in a document followed by a selection step
in which an optimal set of sentences is chosen as the summary.
The scoring step is formulated as a sequence labeling task where
each sentence in a document $\mathcal{D}=\langle s_0,..,s_k,...,s_{|D|} \rangle$ 
is labeled with $y_i \in \{0,1\}$, indicating whether sentence $s_i$ should be selected or not.
A summarization system $M$ assigns score $p(y_i=1 \mid s_i)$ indicating the preference in selecting $s_i$ according to a criteria modeled by $M$.
Then, candidate summary $\hat{S}$ is obtained by concatenating the top-scoring sentences, selected greedily and with a predefined budget in number of tokens.
We focus on informativeness, non-redundancy, and local coherence, as preference modeling criteria.

We build upon the model proposed by \citeA{xiao2020systematically}, consisting of an encoder that incorporates local and global context, a feed-forward layer as a decoder, 
and trained with the Cross-Entropy loss ($\mathcal{L}_{\text{CE}}$) over the sequence labeling task outlined above.
In the rest of this paper, we refer to this supervised model as \textsc{E.LG}.

Then, we adapt previous work on reinforcement learning-based approaches that aim to optimize for informativeness and either redundancy or local coherence.
We define reward $r_{\text{I}}$, aimed at encouraging the selection of informative summaries \shortcite{dong-etal-2018-banditsum}, as
\vspace{-0.2cm}
\begin{equation*}
    r_\text{I}=\frac{1}{3}
    \Big(\text{ROUGE-1} + \text{ROUGE-2} + \text{ROUGE-L}\Big),
\end{equation*}
\noindent where ROUGE F$_1$ scores are calculated using the reference summaries.
Next, we define models employing policy gradient methods that maximize a reward function combining $r_\text{I}$ with redundancy or coherence-aware rewards.

\subsection{Informativeness Encoder}
We employ the model proposed by \citeA{xiao2019extractive} optimized to encode only informativeness during sentence scoring.
The model incorporates local and global information by taking into account the document structure (e.g.\ section separation) and 
The model, which we label \textsc{E.LG} in this chapter, consists of a document encoder and a decoder that classifies whether a sentence should be selected or not.

\textbf{Document Encoder.}
Given document $\mathcal{D}=\langle s_0,..,s_k,...,s_{|D|} \rangle$, where $s_i$ is a sequence of tokens, sentence embedding $h_i$, is defined as the average token embedding of its constituent tokens.
Then, global sentence representations are obtained using a bi-directional RNN \cite{schuster1997bidirectional} with GRU cells \cite{cho2014properties}, i.e.\  $h^g_i = [f_i,b_i]$, where $f_i$ and $b_i$ are the forward and backward hidden state at step i, respectively.
Moreover, let $d=[f_{|D|};b_0]$ be the representation of the whole document.

The document structure is incorporated explicitly with section representations.
Let $\mathcal{D}$ bet organized in sections represented as a list of sentences, $[ [s_0,..,s_i],[s_{i+1},..,s_j],[s_{j+1},..s_k]... ]$,
the embedding of each section is defined as the difference of hidden states corresponding to sentences in the section borders.
For instance, the embedding of section $[s_{i+1},..,s_j]$ is defined as 
$l_1 = [f_{j+1} – f_{i+1};b_{i+1}-b_{j}]$.

\textbf{Decoder.} 
After obtaining sentence as well as global (the entire document) and local context representations (sections), the decoder will combine them using attention, as follows.
Given document embedding $d$, sentence global embedding $h^g_i$, and section embedding $l_t$, where $s_i$ belongs to section $t$, the final sentence representation $z_i$ is obtained as follows,
\begin{align}
    e^d_i &= v^T tanh(W^a [d;h^g_i]), \text{ } e^l_i = v^T tanh(W^a [l_t;h^g_i]), \notag \\
    w^d_i &= \frac{e^d_i}{e^d_i  + e^l_i}, \text{ } w^l_i = \frac{e^l_i}{e^d_i  + e^l_i}, \notag \\
    c_i &= w^d_i d + w^l_i l_t, \notag \\
    z_i &= [h^g_i;c_i],
\end{align}
\noindent where $v^T, W^a$ are weight parameters.
Finally, the probability of selecting $s_i$ is given by $p(y_i=1 | s_i;\theta) = \sigma(ReLU(W^o z_i))$,
where $\theta$ represents the model parameters and $W^o$ is a weight parameter.

The \textsc{E.LG} model just described is trained with the Cross-Entropy loss (CE) over the sequence labeling task outlined at the beginning of this section.

\subsection{Informativeness and Redundancy}
\label{section:elg-coh}
We adapt \textsc{MMR-Select+} \cite{xiao2020systematically}, the strategy most capable of balancing informativeness and redundancy.
Model \textsc{E.LG} is trained using a combined loss that aims to minimize Cross Entropy loss and maximize the expected reward of greedily sampled summary $\hat{S}$ \shortcite{qian2019reducing}, defined as
\vspace{-0.2cm}
\begin{align*}
    \mathcal{L} &= \gamma_\text{R} \cdot \mathcal{L}_{\text{R}} + (1 - \gamma_\text{R}) \cdot \mathcal{L}_{\text{CE}}, \\
    \mathcal{L}_{\text{R}} &= -(r_\text{I}(\hat{S}) - r_\text{I}(\bar{S})) \sum_{s_i \in \hat{S}} \log p(y_i \mid s_i)
\end{align*}
where $r_\text{I}(\bar{S})$ is the informativeness of a baseline summary, used to improve convergence in a self-critic fashion \shortcite{paulusdeep}.
Baseline summary $\bar{S}$ is extracted using greedy selection directly over $p(y_i)$, whereas
 $\hat{S}$ is extracted greedily using redundancy-aware score $p_{\text{MMR}}$
 \vspace{-0.2cm}
\begin{equation*}
    p_{\text{MMR}}(y_i | s_i) = \lambda_{\text{R}} \cdot p(y_i \mid s_i) - (1 - \lambda_{\text{R}}) \cdot \max_{s_j \in \hat{S}} \text{Sim}(s_i,s_j),
\end{equation*}
where $\text{Sim}(s_i,s_j)$ is the cosine similarity between embeddings 
of sentences $s_i$ and $s_j$
and $\lambda_{\text{R}}$ controls the redundancy level in $\hat{S}$.
This scoring strategy is an extension of MMR \cite{carbonell1998use} that aims to minimize semantic similarity between sentences in $\hat{S}$.
In our experiments, we dub this model as \textsc{E.LG-MMRSel+}.
    
\subsection{Informativeness and Local Coherence}
Building upon \citeA{wu2018learning}, we define a 
reward that combines informativeness and local coherence, $\displaystyle r = \lambda_{\text{LC}} \cdot r_\text{I} + (1 - \lambda_{\text{LC}}) \cdot r_{\text{LC}}$,
where $\lambda_{\text{LC}}$ controls the trade-off between informativeness and coherence and $r_{\text{LC}}$ is a local coherence scorer.
Then, \textsc{E.LG} is trained using the REINFORCE algorithm \cite{williams1992simple} with policy gradient
\begin{equation*}
    \nabla \mathcal{L} = -r(\hat{S}) \sum_{s_i \in \hat{S}} \nabla \log p(y_i \mid s_i),
\end{equation*}
where $\hat{S}$ is a candidate summary extracted greedily directly form $p(y_i \mid s_i)$.
In our experiments, we label this model as \textsc{E.LG-CCL}.

\paragraph{Local Coherence Scorer.}
Scorer $r_{\text{LC}}$ receives a multi-sentence text and assigns a score between $[0,1]$ quantifying its local coherence, and it is defined as follows.
Following the methodology of \citeA{steen-markert-2022-find}, we train a RoBERTa model \shortcite{roberta-model} to distinguish shuffled from unshuffled summaries.
The model is trained in a binary classification setup with 
chunks of \num{3} consecutive sentences as positive class and 
their shuffled versions as negative class.
Then, the local coherence score of a summary is defined as the positive class probability,
averaged over windows of \num{3} sentences taken with padding of one sentence.


\section{Unsupervised Control Scenario}
\label{section:unsup-scenario}
In this section, we formulate the second scenario, i.e.\ controlling for informativeness, non-redundancy, and cohesion in candidate summaries in an unsupervised setup.
Similarly to the first scenario, we formulate the task of extractive summarization as a two-step process, sentence scoring and sentence selection.
During sentence scoring, the document is consumed one sentence at a time, updating the score of a subset of sentences at each step.
Then, the top-scoring sentences are selected according to a predefined budget.

This section is organized as follows.
First, we elaborate on the Micro-Macro Structure theory of reading comprehension, KvD, 
explain in detail how it simulates short-term memory, and discuss how its operationalization can be leveraged for sentence scoring in extractive summarization.
Then, we introduce two novel computational implementations of the KvD theory tailored to sentence scoring.

\subsection{The KvD Theory of Human Memory}
\label{section:kvd-model}
Proposed by \citeA{kintsch1978toward}, the Micro-Macro Structure theory describes the cognitive processes involved in
text (or speech) comprehension, and provides a principled way to make predictions about the content 
human subjects would be able to recall later.
In this theory, discourse comprehension is performed at two levels, micro and macro-level, and discourse is represented with a characteristic structure of content at each level.
At the micro level, content structure is modeled after working memory --a type of short-term memory-- and KvD defines precise mechanisms that update and reinforce content in the structure.
Content at this level is discretized in basic meaningful units by means of linguistic propositions.
A proposition is denoted as \texttt{predicate(arg$_1$,arg$_2$,...)} 
where arg$_i$ is a syntactic argument of the predicate (e.g.\ argument to a transitive verb).
As such, propositions can be interpreted as clauses or short sentences and hence provide more expressivity than word units during comprehension.
The advantage of using propositions as content units goes beyond the amount of information it can pack.
A proposition can be linked to another either syntactically or semantically, potentially building entire connected structures of propositions.
According to KvD theory, working memory holds a cohesive organization of content units by making sure that all units are connected e.g.\ in a connected tree.
Hence, the resulting micro-structure models cohesive ties in the text.

At the macro level, content structure represents the global organization of the text and its building is guided by the reader's goals in mind whilst performing the task. For instance, if the task is summarization, KvD defines macro-processes concerned with generalization, fusion, insertion of details from background knowledge, among others.
In this work, we consider only the structures represented at the micro level and leverage them for the task of extractive summarization. Structures and processes at the macro level would require human-like reasoning and intuition and even though recent work on neuro-symbolic systems \shortcite{garcez2020neurosymbolic,bengio2017consciousness} and common-sense reasoning \shortcite{speer2017conceptnet,bosselut2019comet} show a promising development path, we leave this path out of the scope of this work and for future work.

Experimentally, \citeA{kintsch1978toward} tested the model on the tasks of recall and summarization which required human subjects to write down a short text after reading a document.
The recall task aimed to measure how accurately subjects can reproduce specific propositions from the given document,
whereas the summarization task aimed to quantify how many summary-worthy propositions are retrieved.
The KvD theory models the probability of writing down (\textit{to reproduce}) a proposition as a function of the frequency with which it was retained in working memory.
The longer a proposition remained in working memory, either at the micro level or at the macro level, the higher its reproduction probability.
This probability is then used to makes predictions about what content is more likely to be written down in a summary.
Crucial to our work, KvD argued that reproduction probability can be used as a numerical score to rank propositions.
In an extractive summarization setup, this score can be used to rank content and select them accordingly.
We elaborate on how to define such a score function in detail in Section~\ref{section:kvd-sys}, along with our computational implementations of the KvD theory.


\subsubsection{Memory Simulation at Micro Level}

At the micro level, content is organized in a data structure representing working memory called the \textit{memory tree},
where each node corresponds to a proposition and two propositions are connected if any of their arguments overlap.

According to KvD, reading is carried out iteratively in \textit{memory cycles}.
In each cycle, only one new sentence is loaded to the working memory, where its propositions are extracted and added to the current memory tree.
The limits of memory capacity is modeled as a hard constraint in the number of propositions that will be preserved for the next cycle.
Hence, the tree is pruned and some propositions are dropped or \textit{forgotten}.
However, if nodes cannot be attached to the tree in upcoming cycles, forgotten nodes can be recalled and added to the tree, serving as linking ideas that preserve the cohesion represented in the current tree.\footnote{It is worth noting that \citeA{kintsch1978toward} did not specify how many nodes can be recalled at a single time, however, recent implementations \cite{fang2019proposition} limit this number to at most 2.}
Whenever the content in working memory is changed, whether adding propositions or removing them, 
the root is reassigned to the node containing information central to the argumentation represented in working memory.
We now illustrate with an example how content units are captured, forgotten, and recalled during a KvD simulation of reading.

Consider the first three sentences of the introduction section of a biomedical article, along with its abstract, shown in Figure~\ref{fig:example-kvd}.
At the beginning of cycle 1, propositions 1 to 7 are extracted from the incoming sentence and populate an empty working memory, resulting in tree (1a).
Note that the root, node 4, includes the main verb of the sentence and links the main actors (\texttt{antioxidants}, \texttt{species}, and \texttt{people}).
Note also that connected propositions present arguments in common, e.g.\ node 5 and 6 share the argument \texttt{antioxidants}.
Then, the memory capacity constraint is enforced by pruning nodes until the tree is of a predetermined size.
In this example, we set the memory limit to 5 propositions per cycle.
KvD introduced the \textit{leading edge} strategy for pruning, which traverses the tree in depth-first order starting from the root and selects only the most recent node (in order of reading) at each step. In case a leaf node is reached and there is capacity left, the tree is traversed in breath first order starting from the root, and selects nodes with the same criteria, until capacity is reached.
In cycle 1, the selected nodes from tree (1a) are 4, 5, 7, 3, and 2, in that order.
The remaining nodes, 1 and 6, are pruned.
Since content in working memory has been reduced, the root must be reassigned if needed. However, node $4$ remains central, hence it remains as the root
and we move on to the next cycle with tree (1b) as memory tree.
These pruned trees constitute the final product of each cycle and will be used for our content selection experiments.

\begin{figure}[t]

\centering
\small
\begin{tabular}{p{7.5cm}p{5cm}}
\toprule
\multicolumn{2}{p{15cm}}{\textbf{Cycle 1}\newline
In healthy people, reactive oxidant species are controlled by a number of enzymatic and non-enzymatic antioxidants.} \\

\begin{tabular}[t]{p{9cm}}
\\
  \texttt{1: in people(healthy)} \\
  \texttt{2: species(reactive)} \\
  \texttt{3: species(oxidant)} \\
  \texttt{4: are controlled(antioxidants,species, \newline people)} \\
  \texttt{5: of(a number, antioxidants)} \\
  \texttt{6: antioxidants(enzymatic)} \\
  \texttt{7: antioxidants(non-enzimatic)}
\end{tabular}

& \begin{tabular}[t]{c}
\tikzset{every picture/.style={line width=0.75pt}} 
\scalebox{0.9}{
\begin{tikzpicture}[x=0.75pt,y=0.75pt,yscale=-1,xscale=1,baseline=0pt]

\draw  [dash pattern={on 0.84pt off 2.51pt}]  (25,20) -- (48.33,20) ;
\draw    (25,20) -- (49.33,39.67) ;
\draw    (25,20) -- (50.33,58.67) ;
\draw    (25,20) -- (50.33,79.67) ;
\draw  [dash pattern={on 0.84pt off 2.51pt}]  (63.33,81) -- (89.33,81) ;
\draw    (63.33,81) -- (90.33,99.67) ;
\draw    (125,20) -- (149.29,20) ;
\draw    (125,20) -- (149.29,40.43) ;
\draw    (125,20) -- (149.29,60.43) ;
\draw    (163.33,61) -- (179.29,61) ;

\draw (52,33) node [anchor=north west][inner sep=0.75pt]  [align=left] {2};
\draw (52,13) node [anchor=north west][inner sep=0.75pt]  [align=left] {1};
\draw (12,13) node [anchor=north west][inner sep=0.75pt]  [align=left] {4};
\draw (52,53) node [anchor=north west][inner sep=0.75pt]  [align=left] {3};
\draw (52,73) node [anchor=north west][inner sep=0.75pt]  [align=left] {5};
\draw (92,74) node [anchor=north west][inner sep=0.75pt]  [align=left] {6};
\draw (92,93) node [anchor=north west][inner sep=0.75pt]  [align=left] {7};
\draw (152,13) node [anchor=north west][inner sep=0.75pt]  [align=left] {2};
\draw (112,13) node [anchor=north west][inner sep=0.75pt]  [align=left] {4};
\draw (152,33) node [anchor=north west][inner sep=0.75pt]  [align=left] {3};
\draw (152,53) node [anchor=north west][inner sep=0.75pt]  [align=left] {5};
\draw (182,54) node [anchor=north west][inner sep=0.75pt]  [align=left] {7};
\draw (-15,10) node [anchor=north west][inner sep=0.75pt]   [align=left] {(1a)};
\draw (85,10) node [anchor=north west][inner sep=0.75pt]   [align=left] {(1b)};

\end{tikzpicture}
}
\end{tabular} 
\vspace{0.01cm}
\\ 

\multicolumn{2}{p{15cm}}{
\textbf{Cycle 2}\newline
In patients with Cystic Fibrosis (CF), deficiency of nonenzymatic antioxidants is linked to malabsortion of lipid-soluble vitamins.} \\

\begin{tabular}[t]{p{10cm}}
\\
	\texttt{8: with(in patients, Cystic Fibrosis)} \\
	\texttt{9: BE(Cystic Fibrosis,CF)} \\
	\texttt{10: of(deficiency, \$7)} \\
	\texttt{11: is linked (deficiency,malabsortion, \$8)} \\
	\texttt{12: of (malabsortion,vitamins)} \\
	\texttt{13: vitamins(lipid-soluble)}
\end{tabular}

& \begin{tabular}[t]{c}

\tikzset{every picture/.style={line width=0.75pt}} 

\scalebox{0.75}{
\begin{tikzpicture}[x=0.75pt,y=0.75pt,yscale=-1,xscale=1,baseline=0pt]

\draw  [dash pattern={on 0.84pt off 2.51pt}]  (19.29,20.33) -- (34.98,20.33) ;
\draw  [dash pattern={on 0.84pt off 2.51pt}]  (74.33,20.33) -- (90.33,20.33) ;
\draw  [dash pattern={on 0.84pt off 2.51pt}]  (74.33,20.33) -- (91.29,39.86) ;
\draw    (19.29,20.33) -- (28.29,90.86) ;
\draw  [dash pattern={on 0.84pt off 2.51pt}]  (77.33,89.33) -- (87.33,89.33) ;
\draw  [dash pattern={on 0.84pt off 2.51pt}]  (105.33,89.33) -- (117.29,89.33) ;
\draw    (77.33,89.33) -- (86.33,119.33) ;
\draw    (47.33,89.33) -- (59.33,89.33) ;
\draw    (106.33,118.33) -- (117.29,118.33) ;
\draw    (160.33,19.33) -- (172.33,19.33) ;
\draw    (187.33,49.33) -- (197.14,49.33) ;
\draw    (214.33,49.33) -- (224.14,49.33) ;
\draw    (160.33,19.33) -- (170.14,48.86) ;
\draw  [dash pattern={on 0.84pt off 2.51pt}]  (47.29,20.33) -- (60.98,20.33) ;

\draw (31,13) node [anchor=north west][inner sep=0.75pt]  [align=left] {5};
\draw (8,13) node [anchor=north west][inner sep=0.75pt]  [align=left] {7};
\draw (30,82) node [anchor=north west][inner sep=0.75pt]  [align=left] {10};
\draw (62,13) node [anchor=north west][inner sep=0.75pt]  [align=left] {4};
\draw (90,13) node [anchor=north west][inner sep=0.75pt]  [align=left] {2};
\draw (90,33) node [anchor=north west][inner sep=0.75pt]  [align=left] {3};
\draw (60,82) node [anchor=north west][inner sep=0.75pt]  [align=left] {11};
\draw (87,112) node [anchor=north west][inner sep=0.75pt]  [align=left] {12};
\draw (91,82) node [anchor=north west][inner sep=0.75pt]  [align=left] {8};
\draw (117,82) node [anchor=north west][inner sep=0.75pt]  [align=left] {9};
\draw (117,112) node [anchor=north west][inner sep=0.75pt]  [align=left] {13};
\draw (141,13) node [anchor=north west][inner sep=0.75pt]  [align=left] {10};
\draw (172,43) node [anchor=north west][inner sep=0.75pt]  [align=left] {11};
\draw (197,43) node [anchor=north west][inner sep=0.75pt]  [align=left] {12};
\draw (222,43) node [anchor=north west][inner sep=0.75pt]  [align=left] {13};
\draw (173,13) node [anchor=north west][inner sep=0.75pt]  [align=left] {7};
\draw (-20,10) node [anchor=north west][inner sep=0.75pt]   [align=left] {(2a)};
\draw (105,10) node [anchor=north west][inner sep=0.75pt]   [align=left] {(2b)};

\end{tikzpicture}
}
\end{tabular} 
\vspace{0.01cm}
\\ 

\multicolumn{2}{p{15cm}}{\textbf{Cycle 3}\newline
Furthermore, pulmonary inflammation in CF patients also contributes to depletion of antioxidants.} \\

\begin{tabular}[t]{p{6cm}}
\\
	\texttt{14: inflammation(pulmonary)} \\
	\texttt{15: inflammation(in:\$8)} \\
	\texttt{16: contributes(\$15,to:depletion)} \\
	\texttt{17: of(depletion,antioxidants)}
\end{tabular}
 
& \begin{tabular}[t]{c}

\tikzset{every picture/.style={line width=0.75pt}} 
\scalebox{0.75}{
\begin{tikzpicture}[x=0.75pt,y=0.75pt,yscale=-1,xscale=1,baseline=0pt]

\draw  [dash pattern={on 0.84pt off 2.51pt}]  (31.33,20.33) -- (43.33,20.33) ;
\draw  [dash pattern={on 0.84pt off 2.51pt}]  (58.33,50.33) -- (68.14,50.33) ;
\draw  [dash pattern={on 0.84pt off 2.51pt}]  (85.33,50.33) -- (93.4,50.33) ;
\draw    (31.33,20.33) -- (41.14,49.86) ;
\draw    (84.33,80.33) -- (92.4,80.2) ;
\draw  [dash pattern={on 0.84pt off 2.51pt}]  (111.33,80.33) -- (118.4,80.33) ;
\draw    (58.33,50.33) -- (69.4,81.2) ;
\draw   (69.4,71.2) -- (84.4,71.2) -- (84.4,90.2) -- (69.4,90.2) -- cycle ;
\draw    (111.33,80.33) -- (118.4,110.2) ;
\draw  [dash pattern={on 0.84pt off 2.51pt}]  (135.33,110.33) -- (143.4,110.33) ;
\draw   (202.4,40.2) -- (217.4,40.2) -- (217.4,59.2) -- (202.4,59.2) -- cycle ;
\draw    (190,20) -- (202.14,49.86) ;
\draw    (190,20) -- (202,20) ;
\draw    (218,49) -- (230,49) ;
\draw    (247,49) -- (255,49) ;

\draw (12,14) node [anchor=north west][inner sep=0.75pt] [align=left] {10};
\draw (43,44) node [anchor=north west][inner sep=0.75pt]  [align=left] {11};
\draw (68,44) node [anchor=north west][inner sep=0.75pt]  [align=left] {12};
\draw (93,44) node [anchor=north west][inner sep=0.75pt]  [align=left] {13};
\draw (44,14) node [anchor=north west][inner sep=0.75pt]   [align=left] {7};
\draw (72,73) node [anchor=north west][inner sep=0.75pt]  [align=left] {8};
\draw (93,73) node [anchor=north west][inner sep=0.75pt]  [align=left] {15};
\draw (118,73) node [anchor=north west][inner sep=0.75pt]  [align=left] {14};
\draw (118,103) node [anchor=north west][inner sep=0.75pt]  [align=left] {16};
\draw (144,103) node [anchor=north west][inner sep=0.75pt] [align=left] {17};
\draw (172,13) node [anchor=north west][inner sep=0.75pt]  [align=left] {11};
\draw (202,13) node [anchor=north west][inner sep=0.75pt]  [align=left] {10};
\draw (204.4,43.2) node [anchor=north west][inner sep=0.75pt]  [align=left] {8};
\draw (230,42) node [anchor=north west][inner sep=0.75pt]  [align=left] {15};
\draw (255,42) node [anchor=north west][inner sep=0.75pt]  [align=left] {16};
\draw (-20,10) node [anchor=north west][inner sep=0.75pt]   [align=left] {(3a)};
\draw (130,10) node [anchor=north west][inner sep=0.75pt]   [align=left] {(3b)};

\end{tikzpicture}
}
\end{tabular} \vspace{0.01cm}
\\ 
\midrule 

\multicolumn{2}{p{15cm}}{\textbf{Gold Summary}\newline Patients with Cystic Fibrosis (CF) show decreased plasma concentrations of antioxidants due to malabsorption of lipid-soluble vitamins and consumption by chronic pulmonary inflammation.\newline
Carotene is a major source of retinol and therefore is of particular significance in CF. ...} \\ \bottomrule
\end{tabular}
       
\caption{Simulation of KvD reading during three cycles. Each row shows the sentence consumed (top),
	the propositions extracted (left), and memory trees before (1a, 2a, 3a) and after (1b, 2b, 3b) applying a memory constraint of 5 nodes.
	Argument \texttt{\$N} means that proposition \texttt{N} is used as argument.
	Squared nodes are recalled propositions.
	Solid lines connect nodes selected to keep in memory, and dotted lines connect nodes to be pruned. }
\label{fig:example-kvd}
\end{figure}

In cycle 2, propositions \num{8} to \num{13} are added to memory, tree (1b).
In the presence of this new information, the root is reassigned to a proposition central to all the propositions in memory.
In this case, node $7$ is made root because it presents information common to both sentences (\texttt{nonenzimatic antioxidants}), hence being central.
Note also that the new tree (2a) showcases clearly two ramifications of the current topic, namely that \texttt{\$7}\textit{` control a specific kind of molecules'} and \textit{`deficit of} \texttt{\$7} \textit{causes certain condition'}.
Then, we apply the \textit{leading edge} strategy to select nodes \num{7}, \num{10}, \num{11}, \num{12}, and \num{13}, in that order, and prune the rest.
Since the content of the working memory has changed again, node \num{10} is now deemed as central and assigned root status, resulting in tree (2b).

In cycle 3, the newly extracted nodes (\num{14} - \num{17}) cannot be attached to the current tree because the linking node, \texttt{\$8}, was pruned in the previous cycle.
Therefore, proposition \num{8} is \textit{recalled} and re-attached to the tree, shown as a squared node in tree (3a) and (3b).
Then, the selection strategy is applied and node 11 is selected as new root, obtaining (3b).

After analyzing how trees are shaped in each cycle, it is important to point out their importance for the task of extractive summarization.
Next, we elaborate on how memory trees can be leveraged for this end.

\subsubsection{Properties Relevant to Summarization}
\label{section:prop-kvd}
The procedure for content manipulation described above imposes constraints on the shape, size, and content of memory trees during simulation. 
Such constraints bestow memory trees with special properties relevant to the task of summarization,
specifically with respect to cohesion, relevancy, and redundancy.

\paragraph{Local Coherence and Cohesion.}
A memory tree constitutes a connected structure in which two propositions are connected if any two of their arguments refer to the same concept. Connectivity, \citeA{kintsch1978toward} argued, is a consequence of the text being well-structured and locally coherent, although connectivity is not a necessary condition for coherence --a disconnected structure can still be coherent for a reader.
In this way, KvD enforces local coherence in a memory tree in the form of lexical cohesion.
For instance, proposition \num{8} in cycle 3 of Figure~\ref{fig:example-kvd} serves as a bridge to keep the memory tree connected, since propositions talking about 
\textit{CF patients} (propositions \num{8} and \num{9}) were discarded in the previous cycle.

This connectivity property has the following implication for cohesion in a final summary.
By retaining a set of cohesive content units in working memory, their reproduction probability is increased.
Consequently, cohesive groups of propositions will present similar scores at the end of the simulation, encouraging the selection of content that reads more cohesive as a whole.

\paragraph{Relevancy.}
In addition to being locally coherent, memory micro-structure takes the form of a tree for the following reasons.
KvD states that the root of a memory tree should contain information central to the argumentation represented in the working memory; hence, the root is deemed as the most relevant proposition in memory, and the more relevant a proposition is, the closer to the root it will be.
This property could be exploited by a summarization system by designing a scoring function that takes the position of a tree node into account.

However, a KvD-based sentence ranking system that relies on proposition scoring would first need to capture the right propositions in working memory.
Let us look at the first sentence of the gold summary in Figure~\ref{fig:example-kvd}).
On the one hand, many propositions ($7$, $8$, $12$, $13$, and $15$) appear verbatim in this sentence, although sometimes only partially (e.g.\ $7$ and $15$).
The capture of proposition $8$ in cycle 3 highlights the importance of the recall mechanism in KvD to bring back relevant information.
On the other hand, fine-grained information relevant to the summary might also be lost, such as node $14$, in which a crucial property of a noun is not captured (`\textit{pulmonary}').

\paragraph{Redundancy.}
Finally, KvD processes influence redundancy reduction in two accounts.
First, propositions in a memory tree are connected such that each proposition adds new details about a concept without encoding more redundant arguments than necessary.
For instance, consider again proposition \num{2} and \num{3} in Figure~\ref{fig:example-kvd}, 
where both propositions add relevant details (\textit{reactive} and \textit{oxidant}) about a concept (\textit{species}).
Hence, memory trees constitute a representation with the maximum amount of relevant details that can fit in working memory whilst minimizing the redundancy of arguments.

Second, in case the recall mechanism needs to be used, KvD retrieves only the minimum amount of propositions to serve as a bridge and connect the incoming propositions.
Specifically, the recall mechanism only adds one recall path to the memory tree instead of many other alternative paths.
By not loading redundant paths into memory, a system could avoid increasing the score of redundant content and update only one recall path at a time.
This behavior, as we will demonstrate later, contributes immensely to decrease redundancy in the final summary and becomes particularly important for highly redundant documents, e.g.\ scientific articles that repeat information in several sections.

\subsection{Unsupervised Summarization as Human Memory Simulation}
\label{section:kvd-sys}

In this part, two sentence scoring systems are introduced, \textsc{TreeKvD} and \textsc{GraphKvD}, 
which at their core simulate human working memory during reading, according to the KvD theory.
We start by providing an overview of the implemented summarization pipeline.
Then, we elaborate on the procedure used to build propositions from syntactic structures automatically extracted from text.
Finally, we present the proposed sentence scoring systems in detail and discuss the design choices made,
and complement the explanation with a simulation example.

\subsubsection{Pipeline Overview}

\begin{figure}[t]
     \centering
     \includegraphics[width=\textwidth]{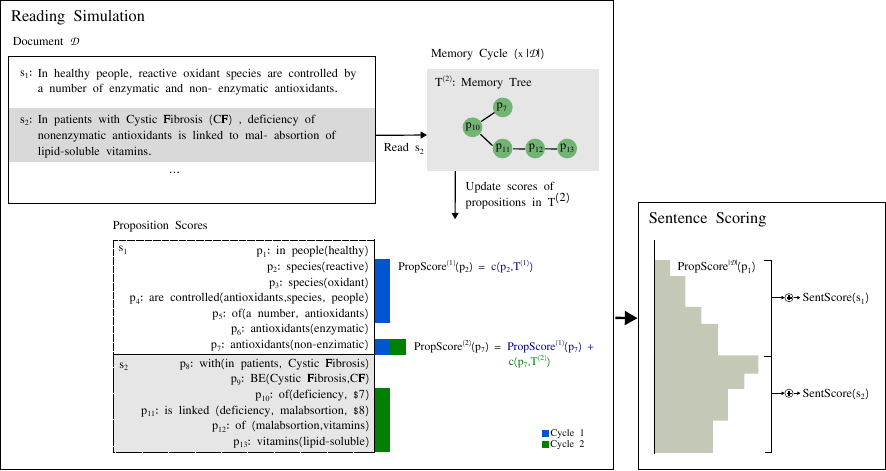}
    \caption{Pipeline of KvD reading simulation and sentence scoring using the simulation example in Fig.2.}
    \label{fig:kvd-pipeline}
\end{figure}

The pipeline for sentence scoring is depicted in Figure~\ref{fig:kvd-pipeline}.
Input document $\mathcal{D}$ is consumed one sentence at a time by the reading simulator.
At each step, one memory cycle is executed and the scores of the propositions in the working memory tree are updated.
Once the document has been completely read, the final score of propositions
is aggregated into sentence scores, which are then used to select the final summary.

\paragraph{Reading Simulation.}

The proposed KvD simulators model how content is moved from working memory to long-term memory and vice versa.
Working memory is represented as a proposition tree, pruned at the end of each cycle in order to simulate short-term memory limitations in humans.
In contrast, long-term memory is represented as an undirected graph of propositions populated by nodes demoted from working memory as reading progresses. 

The outline of the the simulation procedure is presented in Algorithm~\ref{alg:kvd}.
The algorithm consumes a document $\mathcal{D}=\langle s_0,\ldots,s_k,\ldots,s_{|\mathcal{D}|} \rangle$ iteratively in memory cycles, updating working memory and long-term memory in each cycle.
At the beginning of cycle $k$, the algorithm reads sentence $s_k$, extracts its proposition tree $P_k$ (Line 6), and attaches it to the current memory tree $T$ (Line 7).
The resulting tree is pruned to a constant size (Line 10) in order to simulate human memory constraints,
and pruned nodes are added to the long-term memory graph $G$.
Then, the score of proposition $t$ in cycle $k$ (Line 11) is updated to
\vspace{-0.1cm}
\begin{equation}
    \label{eq:pscore}
    \text{PropScore}^k(t) = \text{PropScore}^{k-1}(t) + c(t,T), \forall t \in T,
\end{equation}
\noindent where $c(t,T)$ quantifies the relevance of proposition $t$ by taking into account its position in $T$.
We generalize the idea of reproduction probability proposed by \citeA{kintsch1978toward} by incrementally scoring propositions
based on how often they appeared in memory trees and in which part of said trees they were attached.
Then, the simulation continues to the next cycle until all sentences in $\mathcal{D}$ are consumed.
The specific behavior of subroutines \texttt{getPropositionTree, attachPropositions, memorySelect,} and \texttt{updateScore}
is instantiated by \textsc{TreeKvD} and \textsc{GraphKvD} and their details will be elaborated upon in the following parts of this section.

\begin{algorithm}[t]
\footnotesize
\caption{KvD reading simulation. Subroutines \texttt{getPropositionTree, attachPropositions, memorySelect} and \texttt{updateScore} are 
instantiated by \textsc{TreeKvD} and \textsc{GraphKvD}. }\label{alg:kvd}
\begin{algorithmic}[1]
\Require $\mathcal{D}$, source document as a list of sentences
\Require $\mathtt{WM}$, size of working memory
\Require $\mathtt{\Psi}$, maximum tree persistence
\Procedure{runSimulationKvD}{$\mathcal{D},\mathtt{WM},\mathtt{\Psi}$}

    \State $T \gets \emptyset$ \Comment{Memory tree, initially empty}
    \State $G \gets \emptyset$ \Comment{Long-term memory, initially empty}
    \State $\psi \gets 0$ \Comment{Tree persistance counter}
    \For{$s_k \in \mathcal{D}$}
        \State $P_k \gets $ \texttt{getPropositionTree}($s_k$)
        \State $T,attached \gets $ \texttt{attachPropositions}($P_k,T,G$)
        \If{$attached$}
            \State \texttt{adjustRoot}($T$)
            \State \texttt{memorySelect}($\mathtt{WM},T$)
            \State \texttt{updateScore}($T$)
            \State $\psi \gets 0$
        \Else
            \State $\psi \gets \psi + 1$
        \EndIf
        \If{ $\psi = \Psi$ }
            \State $T \gets \emptyset$
        \EndIf
    \EndFor
\EndProcedure
\end{algorithmic}
\end{algorithm}

\paragraph{Sentence Scoring.}
Once the document has been complete read, the final score of proposition $p$ is
$\text{PropScore}(p) = \text{PropScore}^{|\mathcal{D}|}(p)$.
We define the score of sentence $s_k$ as the sum of the score of all its composing propositions as
\vspace{-0.1cm}
\begin{equation}
    \label{eq:sent-score}
    \text{SentScore}(s_k) = \sum_{p \in V[P_k]} \text{PropScore}(p),
\end{equation}
\noindent where $V[P_k]$ is the set of nodes in proposition tree $P_k$ extracted from $s_k$.

\paragraph{Sentence Selection.}
We resort to a greedy selection strategy, i.e.\ selecting the top-scoring sentences according to Eq.~\refeq{eq:sent-score}
until the budget of \texttt{B} tokens is met.

\subsubsection{Proposition Building}
\label{section:prop-extr}
Propositions are obtained by recursively merging and rearranging nodes in dependency trees, extending the procedure of \citeA{fang2019proposition}.
Given sentence $s=\langle w_0,w_1,...,w_N\rangle$ and its corresponding dependency tree $Q$ with nodes $\{q_0,..,q_N\}$,\footnote{We follow Universal Dependencies \shortcite{de2021universal}, a dependency grammar formalism.}
the objective is to obtain proposition tree $P$ with nodes $\{p_0,...,p_M\}$, $M \leq N$, as follows.

First, we merge dependent nodes into head nodes in $Q$ in a bottom-up fashion.
Given $u,v \in Q$ where $u$ is head of $v$, operation $\text{merge}(u,v)$ adds all tokens contained in $v$ to node $u$
and transplants $\text{children}(v)$ --if any-- to $\text{children}(u)$.
Let $\text{dep}(u,v)$ be the grammatical relation between $u$ and $v$, dependant $v$ is merged into head $u$ if and only if
\begin{itemize}
    \setlength\itemsep{-0.2em}
    \item Node $u$ is a nominal or non-core dependant of a clausal predicate and $v$ is a function word or a discourse modifier (e.g.\ interjections or non-adverbial discourse markers).
    \item Node $u$ is any kind of dependant of a clausal predicate and $v$ is a single-token modifier.
    \item Nodes $u$ and $v$ form part of a multi-word expression or a wrongly separated token (e.g.\ $\text{dep}(u,v)=\text{goeswith}$).
\end{itemize}
Consider the example in Figure~\ref{fig:example-propbuild}.
Starting from dependency tree $Q$ (Fig.~\ref{fig:example-propbuild}a), single-token modifiers are collapsed into their head nodes (e.g.\ merge(\texttt{model},\texttt{this}) and \\
merge(\texttt{galaxy},\texttt{of})),
and compound phrases are joint (e.g.\ merge(\texttt{formation},\texttt{galaxy})).

Second, we promote coordinating conjunctions to head status as follows.
Given $u,v \in Q$, let $v$ be a node with relation $cc$ among children or grandchildren of $u$.
We transplant node $v$ to $u$'s position and put $u$ and all its children with relation \texttt{conj} as children of $v$.
In our example (Fig.~\ref{fig:example-propbuild}.b), node `\texttt{and}' is promoted and nodes `\texttt{galaxy formation}' and `\texttt{the star burst}' are transplanted as its children.
Note that at this point in the procedure, $Q$ is still a tree (Fig.~\ref{fig:example-propbuild}.c) but its nodes might now contain more than one token.

Then, for each non-leaf $u \in Q$ we build proposition $p=w_u (\text{arg}_{v_0},\text{arg}_{v_1},...)$, where $w_u$ is the sequence of tokens contained in node $u$ and $v_i \in \text{children}(u)$. We set $\text{arg}_{v_i}=w_{v_i}$ if $v$ is a leaf node, otherwise $\text{arg}_{v_i}$ is a pointer to the proposition obtained from $v_i$.
For instance, proposition \texttt{3} in Fig.~\ref{fig:example-propbuild}.d, \texttt{and(galaxy formation,\$4)}, presents proposition \texttt{4} as one of its arguments
since node `\texttt{the start burst}', from which proposition \texttt{4} is derived, is not a leaf.

Finally, edges between nodes in $Q$ are used to connect their corresponding propositions and form proposition tree $P$,
and we say that two propositions are connected if one proposition has among its arguments a pointer to the other proposition.
For instance, proposition \texttt{1} in Fig.~\ref{fig:example-propbuild}.d points to propositions \texttt{2} and \texttt{3} and hence, they are connected in $P$.

Under this procedure, the connection among propositions in the same sentence takes a syntactic nature.
However, propositions from different sentences --and hence different proposition trees-- can still be connected if the lexical overlap amongst their arguments is strong enough.
Next, we define connection through proposition overlap and how it is quantified.

\begin{figure}[t]
        \centering
        \scriptsize
\begin{tabular}{p{14.0cm}}
\toprule 
 Input: \textit{`this semi - analytical model predicts galaxy formation and the star burst of galaxies'} \\
\midrule
 \vspace{0.06cm}
{\scriptsize 

\tikzset{every picture/.style={line width=0.75pt}} 

\begin{tikzpicture}[x=0.75pt,y=0.75pt,yscale=-1,xscale=1]

\draw [line width=0.75]    (80.3,31.5) -- (67.5,17.5) ;
\draw    (98.3,54.5) -- (88.3,66.5) ;
\draw    (80.3,94.5) -- (72.3,108.5) ;
\draw    (60.3,134.5) -- (53.3,148.5) ;
\draw    (60.3,134.5) -- (30.3,148.5) ;
\draw    (80.3,94.5) -- (14.3,108.5) ;
\draw    (98.3,54.5) -- (135.3,68.7) ;
\draw    (126.3,94.5) -- (122.3,107.5) ;
\draw    (126.3,94.5) -- (206.3,106.7) ;
\draw    (215.3,135.5) -- (152.3,148.5) ;
\draw    (215.3,135.5) -- (181.3,149.5) ;
\draw    (215.3,135.5) -- (206.3,149.5) ;
\draw    (215.3,135.5) -- (257.3,147.5) ;
\draw    (261.3,172.5) -- (248.3,188.5) ;
\draw    (368.3,44.5) -- (358.3,56.5) ;
\draw    (339.3,74.5) -- (326.3,87.5) ;
\draw    (368.3,44.5) -- (411.3,59.5) ;
\draw    (441.3,73.5) -- (457.3,86.7) ;
\draw    (454.3,113.5) -- (430.3,128.5) ;
\draw    (454.3,113.5) -- (482.3,126.3) ;
\draw    (107.3,264.5) -- (97.3,276.5) ;
\draw    (78.3,294.5) -- (65.3,307.5) ;
\draw    (107.3,264.5) -- (172.3,279.3) ;
\draw    (180.3,293.5) -- (229.3,308.3) ;
\draw    (234.3,324.5) -- (248.3,337.3) ;
\draw    (180.3,293.5) -- (149.3,307.3) ;
\draw    (302.3,264.5) -- (292.3,276.5) ;
\draw    (302.3,264.5) -- (313.3,277.3) ;
\draw    (312.3,293.5) -- (312.3,307.3) ;

\draw (44,3) node [anchor=north west][inner sep=0.75pt]  [font=\scriptsize] [align=left] {$<$ROOT$>$};
\draw (80,31) node [anchor=north west][inner sep=0.75pt]  [font=\scriptsize] [align=left] {predicts};
\draw (69,71) node [anchor=north west][inner sep=0.75pt]  [font=\scriptsize] [align=left] {model};
\draw (-1,111) node [anchor=north west][inner sep=0.75pt]  [font=\scriptsize] [align=left] {this};
\draw (9,151) node [anchor=north west][inner sep=0.75pt]  [font=\scriptsize] [align=left] {semi};
\draw (45,151) node [anchor=north west][inner sep=0.75pt]  [font=\scriptsize] [align=left] {-};
\draw (47,111) node [anchor=north west][inner sep=0.75pt]  [font=\scriptsize] [align=left] {analytical};
\draw (116,71) node [anchor=north west][inner sep=0.75pt]  [font=\scriptsize] [align=left] {formation};
\draw (103,111) node [anchor=north west][inner sep=0.75pt]  [font=\scriptsize] [align=left] {galaxy};
\draw (200,111) node [anchor=north west][inner sep=0.75pt]  [font=\scriptsize] [align=left] {burst};
\draw (142,151) node [anchor=north west][inner sep=0.75pt]  [font=\scriptsize] [align=left] {and};
\draw (171.3,151) node [anchor=north west][inner sep=0.75pt]  [font=\scriptsize] [align=left] {the};
\draw (246,151) node [anchor=north west][inner sep=0.75pt]  [font=\scriptsize] [align=left] {galaxies};
\draw (239,191) node [anchor=north west][inner sep=0.75pt]  [font=\scriptsize] [align=left] {of};
\draw (80,41.5) node [anchor=north west][inner sep=0.75pt]  [font=\scriptsize] [align=left] {\textit{\textcolor[rgb]{0.82,0.01,0.11}{root}}};
\draw (69,81) node [anchor=north west][inner sep=0.75pt]  [font=\scriptsize] [align=left] {\textit{\textcolor[rgb]{0.82,0.01,0.11}{nsubj}}};
\draw (47,121) node [anchor=north west][inner sep=0.75pt]  [font=\scriptsize] [align=left] {\textit{\textcolor[rgb]{0.82,0.01,0.11}{amod}}};
\draw (45,161) node [anchor=north west][inner sep=0.75pt]  [font=\scriptsize] [align=left] {\textit{\textcolor[rgb]{0.82,0.01,0.11}{punct}}};
\draw (9,161) node [anchor=north west][inner sep=0.75pt]  [font=\scriptsize] [align=left] {\textit{\textcolor[rgb]{0.82,0.01,0.11}{amod}}};
\draw (-1,121) node [anchor=north west][inner sep=0.75pt]  [font=\scriptsize] [align=left] {\textit{\textcolor[rgb]{0.82,0.01,0.11}{det}}};
\draw (116,81) node [anchor=north west][inner sep=0.75pt]  [font=\scriptsize] [align=left] {\textit{\textcolor[rgb]{0.82,0.01,0.11}{obj}}};
\draw (103,121) node [anchor=north west][inner sep=0.75pt]  [font=\scriptsize] [align=left] {\textit{\textcolor[rgb]{0.82,0.01,0.11}{compound}}};
\draw (194,152) node [anchor=north west][inner sep=0.75pt]  [font=\scriptsize] [align=left] {star};
\draw (200,121) node [anchor=north west][inner sep=0.75pt]  [font=\scriptsize] [align=left] {\textit{\textcolor[rgb]{0.82,0.01,0.11}{conj}}};
\draw (142,164) node [anchor=north west][inner sep=0.75pt]  [font=\scriptsize] [align=left] {\textit{\textcolor[rgb]{0.82,0.01,0.11}{cc}}};
\draw (171,161) node [anchor=north west][inner sep=0.75pt]  [font=\scriptsize] [align=left] {\textit{\textcolor[rgb]{0.82,0.01,0.11}{det}}};
\draw (194,161) node [anchor=north west][inner sep=0.75pt]  [font=\scriptsize] [align=left] {\textit{\textcolor[rgb]{0.82,0.01,0.11}{compound}}};
\draw (246,161) node [anchor=north west][inner sep=0.75pt]  [font=\scriptsize] [align=left] {\textit{\textcolor[rgb]{0.82,0.01,0.11}{nmod}}};
\draw (239,201) node [anchor=north west][inner sep=0.75pt]  [font=\scriptsize] [align=left] {\textit{\textcolor[rgb]{0.82,0.01,0.11}{case}}};
\draw (350,31) node [anchor=north west][inner sep=0.75pt]  [font=\scriptsize] [align=left] {predicts};
\draw (321,61) node [anchor=north west][inner sep=0.75pt]  [font=\scriptsize] [align=left] {this model};
\draw (288,91) node [anchor=north west][inner sep=0.75pt]  [font=\scriptsize] [align=left] {semi - analytical};
\draw (408,61) node [anchor=north west][inner sep=0.75pt]  [font=\scriptsize] [align=left] {galaxy formation};
\draw (434,91) node [anchor=north west][inner sep=0.75pt]  [font=\scriptsize] [align=left] {the star burst};
\draw (420,131) node [anchor=north west][inner sep=0.75pt]  [font=\scriptsize] [align=left] {and};
\draw (461,131) node [anchor=north west][inner sep=0.75pt]  [font=\scriptsize] [align=left] {of galaxies};
\draw (434,101) node [anchor=north west][inner sep=0.75pt]  [font=\scriptsize] [align=left] {\textit{\textcolor[rgb]{0.82,0.01,0.11}{conj}}};
\draw (420,141) node [anchor=north west][inner sep=0.75pt]  [font=\scriptsize] [align=left] {\textit{\textcolor[rgb]{0.82,0.01,0.11}{cc}}};
\draw (89,251) node [anchor=north west][inner sep=0.75pt]  [font=\scriptsize] [align=left] {predicts};
\draw (60,281) node [anchor=north west][inner sep=0.75pt]  [font=\scriptsize] [align=left] {this model};
\draw (22,311) node [anchor=north west][inner sep=0.75pt]  [font=\scriptsize] [align=left] {semi - analytical};
\draw (114,311) node [anchor=north west][inner sep=0.75pt]  [font=\scriptsize] [align=left] {galaxy formation};
\draw (206,311) node [anchor=north west][inner sep=0.75pt]  [font=\scriptsize] [align=left] {the star burst};
\draw (166,281) node [anchor=north west][inner sep=0.75pt]  [font=\scriptsize] [align=left] {and};
\draw (228,338) node [anchor=north west][inner sep=0.75pt]  [font=\scriptsize] [align=left] {of galaxies};
\draw (299,251) node [anchor=north west][inner sep=0.75pt]  [font=\scriptsize] [align=left] {1};
\draw (289,281) node [anchor=north west][inner sep=0.75pt]  [font=\scriptsize] [align=left] {2};
\draw (309,311) node [anchor=north west][inner sep=0.75pt]  [font=\scriptsize] [align=left] {4};
\draw (309,281) node [anchor=north west][inner sep=0.75pt]  [font=\scriptsize] [align=left] {3};
\draw (334,250) node [anchor=north west][inner sep=0.75pt]  [font=\scriptsize] [align=left] {\texttt{1: predicts(\$2, \$3)}\\\texttt{2: this model(semi - analytical)}\\\texttt{3: and(galaxy formation, \$4)}\\\texttt{4: the star burst(of galaxies)}};
\draw (128,211) node [anchor=north west][inner sep=0.75pt]  [font=\scriptsize] [align=left] {{ (a)}};
\draw (376,211) node [anchor=north west][inner sep=0.75pt]  [font=\scriptsize] [align=left] {{ (b)}};
\draw (129,350) node [anchor=north west][inner sep=0.75pt]  [font=\scriptsize] [align=left] {{ (c)}};
\draw (376,350) node [anchor=north west][inner sep=0.75pt]  [font=\scriptsize] [align=left] {{ (d)}};

\end{tikzpicture}
} \\
 \bottomrule
\end{tabular}
\caption{Step-by-step construction of proposition tree from an input sentence, starting from obtaining its dependency tree in UD format (a),
merging dependent nodes into head nodes (b), promoting coordinating conjunctions to head status (c), to finally build propositions from non-leaf nodes (d).
}
\label{fig:example-propbuild}
\end{figure}

\paragraph{Proposition Overlap.}
We connect propositions from different sentences by quantifying the lexical overlap between their functors --predicates and arguments.
Let $\text{functors}(p)$ be the set of the functors --predicate and arguments-- in propositions $p$.
Given $p_1\in P_x$ and $p_2 \in P_y$,
let $A^*(p_1,p_2)$ be the optimal alignment between $\text{functors}(p_1)$ and $\text{functors}(p_2)$.
Alignment $A^*$ is defined as the maximum matching that can be obtained greedily in the weighted bipartite graph formed from sets $\text{functors}(p_1)$ and $\text{functors}(p_2)$. The edge weight between two functors is defined as $e(a,b)=\text{jaccard}(L_a,L_b)$,
the Jaccard similarity between their sets of lemmas after discarding stopwords, punctuation, and adjectives --$L_a$ and $L_b$.
Then, the average overlap score between $p_1$ and $p_2$, $\phi(p_1,p_2)$, is defined as
\vspace{-0.3cm}
\begin{equation}
\label{eq:prop-overlap}
    \phi(p_1,p_2) = \frac{1}{|A^*|} \sum_{\langle a_1,a_2 \rangle \in A^*} \text{jaccard}(a_1,a_2).
\end{equation}
\vspace{0.01cm}
This overlap score function becomes useful when searching an appropriate place to attach incoming propositions to the current memory tree or 
to pull propositions from long-term memory. We elaborate more on this in the next section.


\subsubsection{TreeKvD}
\label{sec:treekvd}
In this part, we introduce \textsc{TreeKvD}, the first sentence scoring system simulating KvD reading.
The system models working memory and long-term memory as two separate weighted undirected graphs
where each node represents a proposition and an edge connecting two propositions indicates the existence of overlap between their arguments,
with the edge weight quantifying this overlap.
Furthermore, working memory is constrained to be a tree, whereas long-term memory is modeled as a forest of trees pruned from
memory trees during simulation.
Let $s_k$ be the sentence read in cycle $k$, $T$ the working memory tree at the beginning of the cycle, with node set $V[T]$ and edge set $E[T]$.
Similarly, let $G$ be the long-term memory graph with $V[G]$ and $E[G]$ as node and edge set, respectively.
We now elaborate on the details of each step of the \textsc{TreeKvD}'s implementation of Algorithm~\ref{alg:kvd}.

\paragraph{Extracting and Attaching Incoming Nodes.}
First, subroutine \texttt{getPropositionTree} (Line 6) receives $s_k$ as input (as a sequence of tokens) and returns its corresponding proposition tree $P_k$
following the procedure presented in section~\ref{section:prop-extr}.

Then, subroutine \texttt{attachPropositions} (Line 7) attempts to attach $P_k$ to $T$, receiving as input structures $P_k$, $T$, and $G$, and returning the updated tree $T$ along with flag \texttt{attached} to indicate whether $T$ was modified or not.
The attachment of $P_k$ to $T$ and proceeds as follows.
We define the optimal place to attach $P_k$ to $T$ as the pair $(t^*,p^*)$  where $t^* \in V[T], p^* \in V[P_k]$ such that
\begin{equation}
    (t^*,p^*) = \argmax_{t \in V[T], p\in V[P_k]} \phi(t,p),
\end{equation}
\noindent where $\phi(\cdot)$ is the proposition overlap function defined in Equation~\ref{eq:prop-overlap}.
In case no attachment pair can be found, i.e.\ $\phi(t,p)=0, \forall t \in V[T] \wedge \forall p \in V[P_k]$, 
\texttt{attachPropositions} resorts to two cascaded backup plans.

As first backup attachment plan, the procedure \textit{recalls} a path of forgotten propositions from long-term memory $G$ to serve as bridge to connect $P_k$ and $T$.
Let $\mathcal{F}(R)$ be the set of all paths of length at most $R$ in $G$,
we define the optimal attachment place aided by $\textbf{f} \in \mathcal{F}$ as the tuple $(t^*,\textbf{f}^*,p^*)$, such that
\vspace{-0.5cm}
\begin{equation*}
    (t^*,\textbf{f}^*,p^*) = \argmax_{t \in V[T], p\in V[P_k], \textbf{f} \in \mathcal{F}(R)}  \phi(t,f_1) +
    \sum_{i=2}^{n} \phi(f_{i-1},f_i) +
    \phi(f_{n},p),
\end{equation*}
where $\textbf{f}=\langle f_1,...,f_{n} \rangle, f_i \in V[G] \land n\leq R$.
In this way, $P_k$ is attached to $T$ by retrieving a path $\textbf{f}^*$ from $G$ with at most $R$ forgotten nodes that maximizes argument overlap between placement candidates $t^*$ and $p^*$.

In case no suitable recall path can be found (total overlap score is still zero), procedure \texttt{attachPropositions} 
resorts to a second backup attachment strategy, which consists of deciding whether to keep $T$ as memory tree during the current cycle or whether to replace it completely with $P_k$.
Among both trees, we keep the one whose root node presents the highest closeness centrality.
The closeness centrality of a node in an undirected graph is defined as the inverse of the sum of all shortest paths from said node to all other nodes in the graph.
As we will discuss in the root adjustment section, a root closer to all other nodes is an indication of a well balanced tree and allows for efficient pruning, hence a desirable property.
In case $T$ is not replaced, the procedure returns flag \texttt{attached} as $\texttt{False}$.


Now consider the case when \texttt{attachPropositions} fails to attach propositions to $T$ for more than one consecutive cycle.
We name this phenomenon \textit{tree persistence}.
A highly persistent tree is undesirable since it can potentially block important connections between more recently read propositions.
In order to avoid this scenario, we reset the memory tree (line 16 in Algorithm~\ref{alg:kvd}) if its persistence reaches the maximum permissible value, $\Psi$.
Furthermore, we avoid over-scoring nodes in persistent trees by only updating their score if any form of attachment took place (Line 8).


\paragraph{Choosing and Adjusting the Root.}
After attachment takes place, subroutine \texttt{adjustRoot} will select the most appropriate node in the updated $T$ as the root (Line 9).
An important property of working memory trees in the KvD theory is that the root conveys the most central topic at the time of reading.
We build upon \citeA{fang2019proposition} criteria and model this property by selecting the node 
that presents the highest closeness centrality as the root. 
Such a root would facilitate reaching all nodes in the least amount of steps --in average--, a desired property during pruning.

\paragraph{Pruning Working Memory.}
Next, subroutine \texttt{memorySelect} (Line 10) receives as input memory capacity parameter \texttt{WM} and memory tree $T$,
and proceeds to select at most \texttt{WM} nodes from $T$ in the following manner.
Starting from the root, $T$ is traversed in topological order until reaching a leaf node, selecting each node visited along the way.
At this point, if the amount of select nodes is less than \texttt{WM}, nodes are selected in breadth-first traversing order (starting from the root) until capacity is reached or until all nodes are traversed.
Finally, nodes not selected are pruned from $T$ and moved to $G$.

\paragraph{Proposition Scoring.}
Following Eq.~\refeq{eq:pscore}, reproduced here for convenience, the score of propositions is updated as
\begin{equation*}
    \text{PropScore}^k(t) = \text{PropScore}^{k-1}(t) + c(t,T), \forall t \in T,
\end{equation*}
in which subroutine \texttt{updateScore} (Line 11) defines the updating term $c(\cdot)$ as
\vspace{-0.2cm}
\begin{equation}
\label{eq:tkvd-ct}
    c(t,T) = \frac{|T_t|}{|T|} \exp \left( \frac{1}{\text{depth}(t)} \right),
\end{equation}
\noindent where $\text{depth}(t)$ is the depth of node $t$ with respect to the root and $|T_t|$ is the size of the subtree rooted in $t$.
In this way, nodes closer to the root as well as nodes holding more information in their subtree are scored higher.




\paragraph{Limitations.}
\label{section:tkvd-lim}
The presented system closely follows mechanisms of memory organization theorized by \shortciteA{kintsch1978toward}.
As such, the system presents a number of processing limitations inherent to the KvD theory itself which we now elaborate on.

First, the constrained amount of content units in working memory at any given time poses a limitation to how 
much information the system has access to when updating the score of memory tree nodes.
It is entirely possible that some propositions are pruned away and never recalled again, in which case their score will be zero.

Second, \citeA{kintsch1978toward} define the recall mechanism as a routine capable of pulling an unlimited number of 
propositions from long-term memory. Additionally, propositions might not be recalled \textit{verbatim} but simplified, 
given that the difficulty to recall specific details increases over time \cite{postman1965short}.
In system \textsc{TreeKvD}, we limit ourselves to recall previously read propositions verbatim and further limit the maximum number of propositions to recall.
This design choice limits the possibility of recalling important propositions back into working memory.

Third, attachment of an incoming proposition tree to the current memory tree is done by connecting one node in the memory tree to one node in the incoming tree. Whilst this strategy guarantees that the resulting structure remains a tree, as KvD requires, 
many potentially useful connections are ignored.
We address these limitations in the design of the next system.

    



\subsubsection{GraphKvD}
The second proposed system, \textsc{GraphKvD}, considers instead a single underlying structure for long-term memory and short-term memory.
Working memory is modeled as a subgraph of long-term memory that preserves properties of KvD micro-structure, i.e.\ a tree with constrained size.
Such modeling of memory modules allows for richer connections between incoming proposition trees and working memory, in addition to giving the system efficient access to nodes neighboring memory tree nodes, significantly increasing the coverage of content during scoring.
We now proceed to elaborate on how \textsc{GraphKvD} instantiates Algorithm~\ref{alg:kvd}.


\paragraph{Extracting and Attaching Incoming Nodes.}
In the same fashion as in \textsc{TreeKvD}, procedure \texttt{getPropositionTree} extracts $P_k$ from incoming sentence $s_k$ (line 6).
Then, procedure \texttt{attachPropositions} will first attempt to connect $P_k$ to $T$ directly, falling back to two cascaded strategies if unsuccessful.

In contrast with \textsc{TreeKvD}, all nodes in $P_k$ are allowed to connect to $T$. Hence, for each 
$p \in V[P_k]$, its optimal place to be attached to $T$ is defined as the pair $(p,t)$ such that 
$t = \argmax_{\hat{t} \in V[T]} \phi(\hat{t},p)$, where $\phi(\cdot)$ is again the proposition overlap function defined in Equation~\ref{eq:prop-overlap}.
In case no node in $P_k$ could be connected to any node in $T$, \texttt{attachPropositions} employs again two backup plans. Note that these plans are not triggered if at least one node in $P_k$ was connected to $T$.

The first plan consists of a recall mechanism that retrieves paths from $G$ connecting each node in $P_k$ to each node in $T$.
For each node $p \in V[P_k]$, its the optimal attachment place $t^* \in V[T]$ aided by path $\textbf{f}^*=\langle f_1,...,f_n \rangle$, $f_i \in V[G] \land n \leq R$, is defined as
\vspace{-0.25cm}
\begin{equation}
    (t^*,\textbf{f}^*) = \argmax_{t \in V[T], \textbf{f} \subset G} \phi(f_1,t) + \nonumber c(t,T) \left( \sum_{i=2}^{|\textbf{f}|} \phi(f_{i-1},\hat{f}_i) \right) \exp(-|\textbf{f}|) + \phi(f_n,p).
\end{equation}
Note that \textsc{GraphKvD} defines the optimal attachment place differently from \textsc{TreeKvD} in two respects.
First, \textsc{GraphKvD} explicitly favours the attachment of recall paths to highly relevant nodes in $T$, i.e. high $c(\cdot)$ value.
This encourages the memory tree to expand on information about relevant content rather than non-relevant ones.
Second, \textsc{GraphKvD} includes an exponential decay length penalty ($\exp(-|\textbf{f}|)$) 
to favour the retrieval of shorter recall paths.
This penalty is inspired by recent research on how content is gradually forgotten (`decays') in human memory and becomes harder to retrieve \cite{berman2009search}, an idea also applied in the optimization of neural networks \cite{loshchilov2018decoupled}.
In this way, we avoid populating $T$ with long proposition chains that may contain only marginally relevant and potentially redundant information.
Moreover, this approach aims to save memory capacity for other potentially informative attachments.

As second backup plan, procedure \texttt{attachPropositions} will replace $T$ with $P_k$ if $|V[P_k]|>|V[T]|$ and 
the closeness centrality of the root of $P_k$ is greater than that of the root of $T$.
$T$ will also be replaced if the tree persistence has reached its allowed limit, $\psi=\Psi$.
In case $P_k$ is chosen, we \textit{enrich} it by retrieving single nodes from $G$ and connecting them to $P$,
in a similar fashion to the \textit{construction} stage in the Construction-Integration theory of comprehension \cite{kintsch1988role}.
For each node $p \in V[P_k]$, we retrieve candidate nodes in the following order.
First, nodes from the local context, i.e.\ from the current paragraph or article section, are retrieved.
Then, nodes are retrieved in inverse order of processing recency, i.e.\ propositions from sentences processed at the beginning of the simulation are retrieved first.
For each node, searching stops when the argument overlap score of a candidate is greater than zero.\footnote{Experimentally, increasing this threshold does not impact downstream performance significantly.}

This particular retrieval order follows \textit{free recall} accuracy in human subjects \cite{glanzer1972storage}.\footnote{Free recall is a technique used in psycholingusitic studies of human memory in which a subject is presented with a string of items and is free to recall them in any order; in contrast, \textit{serial recall} requires the subject to recall the items in order.}
The tendency to accurately recall the first processed items is known as the \textit{priming effect} \cite{Harley1995}, and is said to depend on long-term memory.
Instead, the tendency to accurately recall the most recent items is called the \textit{recency effect}, and it depends on short-term memory.

\paragraph{Updating Memory Structures.}
After attachment, long-term memory graph $G$ is updated with nodes and edges in $T$.
Note that after executing the attachment procedures described above, the updated memory graph $T$ might no longer be a tree.
However, as mentioned before, the KvD theory models that a valid working memory structure as a tree.
Hence, we reduce $T$ to its maximum spanning tree using the argument overlap score between propositions as edge weights.
Similarly to \textsc{TreeKvD}, the node with the maximum closeness score is chosen as the new root.
Then, $T$ is pruned down to have at most \texttt{WM} nodes using the same strategy as in \S~\ref{sec:treekvd}.

\paragraph{Proposition Scoring.}
The score of nodes in working memory $T$ is updated according to Eq.~\refeq{eq:pscore} and Eq.\refeq{eq:tkvd-ct}.
However, \textsc{GraphKvD} will also update the score of nodes neighboring those in $T$.
In this way, propositions that contribute to the understanding of nodes in $T$ are reinforced, and the more a proposition is selected the more its connections are updated.
For each node $t \in V[T]$,
we define $N(t)=\{ u; u \in V[G] \backslash V[T], \text{ s.t. } (u,v) \in E[G] \}$, the set of nodes neighboring $t$ located in $G$.
Then, the updated score of neighbor node $u$ is
\vspace{-0.2cm}
\begin{equation}
    \text{PropScore}^k(u) = \text{PropScore}^{k-1}(u) + \beta \cdot c(t,T), \forall u \in N(t),
\end{equation}
\noindent where $\beta<1$ is a decay factor.
The consideration of neighboring nodes and a decayed scoring strategy follows the \textit{integration} and \textit{spreading} processing proposed in the Construction-Integration theory.
The objective is to integrate peripheral or related concepts into the memory cycle and spread minimal attentional resources to them in the form of score value,
where parameter $\beta$ controls how much attention is leaked.


    

\subsubsection{Simulation Example}
Next, we illustrate the procedures outlined in Algorithm~\ref{alg:kvd} with an example, showcased in Figure~\ref{fig:ex-rec}.
The example takes two sentences from a scientific article and simulates two memory cycles with \textsc{TreeKvD} (left) and \textsc{GraphKvD} (right).
The propositions involved (middle row) in the cycles are presented alongside the corresponding gold summary (bottom row).
Propositions not directly mentioned in the simulation but necessary for content interpretation are shown in italics.
First, we analyze the processes involved during attachment in a memory cycle, including how the recall mechanism operates.
Then, we relate the properties a memory tree should exhibit according to the KvD theory, and the properties of memory trees obtained with \textsc{TreeKvD} and \textsc{GraphKvD}.

\begin{figure}[t]
\centering
\scriptsize 

\begin{tabular}{p{7cm}p{7.5cm}}
\toprule 
 \multicolumn{2}{p{14.5cm}}{\textbf{Cycle k : `}Therefore, we obtain equal-time and time-dependent structure functions for a shell model for 3D MHD turbulence and, from these, equal-time and dynamic multiscaling exponents. '} \\
\multicolumn{2}{p{14.5cm}}{
\tikzset{every picture/.style={line width=0.75pt}} 
\begin{tikzpicture}[x=0.75pt,y=0.75pt,yscale=-1,xscale=1]

\draw  [dash pattern={on 0.84pt off 2.51pt}]  (53,57) -- (67.5,57) ;
\draw  [dash pattern={on 0.84pt off 2.51pt}]  (84,57) -- (99.5,57) ;
\draw  [dash pattern={on 0.84pt off 2.51pt}]  (113.5,57) -- (129.5,57) ;
\draw  [dash pattern={on 0.84pt off 2.51pt}]  (53,63) -- (67.5,76.07) ;
\draw [color={rgb, 255:red, 208; green, 2; blue, 27 }  ,draw opacity=1 ] [dash pattern={on 0.84pt off 2.51pt}]  (144.5,57.07) -- (158.5,57.07) ;
\draw  [dash pattern={on 0.84pt off 2.51pt}]  (172,57) -- (186.5,57) ;
\draw  [dash pattern={on 0.84pt off 2.51pt}]  (203,57) -- (217.5,57) ;
\draw    (203,77) -- (217.5,77) ;
\draw    (234,77) -- (239.5,77) ;
\draw    (253,77) -- (258.5,77) ;
\draw    (171,62) -- (187.5,76.07) ;
\draw    (94,116) -- (99.5,116) ;
\draw    (113,116) -- (118.5,116) ;
\draw    (74,116) -- (79.5,116) ;
\draw    (54,116) -- (59.5,116) ;
\draw  [dash pattern={on 0.84pt off 2.51pt}]  (349,56) -- (363.5,56) ;
\draw  [dash pattern={on 0.84pt off 2.51pt}]  (380,56) -- (395.5,56) ;
\draw  [dash pattern={on 0.84pt off 2.51pt}]  (409.5,56) -- (425.5,56) ;
\draw  [dash pattern={on 0.84pt off 2.51pt}]  (349,62) -- (363.5,75.07) ;
\draw [color={rgb, 255:red, 208; green, 2; blue, 27 }  ,draw opacity=1 ] [dash pattern={on 0.84pt off 2.51pt}]  (440.5,56.07) -- (454.5,56.07) ;
\draw  [dash pattern={on 0.84pt off 2.51pt}]  (468,56) -- (482.5,56) ;
\draw  [dash pattern={on 0.84pt off 2.51pt}]  (499,56) -- (513.5,56) ;
\draw    (499,76) -- (513.5,76) ;
\draw    (530,76) -- (535.5,76) ;
\draw    (549,76) -- (554.5,76) ;
\draw    (467,61) -- (483.5,75.07) ;
\draw [color={rgb, 255:red, 208; green, 2; blue, 27 }  ,draw opacity=1 ] [dash pattern={on 0.84pt off 2.51pt}]  (433,63) -- (444.5,76.07) ;
\draw [color={rgb, 255:red, 208; green, 2; blue, 27 }  ,draw opacity=1 ] [dash pattern={on 0.84pt off 2.51pt}]  (444.5,76.07) -- (478.5,76.07) ;
\draw [color={rgb, 255:red, 208; green, 2; blue, 27 }  ,draw opacity=1 ] [dash pattern={on 0.84pt off 2.51pt}]  (349,62) -- (359.5,89.07) ;
\draw [color={rgb, 255:red, 208; green, 2; blue, 27 }  ,draw opacity=1 ] [dash pattern={on 0.84pt off 2.51pt}]  (359.5,89.07) -- (542.5,89.07) ;
\draw [color={rgb, 255:red, 208; green, 2; blue, 27 }  ,draw opacity=1 ] [dash pattern={on 0.84pt off 2.51pt}]  (542.5,89.07) -- (542.5,83.07) ;
\draw [color={rgb, 255:red, 208; green, 2; blue, 27 }  ,draw opacity=1 ] [dash pattern={on 0.84pt off 2.51pt}]  (349,62) -- (355.5,94.07) ;
\draw [color={rgb, 255:red, 208; green, 2; blue, 27 }  ,draw opacity=1 ] [dash pattern={on 0.84pt off 2.51pt}]  (355.5,94.07) -- (561.5,94.07) ;
\draw [color={rgb, 255:red, 208; green, 2; blue, 27 }  ,draw opacity=1 ] [dash pattern={on 0.84pt off 2.51pt}]  (561.5,94.07) -- (561.5,81.07) ;
\draw [color={rgb, 255:red, 155; green, 155; blue, 155 }  ,draw opacity=1 ]   (293,10) -- (293,128.07) ;
\draw    (392,117) -- (397.5,117) ;
\draw    (411,117) -- (416.5,117) ;
\draw    (372,117) -- (377.5,117) ;
\draw    (352,117) -- (357.5,117) ;

\draw (40,52) node [anchor=north west][inner sep=0.75pt]   [align=left] {{{71}}};
\draw (70,72) node [anchor=north west][inner sep=0.75pt]   [align=left] {{{80}}};
\draw (70,52) node [anchor=north west][inner sep=0.75pt]   [align=left] {{{77}}};
\draw (100,52) node [anchor=north west][inner sep=0.75pt]   [align=left] {{{78}}};
\draw (129,52) node [anchor=north west][inner sep=0.75pt]   [align=left] {{{79}}};
\draw (159,52) node [anchor=north west][inner sep=0.75pt]   [align=left] {{{\underline{81}}}};
\draw (189,52) node [anchor=north west][inner sep=0.75pt]   [align=left] {{{82}}};
\draw (219,52) node [anchor=north west][inner sep=0.75pt]   [align=left] {{{83}}};
\draw (189,72) node [anchor=north west][inner sep=0.75pt]   [align=left] {{{84}}};
\draw (220,72) node [anchor=north west][inner sep=0.75pt]   [align=left] {{{86}}};
\draw (240,72) node [anchor=north west][inner sep=0.75pt]   [align=left] {{{85}}};
\draw (260,72) node [anchor=north west][inner sep=0.75pt]   [align=left] {{{87}}};
\draw (40,110) node [anchor=north west][inner sep=0.75pt]   [align=left] {{{\underline{81}}}};
\draw (61,110) node [anchor=north west][inner sep=0.75pt]   [align=left] {{{84}}};
\draw (80,110) node [anchor=north west][inner sep=0.75pt]   [align=left] {{{86}}};
\draw (100,110) node [anchor=north west][inner sep=0.75pt]   [align=left] {{{85}}};
\draw (120,110) node [anchor=north west][inner sep=0.75pt]   [align=left] {{{87}}};
\draw (336,52) node [anchor=north west][inner sep=0.75pt]   [align=left] {{{71}}};
\draw (366,72) node [anchor=north west][inner sep=0.75pt]   [align=left] {{{80}}};
\draw (366,52) node [anchor=north west][inner sep=0.75pt]   [align=left] {{{77}}};
\draw (396,52) node [anchor=north west][inner sep=0.75pt]   [align=left] {{{78}}};
\draw (425,52) node [anchor=north west][inner sep=0.75pt]   [align=left] {{{79}}};
\draw (455,52) node [anchor=north west][inner sep=0.75pt]   [align=left] {{{\underline{81}}}};
\draw (485,52) node [anchor=north west][inner sep=0.75pt]   [align=left] {{{82}}};
\draw (515,52) node [anchor=north west][inner sep=0.75pt]   [align=left] {{{83}}};
\draw (485,72) node [anchor=north west][inner sep=0.75pt]   [align=left] {{{84}}};
\draw (516,72) node [anchor=north west][inner sep=0.75pt]   [align=left] {{{86}}};
\draw (536,72) node [anchor=north west][inner sep=0.75pt]   [align=left] {{{85}}};
\draw (556,72) node [anchor=north west][inner sep=0.75pt]   [align=left] {{{87}}};
\draw (10,27) node [anchor=north west][inner sep=0.75pt]   [align=left] {{$T'=\text{aP}( \ \ \ \ \ \ T_{k-1}\ \ \ \ \ \ \ \ \ \ \ ,\ \ \ \ \ \ \ P\ \ \ \ \ \ \ \ \ \ \ ,F)$}};
\draw (10,110) node [anchor=north west][inner sep=0.75pt]   [align=left] {{{$T_k=$}}};
\draw (123,-3) node [anchor=north west][inner sep=0.75pt]   [align=left] {{{TreeKvD}}};
\draw (426,-4) node [anchor=north west][inner sep=0.75pt]   [align=left] {{{GraphKvD}}};
\draw (338,110) node [anchor=north west][inner sep=0.75pt]   [align=left] {{{\underline{81}}}};
\draw (359,110) node [anchor=north west][inner sep=0.75pt]   [align=left] {{{84}}};
\draw (378,110) node [anchor=north west][inner sep=0.75pt]   [align=left] {{{86}}};
\draw (398,110) node [anchor=north west][inner sep=0.75pt]   [align=left] {{{85}}};
\draw (418,110) node [anchor=north west][inner sep=0.75pt]   [align=left] {{{87}}};
\draw (309,110) node [anchor=north west][inner sep=0.75pt]   [align=left] {$T_k=$};
\draw (310,27) node [anchor=north west][inner sep=0.75pt]   [align=left] {{$G'=\text{aP}( \ \ \ \ \ \ T_{k-1}\ \ \ \ \ \ \ \ \ \ \ ,\ \ \ \ \ \ \ P\ \ \ \ \ \ ,G)$}};
\end{tikzpicture}
} \\
\multicolumn{2}{p{14.5cm}}{\textbf{Cycle k+1: `}We then try to see if these suggest any bridge relations. '} \\
\multicolumn{2}{p{14.5cm}}{
\tikzset{every picture/.style={line width=0.75pt}} 
\begin{tikzpicture}[x=0.75pt,y=0.75pt,yscale=-1,xscale=1]

\draw    (23,57) -- (37.5,57) ;
\draw    (54,57) -- (69.5,57) ;
\draw    (83.5,57) -- (99.5,57) ;
\draw    (101,117) -- (106.5,117) ;
\draw    (120,117) -- (125.5,117) ;
\draw    (81,117) -- (86.5,117) ;
\draw    (61,117) -- (66.5,117) ;
\draw [color={rgb, 255:red, 155; green, 155; blue, 155 }  ,draw opacity=1 ]   (293,10) -- (293,128.07) ;
\draw    (392,117) -- (397.5,117) ;
\draw    (411,117) -- (416.5,117) ;
\draw    (372,117) -- (377.5,117) ;
\draw    (352,117) -- (357.5,117) ;
\draw  [dash pattern={on 0.84pt off 2.51pt}]  (111.5,57) -- (127.5,57) ;
\draw    (23,62) -- (43,77) ;
\draw    (43,77) -- (66.51,77) ;
\draw  [dash pattern={on 0.84pt off 2.51pt}]  (83,77) -- (98.5,77) ;
\draw  [dash pattern={on 0.84pt off 2.51pt}]  (113,77) -- (128.5,77) ;
\draw [color={rgb, 255:red, 208; green, 2; blue, 27 }  ,draw opacity=1 ] [dash pattern={on 0.84pt off 2.51pt}]  (142.51,77.07) -- (167.91,57.7) ;
\draw   (69,70) -- (83.51,70) -- (83.51,85.1) -- (69,85.1) -- cycle ;
\draw   (99,70) -- (113.51,70) -- (113.51,85.1) -- (99,85.1) -- cycle ;
\draw   (128,70) -- (142.51,70) -- (142.51,85.1) -- (128,85.1) -- cycle ;
\draw  [dash pattern={on 0.84pt off 2.51pt}]  (183.5,57) -- (199.5,57) ;
\draw  [dash pattern={on 0.84pt off 2.51pt}]  (213.5,57) -- (229.5,57) ;
\draw    (333,57) -- (347.5,57) ;
\draw    (364,57) -- (379.5,57) ;
\draw    (393.5,57) -- (409.5,57) ;
\draw    (421.5,57) -- (437.5,57) ;
\draw  [dash pattern={on 0.84pt off 2.51pt}]  (333,62) -- (353,77) ;
\draw  [dash pattern={on 0.84pt off 2.51pt}]  (353,77) -- (376.51,77) ;
\draw   (379,70) -- (393.51,70) -- (393.51,85.1) -- (379,85.1) -- cycle ;
\draw  [dash pattern={on 0.84pt off 2.51pt}]  (493.5,57) -- (509.5,57) ;
\draw  [dash pattern={on 0.84pt off 2.51pt}]  (523.5,57) -- (539.5,57) ;
\draw [color={rgb, 255:red, 208; green, 2; blue, 27 }  ,draw opacity=1 ] [dash pattern={on 0.84pt off 2.51pt}]  (394.91,77) -- (524.91,77) ;
\draw [color={rgb, 255:red, 208; green, 2; blue, 27 }  ,draw opacity=1 ] [dash pattern={on 0.84pt off 2.51pt}]  (524.91,77) -- (545.91,63.7) ;

\draw (10,52) node [anchor=north west][inner sep=0.75pt]   [align=left] {{{81}}};
\draw (40,52) node [anchor=north west][inner sep=0.75pt]   [align=left] {{{84}}};
\draw (70,52) node [anchor=north west][inner sep=0.75pt]   [align=left] {{{86}}};
\draw (99,52) node [anchor=north west][inner sep=0.75pt]   [align=left] {{{85}}};
\draw (129,52) node [anchor=north west][inner sep=0.75pt]   [align=left] {{{87}}};
\draw (170,52) node [anchor=north west][inner sep=0.75pt]   [align=left] {{{88}}};
\draw (200,52) node [anchor=north west][inner sep=0.75pt]   [align=left] {{{89}}};
\draw (230,52) node [anchor=north west][inner sep=0.75pt]   [align=left] {{{90}}};
\draw (47,110) node [anchor=north west][inner sep=0.75pt]   [align=left] {{{\underline{25}}}};
\draw (68,110) node [anchor=north west][inner sep=0.75pt]   [align=left] {{{81}}};
\draw (87,110) node [anchor=north west][inner sep=0.75pt]   [align=left] {{{84}}};
\draw (107,110) node [anchor=north west][inner sep=0.75pt]   [align=left] {{{86}}};
\draw (127,110) node [anchor=north west][inner sep=0.75pt]   [align=left] {{{85}}};
\draw (5,27) node [anchor=north west][inner sep=0.75pt]   [align=left] {$T'=\text{aP}( \ \ \ \ \ \ T_{k}\ \ \ \ \ \ \ \ \ \ \ \ ,\ \ \ \ \ \ \ \ \ \ \ P\ \ \ \ \ \ \ \ \ \ \ ,F)$};
\draw (123,-3) node [anchor=north west][inner sep=0.75pt]   [align=left] {{{TreeKvD}}};
\draw (426,-4) node [anchor=north west][inner sep=0.75pt]   [align=left] {{{GraphKvD}}};
\draw (338,110) node [anchor=north west][inner sep=0.75pt]   [align=left] {{{\underline{81}}}};
\draw (359,110) node [anchor=north west][inner sep=0.75pt]   [align=left] {{{84}}};
\draw (378,110) node [anchor=north west][inner sep=0.75pt]   [align=left] {{{86}}};
\draw (398,110) node [anchor=north west][inner sep=0.75pt]   [align=left] {{{85}}};
\draw (418,110) node [anchor=north west][inner sep=0.75pt]   [align=left] {{{87}}};
\draw (302,27) node [anchor=north west][inner sep=0.75pt]   [align=left] {{$G'=\text{aP}( \ \ \ \ \ \ \ T_{k}\ \ \ \ \ \ \ \ \ \ \ \ \ \ ,\ \ \ \ \ \ \ \ \ \ P\ \ \ \ \ \ ,G)$}};
\draw (70,72) node [anchor=north west][inner sep=0.75pt]   [align=left] {{{\underline{25}}}};
\draw (100,72) node [anchor=north west][inner sep=0.75pt]   [align=left] {{{24}}};
\draw (129,72) node [anchor=north west][inner sep=0.75pt]   [align=left] {{{21}}};
\draw (8,110) node [anchor=north west][inner sep=0.75pt]   [align=left] {{{$T_{k+1}=$}}};
\draw (320,52) node [anchor=north west][inner sep=0.75pt]   [align=left] {{{\underline{81}}}};
\draw (350,52) node [anchor=north west][inner sep=0.75pt]   [align=left] {{{84}}};
\draw (380,52) node [anchor=north west][inner sep=0.75pt]   [align=left] {{{86}}};
\draw (409,52) node [anchor=north west][inner sep=0.75pt]   [align=left] {{{85}}};
\draw (439,52) node [anchor=north west][inner sep=0.75pt]   [align=left] {{{87}}};
\draw (480,52) node [anchor=north west][inner sep=0.75pt]   [align=left] {{{88}}};
\draw (510,52) node [anchor=north west][inner sep=0.75pt]   [align=left] {{{89}}};
\draw (540,52) node [anchor=north west][inner sep=0.75pt]   [align=left] {{{90}}};
\draw (380,72) node [anchor=north west][inner sep=0.75pt]   [align=left] {{{79}}};
\draw (303,110) node [anchor=north west][inner sep=0.75pt]   [align=left] {{{$T_{k+1}=$}}};
\end{tikzpicture}
} \\
\midrule 
\multicolumn{2}{p{14.5cm}}{\textbf{Propositions}} \\
{24: must be generalized(that, \$21, \$25) \newline
21: the simple scaling(\$22) \newline
\textit{22: see(we, at most critical points)} \newline
\textcolor{blue}{25: to multiscaling(in turbulence)} \newline
\textcolor{blue}{71: behooves(therefore, it, us, \$72, \$75, \$77)} \newline
\textit{72: to examine first the dynamic multiscaling(of structure functions, \$73)} \newline
\textit{73: in a shell model for MHD(three dimensional, 3D MHD)} \newline
\textit{75: are related(dynamic multiscaling exponents, by linear bridge relations to equal time multiscaling exponents)} \newline
77: have not been able(we, \$78)
} &
{78: to find(\$79, so far) \newline
79: such relations(for MHD turbulence) \newline
80: and(\$71; scalar turbulence) \newline
\textcolor{blue}{81: obtain(therefore, we, \$82, \$84)} \newline
\textcolor{blue}{82: and(equal time, \$83)} \newline
\textcolor{blue}{83: time \_dependent structure functions(for a shell model)} \newline
\textcolor{blue}{84: for 3D MHD turbulence from these(and, \$86)} \newline
\textcolor{blue}{85: equal time (dynamic, \$87)} \newline
86: and(\$85) \newline
\textcolor{blue}{87: multiscaling(exponents)} \newline
88: try(then, we, \$89) \newline
89: to see(\$90) \newline
90: suggest(if, these, any bridge relations)
} \\
\midrule
\multicolumn{2}{p{14.5cm}}{\textbf{Gold Summary}\newline We present the first study of the \textcolor{blue}{multiscaling of time-dependent velocity and magnetic-field structure functions} in homogeneous, isotropic \textcolor{blue}{Magnetohydrodynamic (MHD) turbulence in three dimensions} .\newline
We generalize the formalism that has been developed for analogous studies of time-dependent structure functions in fluid turbulence to MHD. \newline
By carrying out detailed numerical studies of such \textcolor{blue}{time-dependent structure functions in a shell model for three-dimensional MHD turbulence, we obtain both equal-time and dynamic scaling exponents} .} 
\\ \bottomrule
\end{tabular}
\caption{Simulation example of \textsc{TreeKvD} (left) and \textsc{GraphKvD} (right). 
Each memory cycle shows the input sentence, extracted propositions, and the derived memory tree.
Function \text{aP} refers to subroutine \texttt{attachPropositions} in Algorithm~\ref{alg:kvd}.
Solid line: edge in final memory tree; dotted line: pruned edge; red dotted line: edge connecting $T$ and $P$.
Squared nodes: propositions recalled from long-term memory; underlined node: new root of memory tree. 
Relevant content common in propositions and the gold summary is coloured in blue.
}
\label{fig:ex-rec}
\end{figure}

\paragraph{Memory Cycles.}
In cycle $k$, both systems manage to attach the incoming proposition tree $P$ directly to the current memory tree $T_{k-1}$, with such connections illustrated as red dotted lines in Figure~\ref{fig:ex-rec}.
Notice that \textsc{TreeKvD} is allowed to make only one connection ($79 \mapsto 81$) so that the resulting structure, $T'$, remains a tree. 
In contrast, \textsc{GraphKvD} is allowed to connect each node in $P$ back to $T_{k-1}$ (e.g.\ $84 \mapsto 79$, $85 \mapsto 71$), which results in structure $G'$, an undirected weighted graph.
After choosing the new root (node $81$), the retention process (function $\texttt{memorySelect}$) selects the new memory tree $T_k$.

In the next cycle, $k+1$, the incoming $P$ cannot be attached directly to $T_k$ and hence, the recalled mechanism is used.
\textsc{TreeKvD} recalls a 3-node path to connect node $88$ to $81$, 
linking information about proposed models (\textit{`models for turbulence'} in $81$) to methodology
(\textit{`scaling methods'} in $25,24,21$) and hypothesis exploration (\textit{`we try to see if these suggest'} in $88$).
In contrast, \textsc{GraphKvD} recalls a single node linking the studied phenomenon (\textit{`MHD turbulence'} in $81$) to its properties of interest (\textit{`such relations'} in $79$, making reference to information in $75$)
and to the specific property being studied (\textit{`bridge relations'} in $90$).

\paragraph{Properties of Memory Trees.}
Properties of memory structures at the micro level, as discussed in Section~\ref{section:prop-kvd},
have the potential to greatly influence the level of lexical cohesion and redundancy in output summaries, in addition to identifying relevant content to be included.
We now elaborate on how this influence manifests in our example.

First, regarding lexical cohesion, a connected memory tree is evidence that content units currently held in memory are not a disjoint set of mutually exclusive concepts but a set that can be interpreted in a coherent manner. For instance, the content in $T_{k-1}$ could be verbalized in the following manner:
\begin{displayquote}
\textit{We examine dynamic multiscaling...in a shell model for 3D MHD [71,72] and scalar turbulence [80].
Dynamic multiscaling exponents are related by linear bridge relations to equal time multiscaling exponents [75].
We have not been able to find such relations for MHD turbulence so far [77,78,79].}
\end{displayquote}
\noindent where the propositions used to verbalize each phrase or sentence are indicated inside square brackets.
As can be seen, the text above reads smoothly and exhibits an acceptable level of lexical cohesion and co-referential coherence.
By updating the score of a set of propositions capable of forming a coherent text, a KvD system encourages the similar ranking of mutually coherent propositions.
Hence, a content selector is also encouraged to select a set of sentences exhibiting a non-trivial level of lexical cohesion.

A similar reasoning can be applied to explain the influence of memory simulation over redundancy in output summaries.
As claimed in Section~\ref{section:prop-kvd}, a memory tree constitutes a non-redundant set of propositions, with each proposition adding details of an entity or topic shared with the propositions it is connected to.
For instance, node \num{81} adds information about `MHD turbulence' to $T_{k-1}$ when connected to node \num{79}.
Moreover, when the recall mechanism is used in cycle $k+1$, only one recall path is added to $T_k$ (${25,24,21}$ in \textsc{TreeKvD} and ${79}$ in \textsc{GraphKvD}) instead of many potentially redundant recall paths.
Hence, by updating the score of a minimally redundant set of propositions in each cycle, a KvD system encourages non-redundant content to be ranked closely 
and by extension, the content selector is encouraged to select sentences with an acceptable level of redundancy.


Finally, memory trees are capable of identifying and ranking relevant propositions, hence encouraging a selector to pick sentences with relevant content.
In our example, we observe that both \textsc{TreeKvD} and \textsc{GraphKvD} retain propositions 81, 84, 85, 86 in $T_k$ and $T_{k+1}$.
These propositions cover information directly mentioned in the gold summary, coloured in blue in Figure~\ref{fig:ex-rec}.


\section{Experimental Setup}
\label{section:exp-setup}

In this section, we present the experimental setup for assessing the trade-off between informativeness, redundancy, and cohesion, under the two control scenarios
defined in previous sections, reward-guided and unsupervised.
We evaluate our models on the task of extractive summarization of scientific articles and
define appropriate automatic evaluation metrics to capture the analyzed summary properties.
Moreover, we design two human evaluation campaigns aimed at quantifying the perceived informativeness and cohesion of summaries produced by the proposed unsupervised systems,
\textsc{TreeKvD} and \textsc{GraphKvD}.
In the following, we elaborate on the datasets used and the preprocessing employed,
the comparison systems, and the setup for automatic and human evaluation.

\subsection{Datasets}
We used \textsc{PubMed} and \textsc{arXiv} datasets \shortcite{cohan2018discourse}, consisting of scientific articles in English in the Biomedical and Computer Science, Physics domains, respectively.
For each article, the source document is defined as the concatenation of all section texts, and 
the abstract is used as reference summary.
We further preprocessed both datasets after noticing substantial sentence tokenization errors and pollution of latex code.
Instances with documents with less than 5 tokens in the abstract are ignored.
Sentences are capped at 200 tokens, and sentences with more than 3 latex code keywords (e.g.\ \textit{usepackage, documentclass}) and less than 5 tokens are ignored.
Following previous work \cite{xiao2020systematically,gu2022memsum}, we use a budget of \texttt{B}$=200$ tokens for both \textsc{arXiv} and \textsc{PubMed}.

\subsection{Comparison Systems}

In addition to the discussed and proposed models, we report results on a range of standard heuristic and unsupervised baseline systems.
As heuristic baselines, we include the following: 
extractive oracle, \textsc{Ext-Oracle}, which consists of greedily selecting a set of sentences that maximize the sum of ROUGE-1 and ROUGE-2 scores w.r.t.\ the reference summary; 
\textsc{Lead}, selecting the leading sentences of a document until the budget is met; and \textsc{Random}, randomly sampling sentences following a uniform distribution.
Next, we elaborate on the training details and hyper-parameter configuration of our reward-based and unsupervised systems.

\paragraph{Supervised and Reinforcement Learning Systems.}
We report the performance of \textsc{E.LG} as a reference for an informativeness-oriented baseline, and
use checkpoints provided by \cite{xiao2020systematically}.
For redundancy-oriented model \textsc{E.LG-MMRSel+}, we use the default hyper-parameter configuration \cite{xiao2020systematically} and set $\lambda_\text{R}=0.6$, $\gamma_\text{R}=0.99$.
For local coherence-oriented model \textsc{E.LG-CCL}, we tune $\lambda_{\text{LC}}$ over validation sets and set it to $\lambda_{\text{LC}}=0.2$.
Both models were trained using Adam optimizer \cite{loshchilov2018decoupled}, batch size of \num{32}, learning rate of $10^{-7}$,
and trained for \num{20} epochs, with the best checkpoint selected based on the sum of ROUGE-1 and ROUGE-2 scores.

In addition, we compare against \textsc{MemSum} \cite{gu2022memsum}, a model that employs a multi-step episodic Markov decision process
that samples a candidate summary sentence by sentence instead of sampling the complete summary via a single action \cite{narayan2018ranking,dong-etal-2018-banditsum}.
Crucially, \textsc{MemSum} incorporates an \textit{extraction history} module that informs the agent about the information already selected and hence, minimize redundancy in the final summary.
Although the model is trained to produce a \textit{stop} action, we stop extraction once the budget is met in order to have a fairer comparison with other baselines
in terms of summary length.

Finally, we do not include supervised baselines that require the calculation of coreference chains or rhetorical structure trees over the input document, such as DiscoBERT \cite{xu2020discourse}, because of their limited applicability in out-of-domain scenarios and their inability to process documents of the length analyzed in this paper.

\paragraph{Unsupervised Systems.}
For the proposed KvD systems, we perform hyper-parameter tuning over the validation sets and set the maximum recall path length $R=5$, maximum tree persistence $\Psi=8$, working memory capacity $\texttt{WM}=100$ for both \textsc{TreeKvD} and \textsc{GraphKvD}.
For proposition scoring in \textsc{GraphKvD}, the decay factor is set to $\beta=0.01$.

We compare against unsupervised systems that model a document as a graph of sentences and employ node centrality as a proxy for informativeness.
First, we report on \textsc{TextRank} \cite{mihalcea2004textrank},\footnote{We use implementation in the Gensim library \cite{rehurek2010software}.} a system that employs TF-IDF as edge score between sentences and the PageRank algorithm \cite{brin1998anatomy} to obtain node centrality.
Second, we benchmark PacSum \cite{zheng2019pacsum}, which learns a specialized edge scorer and also uses PageRank.
For computational purposes, we limit connection to sentences in a window of size \num{200}.\footnote{Such a limitation was possibly not considered by \citeA{zheng2019pacsum} since their model was not designed for long documents, it was tested on the CNN/DM dataset in which documents are 50 sentences long in average.}
We report results using a SciBERT \cite{beltagy2019scibert} sentence embedded and two configurations: \textsc{PacSum}, using the default hyper-parameters reported by \citeA{zheng2019pacsum},
and \textsc{PacSum-FT}$^*$, finetuned over a sample of \num{1000} documents following the procedure therein.

Moreover, we investigate the appropriateness of constraining the size of working memory during KvD simulation, 
and define baseline \textsc{FullGraph}, which simulates all steps of KvD reading in Alg.~\ref{alg:kvd} except subroutine \texttt{memorySelect}.
Similarly to \textsc{PacSum}, proposition connection is limited to those in the previous \num{50} sentences.
Finally, we compared our proposed models against a previous implementation of the KvD theory \cite{fang2019proposition}, labeled as \textsc{FangKvD}.

\subsection{Automatic Evaluation}
\label{section:auto-eval}
We evaluate the intrinsic performance of the analyzed models in terms of informativeness, redundancy, and lexical cohesion.

\subsubsection{Informativeness}
We report F$_1$ ROUGE \cite{lin2004rouge} which measures lexical --n-gram-- overlap between extracted summaries and reference summaries, serving as an indicator for informativeness and relevancy.
Even though many issues have been identified when using ROUGE outside its proposed setting \cite{liu-liu-2008-correlation,cohan2016revisiting},
many variations of the original metric have shown a strong correlation with human assessment
\shortcite{graham2015re,shafieibavani2018graph,fabbri2021summeval}.

Nevertheless, ROUGE is not designed to appropriately reward semantic and syntactic variation in summaries.
For this reason, the semantic relevancy of summaries is assessed using F$_{1}$ BertScore \shortcite{zhang2019bertscore} which addresses semantic similarity by comparing contextual embeddings given by a pretrained BERT model.
\textsc{BertScore} has been proven a reliable metric when equipped with importance weighting in highly technical domains such as medical texts \shortcite{miura2021improving,hossain2020covidlies}.
In all our experiments, we report scores using RoBERTa \cite{roberta-model} as underlying model,
and apply importance weighting to diminish the effect of non-content words, e.g.\ function words.\footnote{IDF statistics were obtained from documents in the training set of each dataset.}

\subsubsection{Redundancy}
We assess redundancy in a text with the following metrics, each of which computes a value in the range of $[0;1]$, the higher it is the more redundant a text will be.

\paragraph{Inverse Uniqueness (IUniq).}
Defined as $\text{IUniq}= 1 - \text{Uniq}$,
where Uniq refers to \textit{uniqueness} \cite{peyrard-etal-2017-learning}, a metric that measures the ratio of unique n-grams to the total number of n-grams.
We report the mean among values for unigrams, bigrams, and trigrams.



\paragraph{Sentence-wise ROUGE (RdRL).}
Defined as the average F$_1$ ROUGE-L score among all pairs of sentences \cite{bommasani2020intrinsic}.
Given candidate summary $\hat{S}$,
\begin{equation*}
    \text{RdRL} = \mean_{(x,y)\in \hat{S} \times \hat{S}, x\neq y} \text{ROUGE-L}(x,y).
\end{equation*}

\subsubsection{Cohesion}
The following measures of cohesion are used.


\paragraph{Extended Entity Grid (EEG).}
The Entity Grid \cite{barzilay2008modeling} 
models cohesion in a text by obtaining the probability of an entity appearing in a 
determined syntactic role (subject, object, or other) in a sentence, given its role in the previous two sentences.
Then, a discriminative model learns a score using entity role transition probabilities and saliency features such as frequency.
Later, the feature set was extended to include entity-specific features such as the presence of proper mentions, the number of modifiers, among others \citeA{elsner2011extending}.
We use the implementation part of the Brown Coherence Toolkit\footnote{ \textbf{https://web.archive.org/web/20200505174052/https://bitbucket.org/melsner/ \\ browncoherence} }
and train our models over \num{50000} uniformly chosen samples from each training set.

\paragraph{Entity Graph (EGr).} \cite{guinaudeau2013graph}
Models a text as a graph of sentences with edges connecting sentences that have at least one noun in common.
Following \citeA{zhao2023discoscore}, averaged adjacency matrix is reported as a proxy for cohesion.

\subsubsection{Local Coherence}
The local coherence of a summary is assessed using the CCL scorer defined in \S~\ref{section:elg-coh}
with a sentence window of \num{3} and padding of \num{1}.

\subsubsection{Metric Reliability}
\label{section:metric-rel}
The automatic metrics for cohesion and local coherence
used in this article present the following limitations that might impact their reliability.
Regarding metrics of cohesion, their reliability depends on the accuracy of noun extraction.
EEG employs a co-reference resolution tool \cite{ng2002improving} that uses lexical, grammatical, and semantic features, in order to extract and link nouns from sentences.
This method --rather limited to modern NLP standards-- is complemented by metric EGr, which instead employs strong neural taggers for noun extraction.

In the case of local coherence, reliability might be impacted by the length (in wordpieces) being scored at a time by the model \cite{steen-markert-2022-find}.
In this article, we train our CCL scorers using binary cross-entropy with positive and negative examples taken from different documents, hence mitigating the model bias for chunk length.

\subsection{Human Evaluation}
In addition to automatic metrics, we elicit human judgments to assess informativeness and cohesion
in two separate studies conducted on the Amazon Mechanical Turk platform.
We sampled 30 documents from the test set of \textsc{PubMed} and the respective summaries extracted by unsupervised systems optimizing for cohesion, i.e.\ \textsc{TreeKvD}, \textsc{GraphKvD}, and \textsc{PacSum}.

Annotators were awarded $\$1$ per Human Intelligence Task (HIT), translating to more than
\$15 per hour. These rates were calculated by measuring the average annotation time per HIT in a pilot study.
In order to ensure the quality of annotations, we required annotators to have an HIT approval rate higher than $99\%$, a minimum of \num{10000} approved HITs, be proficient in the English language, and have worked in the healthcare or medical sector before.
Furthermore, we implemented the following catch controls: 
(i) we asked participants to check checkboxes confirming they had read the instructions and examples provided,
and (ii) we discard HITs that were annotated in less than \num{5} minutes.\footnote{Time threshold obtained from pilot study measurements.}
Annotations that failed the controls were discarded in order to maximize the quality.
We now elaborate on the details of each study.


\paragraph{Informativeness.}
In the first study, subjects were shown the abstract and the introduction of a scientific article along with two system summaries. Subjects were then asked to select the most informative summary among them with the possibility to select both in case of a tie, following previous work \cite{fabbri2021summeval,wu2018learning}.
In each system pair comparison, a system is assigned rank 1 if its summary was selected as most informative, and rank 2 otherwise. In case of a tie, both systems are assigned rank 1.
Then, the score of a system is defined as its average ranking.
We collected three annotations per system-pair comparison and made sure that the same annotator was not exposed to the same document twice.
As an additional catch trial, we included in each annotation batch an extra instance with summaries extracted by the extractive oracle and the random baseline. 

\paragraph{Cohesion.}
Lexical chains are sequences of semantically related words \cite{morris1991lexical}, and the distribution of these chains across a text has been shown to be a strong indicator of cohesion \cite{barzilay-elhadad-1997-using,galley2003improving}.
We relax the concept of lexical chains and extend it to that of \textit{chains of summary content units (SCUs)}, where all SCUs in a chain cover semantically related content.

In our second study, we aimed to capture cohesive ties between sentences in a system summary by asking 
participants to identify SCU chains.
Following previous work on semi-automation of the pyramid method \cite{zhang2021finding},
we employ propositions --as extracted in Section~\ref{section:prop-extr}-- as surrogates for SCUs. 
Hence, a propositional chain is defined as a set of propositions that exhibit semantically related arguments.

\begin{figure}[t]
     \centering
     \includegraphics[width=\textwidth]{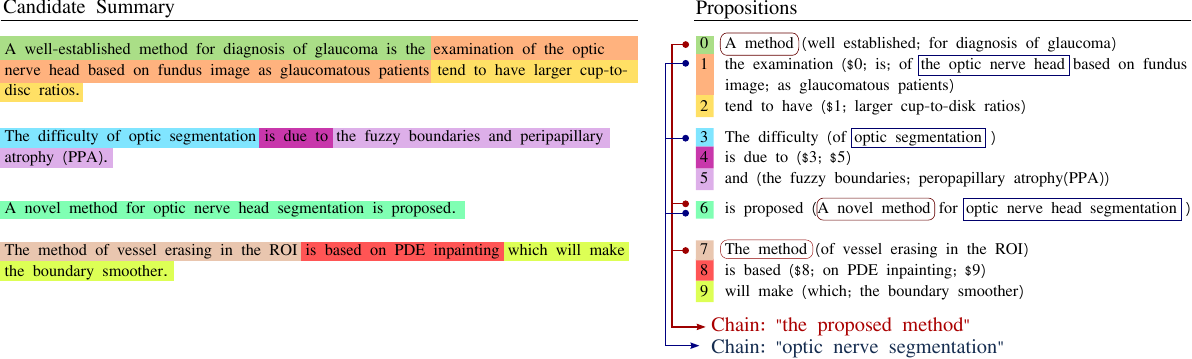}
    \caption{Example of proposition chain annotation in our cohesion evaluation campaign. Each coloured chunk in the candidate summary corresponds to a pre-extracted proposition. Users are tasked to group text chunks that share information by clicking on them. Best seen in colour.}
    \label{fig:coh-st}
\end{figure}

Participants were shown a single system summary as a list of sentences where tokens that belonged to the same proposition were colored the same, as depicted in the example in Figure~\ref{fig:coh-st}.
Then, the task consists of selecting chains of colored text chunks that share content among them.
For instance, in our example proposition chain $\{0,6,7\}$ is connected through information about \textit{the proposed method}, 
whereas chain $\{1,3,6\}$, through \textit{optic nerve segmentation}.
Chains were allowed to be non-exclusive, i.e.\ propositions can be selected in more than one group.
Similarly to the previous study, we collected three annotations per system summary 
and include the gold summary of an extra system in the campaign.

Finally, based on annotations of propositional chains, we define the following measurements of lexical cohesion:
(i) \textit{chain spread}, defined as the average number of sentences between two consecutive propositions in a chain;
(ii) \textit{chain density}, the number of chains covering the same sentence\footnote{We say that a chain \textit{covers} a sentence if at least one of the chain's proposition belongs to said sentence.};
and (iii) \textit{sentence coverage}, the number of sentences covered by at least one chain.
Intuitively, a text with less spread propositional chains exhibits cohesive ties that link sentences that are closer to each other, making the topic transition between sentences smoother \shortcite{halliday1976cohesion}.
Chain density can be interpreted as an indicator of the topic density in a sentence as well as 
how well a sentence connects to preceding and posterior sentences, e.g.\ by connecting to a preceding sentence through one chain
and connecting to a posterior one through another chain.
Finally, sentence coverage constitutes a straightforward measurement of how many sentences are connected through cohesive ties in a summary.

Agreement between human annotators is obtained by calculating the average text overlap between proposition chains, as follows.
Given candidate summary $\hat{S}$, let $C_A$ and $C_B$ be sets of chains extracted from $\hat{S}$ by annotators $A$ and $B$, respectively.
Given chains $a \in C_A$ and $b \in C_B$, we define Precision, Recall, and F$_1$ score as follows,
\begin{align*}
    \text{P}^{ov}(a,b) &= \frac{\sum_{p \in a} \max_{q \in b} |\text{LCS}(p,q)|}{ \sum_{p \in a} |p|}, \\
    \text{R}^{ov}(a,b) &= \frac{\sum_{q \in b} \max_{p \in a} |\text{LCS}(p,q)|}{ \sum_{q \in b} |q|}, \\
    \text{F}^{ov}_1(a,b) &= \frac{2 \cdot \text{P}^{ov} \cdot \text{R}^{ov}}{ \text{P}^{ov} + \text{R}^{ov} },
\end{align*}
\noindent where $p$ and $q$ are propositions included in chains $a$ and $b$, respectively,
$\text{LCS}(p,q)$ is the longest token sequence common to $p$ and $q$, and $|p|$ indicates the number of tokens covered by $p$.
Then, the overlap score between annotator $A$ and $B$ is defined as
\vspace{-0.2cm}
\begin{equation}
\label{eq:chain-overlap}
    \text{ChainOverlap}(A,B)= \frac{1}{|C_A|\cdot |C_B|} \sum_{a \in C_A, b \in C_B} \text{F}^{ov}_1 (a,b) .
\end{equation}
\noindent Finally, we report the average overlap score over all pairs of annotators, averaged over all system summaries.

\section{Results and Discussion}  
\label{section:res}

In this section, we present results for our proposed systems, \textsc{TreeKvD} and \textsc{GraphKvD}, and comparison systems on the \textsc{PubMed} and \textsc{arXiv} datasets.
First, we discuss the trade-offs systems incur when aiming to balance informativeness, redundancy, and lexical cohesion,
under varying setups of training supervision.
Then, we investigate how systems apply these trade-offs across increasing levels of source document redundancy.
Finally, we present a thorough analysis, both quantitative and qualitative, of how properties of simulated cognitive processes affect final summaries.
In all our experiments, statistical significance at the 95\% confidence level is estimated using bootstrap resampling \cite{davison1997bootstrap}.

\subsection{Informativeness, Redundancy, and Cohesion}

We start by analyzing the performance of our models in terms of relevancy, redundancy, and cohesion.
Results on informativeness are summarized in Table~\ref{c2-table:relv},
whereas results on redundancy, cohesion, and local coherence metrics are presented in Table~\ref{c2-table:red-coh}.
Both tables are organized in three sections: heuristic systems (\textit{Heur.}), supervised and reinforcement learning-based systems (\textit{Sup., R.L.\ }), 
and unsupervised systems (\textit{Unsup.}).
Systems are color-coded according to which summary properties they aim to optimize, such as informativeness (I), redundancy (R), and cohesion (C).
For completeness, we also report redundancy and cohesion of reference summaries (\textsc{Gold}, last row in Table~\ref{c2-table:red-coh}) to have a reference point for a desirable level of redundancy and cohesion.

Statistical significance at the system level is tested pairwise using Bootstrap resampling \cite{davison1997bootstrap} with a $95\%$ confidence interval.
For \textsc{PubMed}, we found no pairwise statistical difference between R1 scores of systems 
\textsc{TreeKvD} and \textsc{GraphKvD}; and between systems \textsc{E.LG}, \textsc{E.LG-MMRSel+}, and \textsc{E.LG-CCL}.
For \textsc{arXiv}, no pairwise statistical difference in R1 scores was found between systems 
\textsc{TreeKvD} and \textsc{GraphKvD}; and between systems \textsc{E.LG}, \textsc{E.LG-MMRSel+}, \textsc{MemSum}, and \textsc{E.LG-CCL}.
Analogously, Table~\ref{c2-table:relv} and ~\ref{c2-table:red-coh} indicate system groups in which no pairwise difference was found, one group per marker, for each metric reported.

\begin{table}[t]
\centering
\footnotesize
\begin{tabular}{llcccc|cccc}
\toprule
                                        &                                        & \multicolumn{4}{c}{\textbf{PubMed}}                                                                                                                          & \multicolumn{4}{c}{\textbf{arXiv}}                                                                                                                           \\ 
\multicolumn{1}{l}{\multirow{-2}{*}{\textbf{Aim}}}     & \multirow{-2}{*}{\textbf{System}}      & \multicolumn{1}{c}{\textbf{R1}}       & \multicolumn{1}{c}{\textbf{R2}}       & \multicolumn{1}{c}{\textbf{RL}}       & \multicolumn{1}{c|}{\textbf{BSc}}     & \multicolumn{1}{c}{\textbf{R1}}       & \multicolumn{1}{c}{\textbf{R2}}       & \multicolumn{1}{c}{\textbf{RL}}       & \multicolumn{1}{c}{\textbf{BSc}}     \\ \midrule
-                             & {\color{BrickRed} Ext-Oracle}      & {\color{BrickRed} 59.62}          & {\color{BrickRed} 35.14}          & {\color{BrickRed} 54.52}          & {\color{BrickRed} 88.22}          & {\color{BrickRed} 58.66}          & {\color{BrickRed} 30.28}          & {\color{BrickRed} 52.28}          & {\color{BrickRed} 87.12}          \\
-                             & {\color{OliveGreen} Lead}            & {\color{OliveGreen} 37.07}          & {\color{OliveGreen} 12.73}          & {\color{OliveGreen} 33.28}          & {\color{OliveGreen} 82.94}          & {\color{OliveGreen} 36.46}          & {\color{OliveGreen} 9.78}           & {\color{OliveGreen} 32.02}          & {\color{OliveGreen} 82.58}          \\
-                             & Random                                 & 36.11                                 & 10.43                                 & 32.70                                 & 82.21                                 & 33.02                                 & 6.52                                  & 30.09                                 & 80.51                                 \\ \midrule
{\color{BrickRed} I}      & {\color{BrickRed} E.LG}            & {\color{BrickRed} 47.34}\ddag          & {\color{BrickRed} 21.04}\ddag          & {\color{BrickRed} 42.42}\ddag          & {\color{BrickRed} 85.17}\ddag          & {\color{BrickRed} 46.38}\ddag          & {\color{BrickRed} 18.66}\ddag          & {\color{BrickRed} 40.77}          & {\color{BrickRed} 85.01}\ddag          \\
{\color{Blue} I,R}    & {\color{Blue} E.LG-MMRSel+}    & {\color{Blue} 47.55}\ddag          & {\color{Blue} 21.20}\ddag          & {\color{Blue} 42.70}\ddag          & {\color{Blue} 85.21}\ddag          & {\color{Blue} 46.52}\ddag          & {\color{Blue} 18.69}\ddag          & {\color{Blue} \textbf{41.06}}\ddag & {\color{Blue} 85.00}\ddag          \\
{\color{Blue} I,R}    & {\color{Blue} MemSum}          & {\color{Blue} \textbf{48.02}} & {\color{Blue} \textbf{22.06}} & {\color{Blue} \textbf{43.16}} & {\color{Blue} \textbf{85.63}} & {\color{Blue} \textbf{46.69}}\ddag & {\color{Blue} \textbf{19.50}} & {\color{Blue} 41.02}\ddag          & {\color{Blue} \textbf{85.13}} \\
{\color{OliveGreen} I,C}   & {\color{OliveGreen} E.LG-CCL}        & {\color{OliveGreen} 47.42}\ddag          & {\color{OliveGreen} 21.21}\ddag          & {\color{OliveGreen} 42.57}\ddag          & {\color{OliveGreen} 85.34}          & {\color{OliveGreen} 46.35}\ddag          & {\color{OliveGreen} 18.74}\ddag          & {\color{OliveGreen} 40.80}          & {\color{OliveGreen} 85.05}\ddag          \\
{\color{OliveGreen} I,C}      & \textit{\color{OliveGreen} PacSum-FT $^*$}       & {\color{OliveGreen} 40.05}          & {\color{OliveGreen} 13.66}          & {\color{OliveGreen} 36.29 }          & {\color{OliveGreen} 83.86} & {\color{OliveGreen} 38.05}          & {\color{OliveGreen} 9.87}           & {\color{OliveGreen} 34.18} & {\color{OliveGreen} 83.06} \\  \midrule
{\color{BrickRed} I}      & {\color{BrickRed} FullGraph}       & {\color{BrickRed} 35.48}          & {\color{BrickRed} 11.06}          & {\color{BrickRed} 30.28}          & {\color{BrickRed} 81.89}          & {\color{BrickRed} 27.44}          & {\color{BrickRed} 6.61}           & {\color{BrickRed} 22.75}          & {\color{BrickRed} 78.73}          \\
{\color{BrickRed} I}      & {\color{BrickRed} TextRank}        & {\color{BrickRed} \textbf{41.51}} & {\color{BrickRed} \textbf{15.37}} & {\color{BrickRed} \textbf{35.78}} & {\color{BrickRed} \textbf{83.59}}          & {\color{BrickRed} \textbf{40.32}} & {\color{BrickRed} \textbf{12.67}} & {\color{BrickRed} \textbf{34.06}}          & {\color{BrickRed} \textbf{82.68}}          \\
{\color{OliveGreen} I,C}      & {\color{OliveGreen} PacSum}          & {\color{OliveGreen} 37.01}          & {\color{OliveGreen} 10.07}          & {\color{OliveGreen} 33.55}          & {\color{OliveGreen} 82.98}          & {\color{OliveGreen} 33.41}          & {\color{OliveGreen} 6.54}           & {\color{OliveGreen} 30.48}          & {\color{OliveGreen} 81.70}          \\
{\color{Purple} I,R,C} & {\color{Purple} FangKvD}         & {\color{Purple} 35.80}          & {\color{Purple} 10.94}          & {\color{Purple} 30.97}          & {\color{Purple} 82.17}          & {\color{Purple} 32.76}          & {\color{Purple} 8.31}           & {\color{Purple} 27.81}          & {\color{Purple} 80.60}          \\
{\color{Purple} I,R,C} & {\color{Purple} TreeKvD (ours)}  & {\color{Purple} 37.22}\dag          & {\color{Purple} 11.40}\dag          & {\color{Purple} 32.37}\dag          & {\color{Purple} 82.61}\dag          & {\color{Purple} 34.90}\dag          & {\color{Purple} 9.06}\dag           & {\color{Purple} 29.85}\dag          & {\color{Purple} 81.16}\dag          \\
{\color{Purple} I,R,C} & {\color{Purple} GraphKvD (ours)} & {\color{Purple} 37.21}\dag          & {\color{Purple} 11.42}\dag          & {\color{Purple} 32.25}\dag          & {\color{Purple} 82.57}\dag          & {\color{Purple} 34.98}\dag          & {\color{Purple} 9.19}\dag           & {\color{Purple} 29.73}\dag          & {\color{Purple} 81.14}\dag          \\ \bottomrule
\end{tabular}
\caption{Performance of systems over \textsc{PubMed} and \textsc{arXiv} test sets in terms of ROUGE F$_1$ (R1, R2, RL) and BERTScore (BSc).
Optimization Aim (Aim) indicates whether a system was optimized for (I)nformativeness, (R)edundancy, Cohesion (C), or a combination of these, grouped by color.
Best models in each section are \textbf{bolded}.
(\dag,\ddag): no statistical difference between systems in the same section and column.
(*): non-completely supervised system.}
\label{c2-table:relv}
\end{table}

\begingroup
\setlength{\tabcolsep}{4pt} 

\begin{table}[t!]
\centering
\footnotesize
\begin{tabular}{llccccc|ccccc}
\toprule
                                        &                                        & \multicolumn{5}{c}{\textbf{PubMed}}                                                                                                                        & \multicolumn{5}{c}{\textbf{arXiv}}                                                                                                                         \\
\multicolumn{1}{l}{\multirow{-2}{*}{\textbf{Aim}}}              & \multirow{-2}{*}{\textbf{System}}      & \textbf{RdRL}                         & \textbf{IUniq}                          & \textbf{EEG}                      & \textbf{EGr}       & \textbf{CCL}                       & \textbf{RdRL}                         & \textbf{IUniq}                          & \textbf{EEG}                         & \textbf{EGr}  & \textbf{CCL}                         \\ \midrule
-                             & {\color{BrickRed} Ext-Oracle}      & {\color{BrickRed} 14.07}          & {\color{BrickRed} 18.72}          & {\color{BrickRed} 0.76}	& {\color{BrickRed} 0.84} & {\color{BrickRed} 0.58}               & {\color{BrickRed} 14.98}          & {\color{BrickRed} 18.78}          & {\color{BrickRed} 0.71}	& {\color{BrickRed} 0.72} & {\color{BrickRed} 0.40}                  \\
-                             & {\color{OliveGreen} Lead}            & {\color{OliveGreen} 12.75}          & {\color{OliveGreen} 18.25}          & {\color{OliveGreen} 0.72}	& {\color{OliveGreen} 0.78}	& {\color{OliveGreen} 0.76}
          & {\color{OliveGreen} 13.95}          & {\color{OliveGreen} 19.32}          & {\color{OliveGreen} 0.68}	& {\color{OliveGreen} 0.96}	& {\color{OliveGreen} 0.77}        \\
-                             & Random                                 & 11.36                                 & 18.29                                 & 0.63	& 0.69	& 0.41\ddag                        & 10.78                                 & 20.67                                 & 0.61	& 0.61	& 0.24 \\ \midrule
{\color{BrickRed} I}      & {\color{BrickRed} E.LG}            & {\color{BrickRed} 16.19}          & {\color{BrickRed} 21.60}\ddag          & {\color{BrickRed} 0.75}\ddag	& {\color{BrickRed} 1.03}	& {\color{BrickRed} 0.18}          & {\color{BrickRed} 16.71}\dag          & {\color{BrickRed} 21.20}\ddag          & {\color{BrickRed} 0.70}	& {\color{BrickRed} 1.01}	& {\color{BrickRed} 0.21}\ddag          \\
{\color{Blue} I,R}    & {\color{Blue} E.LG-MMRSel+}    & {\color{Blue} \textbf{15.03}} & {\color{Blue} \textbf{20.69}} & {\color{Blue} 0.75}\ddag	& \color{Blue}{0.96}	& {\color{Blue} 0.16}          & {\color{Blue} \textbf{14.58}} & {\color{Blue} \textbf{20.66}} & {\color{Blue} \textbf{0.71}}\dag	& {\color{Blue} 0.91} &	{\color{Blue} 0.21}\ddag \\
{\color{Blue} I,R}    & {\color{Blue} MemSum}          & {\color{Blue} 17.24}          & {\color{Blue} 24.01}          & {\color{Blue} 0.75}	& {\color{Blue} 0.75}	& {\color{Blue} 0.48}          & {\color{Blue} 16.80}\dag          & {\color{Blue} 21.89}\ddag          & {\color{Blue} 0.69}	& {\color{Blue} 1.03}	& {\color{Blue} 0.44}\dag \\
{\color{OliveGreen} I,C}   & {\color{OliveGreen} E.LG-CCL}        & {\color{OliveGreen} 16.92}          & {\color{OliveGreen} 21.21}\ddag          & {\color{OliveGreen} 0.75}\ddag	& {\color{OliveGreen} \textbf{1.04}}\dag	& {\color{OliveGreen} \textbf{0.51}}      & {\color{OliveGreen} 16.92}\dag          & {\color{OliveGreen} 21.21}\ddag          & {\color{OliveGreen} 0.70}\ddag	& {\color{OliveGreen} \textbf{1.05}}	& {\color{OliveGreen} \textbf{0.45}}\dag     \\
{\color{OliveGreen} I,C}      & \textit{\color{OliveGreen} PacSum-FT$^*$}       & {\color{OliveGreen} 12.92}          & {\color{OliveGreen} 18.76} & {\color{OliveGreen} 0.73}	& {\color{OliveGreen} 0.77}	& {\color{OliveGreen} 0.61}          & {\color{OliveGreen} 11.42}\ddag          & {\color{OliveGreen} 16.93} & {\color{OliveGreen} 0.72}	& {\color{OliveGreen} 0.67}\ddag	& {\color{OliveGreen} 0.56} \\ \midrule
{\color{BrickRed} I}      & {\color{BrickRed} FullGraph}       & {\color{BrickRed} 15.82}          & {\color{BrickRed} 23.79}          & {\color{BrickRed} 0.73}	& {\color{BrickRed} 0.68} &	{\color{BrickRed} 0.45}          & {\color{BrickRed} 11.65}\ddag          & {\color{BrickRed} 33.22}          & {\color{BrickRed} 0.56}	& {\color{BrickRed} 0.67}\ddag	& {\color{BrickRed} 0.24}  \\
{\color{BrickRed} I}      & {\color{BrickRed} TextRank}        & {\color{BrickRed} 22.08}          & {\color{BrickRed} 26.76}          & {\color{BrickRed} \textbf{0.78}}	& {\color{BrickRed} \textbf{1.05}}\dag	& {\color{BrickRed} 0.41}\ddag & {\color{BrickRed} 17.55}          & {\color{BrickRed} 22.25}          & {\color{BrickRed} \textbf{0.72}}\dag	& {\color{BrickRed} \textbf{1.02}} &	{\color{BrickRed} 0.26} \\
{\color{OliveGreen} I,C}      & {\color{OliveGreen} PacSum}          & {\color{OliveGreen} 11.66}          & {\color{OliveGreen} 20.84}\dag          & {\color{OliveGreen} 0.64}	& {\color{OliveGreen} 0.71}\ddag	& {\color{OliveGreen} \textbf{0.49}\dag}        & {\color{OliveGreen} 10.17}          & {\color{OliveGreen} \textbf{19.27}}          & {\color{OliveGreen} 0.62}	& {\color{OliveGreen} 0.44} &	{\color{OliveGreen} \textbf{0.40} }          \\
{\color{Purple} I,R,C} & {\color{Purple} FangKvD}         & {\color{Purple} 12.59}          & {\color{Purple} \textbf{20.45}\dag}          & {\color{Purple} 0.74}	& {\color{Purple} 0.70}\ddag	& {\color{Purple} \textbf{0.50}\dag}          & {\color{Purple} 12.15}          & {\color{Purple} 26.11}          & {\color{Purple} 0.66} &	{\color{Purple} 0.69}	& {\color{Purple} 0.34}  \\
{\color{Purple} I,R,C} & {\color{Purple} TreeKvD (ours)}  & {\color{Purple} 13.06}          & {\color{Purple} \textbf{20.62}\dag}          & {\color{Purple} 0.75}\dag	& {\color{Purple} 0.83} &	{\color{Purple} \textbf{0.49}\dag}        & {\color{Purple} 12.72}          & {\color{Purple} 24.22\dag}          & {\color{Purple} 0.70}\ddag	& {\color{Purple} 0.83}\dag &	{\color{Purple} 0.36}   \\
{\color{Purple} I,R,C} & {\color{Purple} GraphKvD (ours)} & {\color{Purple} \textbf{13.74}} & {\color{Purple} 21.00}\ddag          & {\color{Purple} 0.75}\dag	& {\color{Purple} 0.85}	& {\color{Purple} 0.44}      & {\color{Purple} \textbf{13.46}} & {\color{Purple} 24.57\dag}          & {\color{Purple} \textbf{0.71}}\dag	& {\color{Purple} 0.84}\dag &	{\color{Purple} 0.31}          \\ \midrule
& \textbf{Gold}                          & 13.54                                 & 19.12                                 & 0.70	& 0.96	& 0.91                                 & 14.83                                 & 17.27                                 & 0.72	& 0.87	& 0.89                                 \\ \bottomrule
\end{tabular}
\caption{Redundancy (RdRL, IUniq), cohesion (EEG, EGr), and local coherence (CCL) levels in candidate summaries over \textsc{PubMed} and \textsc{arXiv} test sets.
See Table 1 for details on Optimization Aim (Aim) and color coding.
Best models in each section are \textbf{bolded}, according to redundancy (those closest to \textsc{Gold}), cohesion and coherence (the higher the better).
(\dag,\ddag): no statistical difference between systems in the same section and column.
(*): non-completely supervised system.}
\label{c2-table:red-coh}
\end{table}

\endgroup

\paragraph{Heuristics.}
It is worth noting that the extractive oracle, \textsc{Ext-Oracle}, even though optimized for informativeness by design,
can still be used as a good-enough reference for redundancy in an extractive summary, given that RdRL and IUniq scores remain tightly close to those of \textsc{Gold}.
However, note that summaries extracted by \textsc{Ext-Oracle} need not be lexically cohesive, as indicated by its lower CCL scores than systems optimized for cohesion.
Instead, \textsc{Lead} does obtain high EEG, EGr, and CCL scores, and low RdRL and IUniq scores, a trend also present in \textsc{Gold}.
These measures indicate that such a trend is proper of cohesive text.
Notice, however, that source documents in \textsc{arXiv} might showcase lower lexical cohesion than those in \textsc{PubMed}, as indicated by their EEG and EGr scores.
Finally, it can be observed that the organization of information in scientific articles poses a challenge for trivial baselines,
as evidenced by the low ROUGE scores of \textsc{Lead} and \textsc{Random}.

\textbf{Supervised and Reinforcement Learning Systems.}
When optimizing one extra summary property besides informativeness in a reinforcement learning setup, the following insights can be drawn.
First, it is possible to reduce redundancy or improve lexical cohesion without losing informativeness:
\textsc{E.LG-MMRSel+} and \textsc{E.LG-CCL} obtain comparable \\ ROUGE scores to \textsc{E.LG}, a supervised system optimized only for informativeness.
\textsc{E.LG-MMRSel+} obtains the lowest redundancy scores (RdRL and IUniq) and \textsc{E.LG-CCL}, the highest cohesion and local coherence scores in terms of EGr and CCL, respectively.
However, optimizing for redundancy or informativeness alone incurs a huge sacrifice in terms of cohesion, as indicated by the low CCL scores.
On the other hand, optimizing for cohesion entails maintaining a non-trivial level of redundancy, as indicated by the RdRL and IUniq scores 
in \textsc{E.LG-CCL}, which are higher than those of \textsc{E.LG} and \textsc{E.LG-MMRSel+}.

Second, we find that tackling redundancy in the model architecture itself, i.e.\ \textsc{MemSum}, works consistently better than using a redundancy-aware reward during training, i.e.\ \textsc{E.LG-MMRSel+}.
Not only does \textsc{MemSum} obtain higher ROUGE scores, but seems to better balance cohesion and redundancy.
Even though \textsc{MemSum}'s CCL scores are lower than \textsc{E.LG-CCL} in both datasets, they are significantly higher than those of \textsc{E.LG-MMRSel+}.
Once again, we observe the trade-off between cohesion and redundancy, as indicated by the higher redundancy scores in \textsc{MemSum}.

\textbf{Unsupervised Systems.}
When comparing proxies for relevancy,
we find that sentence centrality (as in \textsc{TextRank} and \textsc{PacSum-FT}) performs better than sentence scoring based on reading comprehension, such as in our proposed KvD systems.
However, whilst \textsc{TextRank} obtains the highest ROUGE-1 and 2 scores in both datasets, it also obtains the highest redundancy scores (in terms of RdRL) and low CCL scores
(lowest in \textsc{PubMed} and second to lowest in \textsc{arXiv}).
A similar trend can be observed for \textsc{FullGraph}.
Since both \textsc{FullGraph} and \textsc{TextRank} use PageRank to rank content, we can conclude 
that lexical overlap at the sentence level is more beneficial than
overlap at the proposition argument level, as done by \textsc{FullGraph}.
Interestingly, EEG and EGr scores for \textsc{TextRank} are surprisingly high in both datasets.
Upon closer inspection, we found that EEG detects very few entity chains --most of the time a single one-- with high probability.
For EGr, this translates into having a sentence graph where edges are a result of co-occurrence of the same very few nouns.
This phenomenon can be interpreted as a sign of poor content coverage and high redundancy.

Consider now systems \textsc{PacSum} and \textsc{PacSum-FT}.
First, we notice that perhaps unsurprisingly, finetuning over in-domain data gives huge improvements in relevancy and a better cohesive-redundancy trade-off.
Second, unlike the supervised scenario, we observe that adding a proxy for cohesion during training significantly hurts relevancy.
This can be observed by the higher ROUGE-1 and 2 scores of \textsc{TextRank} against \textsc{PacSum-FT}.
Notice, however, that fluency (ROUGE-L) and semantic relevancy (BertScore) do experiment an improvement.
Moreover, \textsc{PacSum-FT} obtains more cohesive summaries than \textsc{Ext-Oracle} and even the supervised baseline optimized for local coherence, \textsc{E.LG-CCL}.
We hypothesize that \textsc{PacSum} and \textsc{PacSum-FT} model a strong proxy for cohesion by encouraging strong connections between neighboring sentences.

When comparing KvD systems in terms of relevancy scores (ROUGE-1 and 2), we observe that \textsc{GraphKvD} and \textsc{TreeKvD} significantly outperform other unsupervised baselines, except \textsc{TextRank}.
Notice, once again, that \textsc{PacSum} obtains better fluency (ROUGE-L) and semantic relevancy (BERTScore).
Whilst \textsc{PacSum} aims to optimize local coherence, it does not explicitly encourage lexical cohesion, as indicated by its EEG and EGr scores, lower than KvD systems.
In contrast, KvD systems improve lexical cohesion, which translates into higher EEG and EGr scores and in turn, slightly higher redundancy scores.
The contrast is more defined when the source documents present low lexical cohesion, as is the case for \textsc{arXiv}.

It is worth noting the advantage of the proposed KvD systems against a previous implementation of the KvD theory, \textsc{FangKvD}.
We hypothesize two reasons behind this result.
First, \textsc{FangKvD} relies on external domain-dependant resources like WordNet, which makes it hard to apply in highly domain-specific applications such as the scientific domain.
Second, \textsc{GraphKvD} and \textsc{TreeKvD} score propositions based on their position on the memory tree during simulation, whereas \textsc{FangKvD} only counts how many times a proposition has appeared in a memory cycle.
Note also that our proposed KvD systems outperform \textsc{FullGraph}, highlighting the importance of constraining working memory in each cycle.
In terms of cohesion-redundancy trade-off, we observe that \textsc{TreeKvD} obtains a comparable balance to \textsc{FangKvD} in \textsc{PubMed} but a better balance for \textsc{arXiv}.
Notice that in both datasets, \textsc{GraphKvD} obtains redundancy scores closest to \textsc{Gold} w.r.t.\ RdRL but lower CCL scores than \textsc{TreeKvD}.
In contrast, EEG and EGr scores indicate that \textsc{GraphKvD} maintains a comparable level of lexical cohesion to \textsc{TreeKvD}.


Lastly, it is important to point out the limited expressivity of the EEG metric (i.e.\ score gap at the system level) and the difference between trends in cohesion metrics and trends in CCL.
As mentioned in \S~\ref{section:metric-rel}, EEG is
limited by data sparsity —the limited lexical matching between nouns and entities– and the
performance of coreference resolution tools they use.
Hence, its expressivity is highly dependent on the accuracy of noun detection.
Regarding metric trends, metrics EEG and EGr were designed to capture lexical and semantic links between sentences in a text, therefore measuring cohesion.
While cohesion is considered a device to achieve local coherence, it does not model discourse structure.
In contrast, CCL was trained to capture sensible sentence orderings as a proxy for discourse organization on nearby sentences.
As such, it is capable of capturing not only lexical cohesive ties but also rhetorical orderings in a text.


\subsection{Effect of Document Redundancy}
\begin{figure}[t]
     \centering
     \begin{subfigure}[t]{\textwidth}
         \centering
         \includegraphics[width=\textwidth]{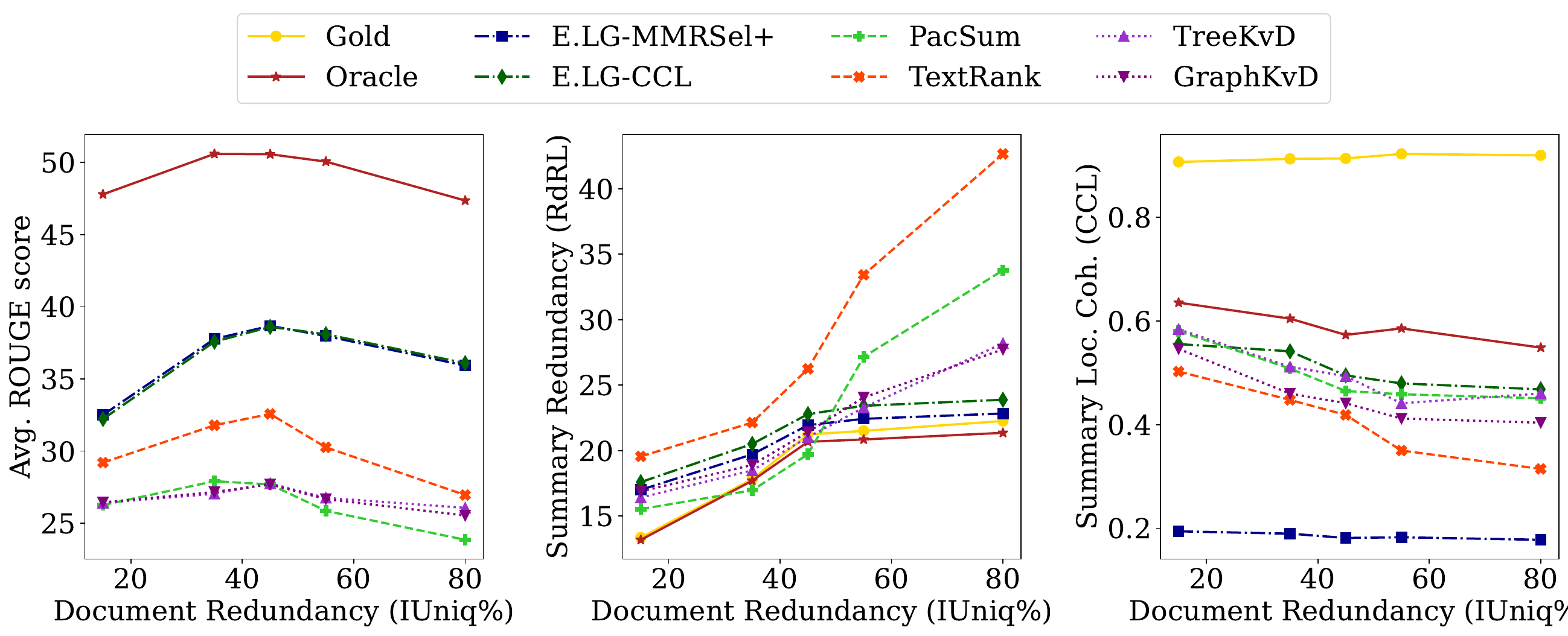}
         \caption{PubMed}
     \end{subfigure}
     \\ \vspace{1em}
     \begin{subfigure}[b]{\textwidth}
         \centering
         \includegraphics[width=\textwidth]{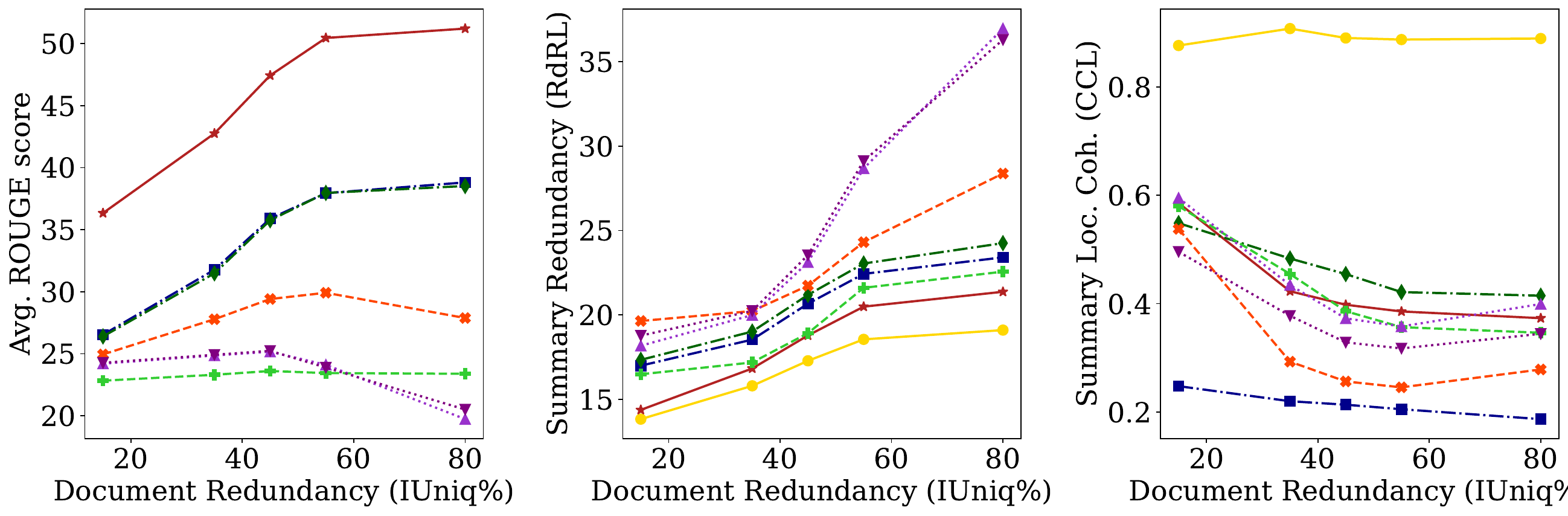}
         \caption{arXiv}
     \end{subfigure}
     
    \caption{Informativeness (left), summary redundancy (mid), and summary local coherence (right) across increasing levels of document redundancy. Metric values are averaged over each document redundancy range.}
    \label{c2-fig:red-bin}
\end{figure}

Next, we take a closer look at the redundancy and cohesion levels in summaries extracted from increasingly redundant documents.
Figure~\ref{c2-fig:red-bin} shows the performance of summarization systems in terms of informativeness (average ROUGE score, $\displaystyle (\text{ROUGE-1}+\text{ROUGE-2}+\text{ROUGE-L})/3$), redundancy (RdRL), and local coherence (CCL)
across different levels of document redundancy (IUniq).
Test sets were divided into bins according to their document redundancy score and the average metric value per bin is reported.
For simplicity, we only plot the performance of representative systems in each section.

\paragraph{Reinforcement Learning Systems.}
In general, we observe that performance in informativeness and redundancy degrades slightly but surely as redundancy increases in the source document.
Most notably, \textsc{E.LG-MMRSel+} and \textsc{E.LG-CCL} show comparable robustness in informativeness and redundancy, whilst
\textsc{E.LG-CCL} shows significantly better robustness in local coherence, highlighting the importance of 
optimizing for cohesion instead of redundancy.

\paragraph{Unsupervised Systems.}
In \textsc{PubMed}, we observe that \textsc{PacSum} and \textsc{TextRank} are highly susceptible to document redundancy, showing quick degradation in informativeness and redundancy as document redundancy increases. Whilst \textsc{PacSum} remains robust in terms of cohesion, \textsc{TextRank} exhibits a significant drop.
In contrast, \textsc{TreeKvD} and \\ \textsc{GraphKvD} show more robustness w.r.t.\ informativeness, remain closer in redundancy to \textsc{Gold}, and show local coherence levels comparable to \textsc{E.LG-CCL}.
Notably, our KvD systems show comparable redundancy to the RL-based baselines at low and mid levels of document redundancy.
This indicates that our systems manage to successfully balance informativeness, redundancy, and cohesion across increasing levels of document redundancy.

In \textsc{arXiv}, however, a few differences can be observed.
First, \textsc{PacSum} shows notable robustness to document redundancy, and remains closer in redundancy to \textsc{Gold} than all other unsupervised systems.
Our KvD systems exhibit a degradation in informativeness and redundancy, although robustly keeping high levels of cohesion.
We hypothesize that KvD systems prioritize cohesion above informativeness and redundancy.
In addition, we point out that \textsc{arXiv} is composed of noisier text than \textsc{PubMed}, exhibiting a number of preprocessing errors that might affect the quality of the proposition extraction.\footnote{Such errors include sentence tokenization errors, incomplete equations, bibliography text included in the document, among others. Even though we re-processed the dataset, many of these errors persisted.}


\subsection{Human Evaluation}
The results of our human evaluation campaigns are showcased in Table~\ref{table:hum-eval-res}.
In both studies, statistical significance between system scores was assessed by making pairwise comparisons between all systems using a one-way ANOVA with posthoc Tukey tests with $95\%$ confidence interval.

\paragraph{Informativeness.}
After discarding annotations that failed the controls, we are left with \num{229} out of \num{270} instances (\num{30} documents, \num{3} system pairs, and \num{3} annotations per pair).
Inter-annotator agreement --Krippendorff’s alpha \cite{krippendorff2011computing}-- was found to be $0.73$.

We found that humans shown significantly higher preference for \textsc{TreeKvD} summaries compared to \textsc{PacSum} summaries ($p<0.01$),
highlighting the advantage of modeling informativeness using KvD reading simulation 
compared to using a sentence centrality proxy in an unsupervised setup.
All other system pair differences are not statistically significant.

\paragraph{Lexical Cohesion.}
We obtained \num{343} out of \num{360} summary-level annotation instances (\num{30} documents, \num{4} systems --including gold summaries--, and \num{3} annotations per summary) after applying the control filters.
In average, annotators identified $2.71$ groups per summary and $3.89$ propositions per group.
Chain overlap, as defined in~Equation~\ref{eq:chain-overlap}, was calculated at $0.97$.
Score differences between system pairs \textsc{TreeKvD}--\textsc{PacSum} and \textsc{GraphKvD}--\textsc{PacSum} were found to be statistically significant, for all the analyzed measurements of cohesion.
Similarly, gold summary scores are significantly different from all systems in chain spread and chain density, and different from \textsc{PacSum} in sentence coverage.

The following insights can be drawn from these results.
First, gold summaries present chains that span sentences that are either adjacent to each other or separated by one other sentence, as indicated by its chain spread scores.
Chains in \textsc{GraphKvD} summaries mostly span adjacent sentences, in stark contrast with \textsc{PacSum} chains which are separated by two sentences on average. 
Second, chain density scores indicate that sentences in gold summaries are covered by either one or two chains, whereas \textsc{GraphKvD} summary sentences are covered by two chains in average.
On the one hand, this indicates that KvD summaries present a smooth topic transition by linking a summary sentence to the previous one through one chain and to the following sentence through another chain.
On the other hand, we note that gold summaries show lower chain density than \textsc{GraphKvD} summaries on average.
We hypothesize that the lower chain density in gold summaries is due to 
the high technicality of the scientific domain, making it harder for annotators to identify cohesive ties of non-lexical nature.
Nevertheless, sentence coverage scores indicate that chains in \textsc{TreeKvD} and \textsc{GraphKvD} cover a comparable amount of sentences as chains in gold summaries.
In contrast, the low chain density and low sentence coverage scores of \textsc{PacSum}
indicate that fewer sentences (around only $54\%$ of them) in its summaries are connected through cohesive links, the rest being perceived as isolated.

In summary, explicitly modeling lexical cohesive links during reading allows 
our KvD systems to extract summaries that exhibit a smooth topic transition between
adjacent or near-adjacent sentences,
with cohesive links connecting significantly more sentences than 
\textsc{PacSum} summaries.


\begin{table}[t]
\centering
\begin{tabular}{lrrrr}
\toprule
\textbf{Criteria}   & \textbf{TreeKvD} & \textbf{GraphKvD} & \textbf{PacSum} & \textbf{Gold} \\ \midrule
(I) Ranking $\downarrow$      & \textbf{1.44}             & 1.47              & 1.59               & -             \\ \midrule
(C) Chain Spread $\downarrow$        & 1.15             & \textbf{1.08}              & 2.14               & 1.59          \\
(C) Chain Density $\uparrow$       & 1.89             & \textbf{2.29}              & 0.95               & 1.63          \\
(C) Sent. Coverage (\%) $\uparrow$ & 72.10             & \textbf{77.33}             & 54.64              & 73.82         \\
\bottomrule
\end{tabular}
\caption{Informativeness ranking (I) and cohesion scores (C) as a function of propositional chain properties,
according to human judgments.($\uparrow,\downarrow$): higher, lower is better.}
\label{table:hum-eval-res}
\end{table}

\subsection{Qualitative Analysis}

We performed a qualitative analysis of system summaries extracted by the compared systems (Figure~\ref{fig:ex-rl-pubmed-output} and ~\ref{fig:ex-unsup-arxiv-output})
by annotating the lexical chains in them and analyzing the spread of chains as well as their relevance and coverage.
Each sample is accompanied by its gold summary, informativeness (average ROUGE score), redundancy (RdRL), and local coherence level (CCL).

\paragraph{Reinforcement Learning Systems.}
Consider the example in Figure~\ref{fig:ex-rl-pubmed-output}, showing summaries extracted by \textsc{E.LG-MMRSel+}, \textsc{E.LG-CCL}, and \textsc{MemSum} from a document in \textsc{PubMed}.
First, it can be observed that the gold summary covers \num{6} lexical chains (all colored differently) and that
these chains can appear throughout the entire text but always span windows of three to four sentences at a time.
Note that chains spanning more than one sentence imply a non-trivial level of redundancy, as shown by RdRL$>0$.
These smooth transitions are detected by our local coherence classifier --which scores a text by sliding a window of 3 sentences-- and assigns a high CCL score.



Second, we can observe how \textsc{E.LG-MMRSel+} trades off informativeness for redundancy by noting that the candidate summary exhibits one dominant chain (\{miRNA expression\}), possibly regarded as most promising relevancy-wise.
Redundancy reduction is translated in poor coverage of other chains (e.g.\ \{miRNA\}, \{analysis\}), being also too spread out (e.g.\ \{biomarkers\}),
which is reflected in the low cohesion score of the summary.
In stark contrast, \textsc{E.LG-CCL} exhibits most chains spreading in spans of three sentences whilst still favouring a highly relevant chain (\{miRNA expression\}).
Note that this improvement in cohesion implied an increment in redundancy, as shown by the higher CCL score and slightly higher RdRL score.

Finally, \textsc{MemSum} exhibits two dominant chains (\{miRNA expression\} and \{CAD patients\}) which are highly informative, justifying the high ROUGE score of the system.
However, we observed a lower cohesion score compared to \textsc{E.LG-CCL}, which can be explained by how the chains are spread out in the summary.
Whilst some chains do span adjacent sentences (e.g.\ the two dominant chains), others spread further (e.g.\ \{biomarkers\}, \{control\}).
In terms of redundancy, the higher levels can be explained by the fact that chains have items with longer n-grams.
This could lead to higher RdRL scores since the metric calculates the longest common n-gram subsequence in two strings.
Moreover, one particular chain (\{miRNA\}) contains a high number of items, increasing the chance of higher lexical overlap between the sentences this chain covers.


  

    

\paragraph{Unsupervised Systems.}
Consider the example in Figure~\ref{fig:ex-unsup-arxiv-output}, showing summaries extracted by \textsc{TextRank}, \textsc{TreeKvD}, and \textsc{GraphKvD} from a highly redundant document ($\text{IUniq}=63.34\%$) in \textsc{arXiv}.
As observed in the previous example, the gold summary exhibits abundant lexical chains, although with varying degrees of coverage.
We notice two main chains spanning the entire summary, with the rest being mentioned only once or twice.
This sign of seemingly low lexical cohesion was observed to be a common property in \textsc{arXiv} articles,
perhaps attributed to the rather mathematical formality in the writing style, as opposed to articles in \textsc{PubMed}.
Nevertheless, our cohesion classifier is able to pick non-lexical cues and assign a high cohesion score.

Regarding \textsc{TextRank}, we observe that its centrality-based scoring steers the model to focus mainly on two chains,
although only one of them ended up being informative (\{frequencies\}).
The high ROUGE scores and extremely high redundancy score confirm that centrality is a strong proxy for relevancy but without
any redundancy reduction mechanism, the system will degrade into selecting repeating content.
High repetition, in turn, proves to affect cohesion negatively, as indicated by the low CCL score.
Most critically, \textsc{TextRank} is susceptible to select sentences with high --if not complete-- token overlap between them, e.g.\ \textit{`monopole'}, \textit{`frequency'}, and \textit{`ground state'}.
Upon closer inspection, we found that some documents present repeated sentences in different sections, e.g.\ repeating a claim or conclusion. 

In contrast, \textsc{TreeKvD} shows noticeably less repetitions and a more balanced coverage of lexical chains, as indicated by the lower redundancy score and comparable ROUGE score.
Most of the chains spread consistently across the entire summary, which translates into a perceived and measured improvement in cohesion.
Moreover, the system manages to recover the same two main chains present in the gold summary, and even covers short chains not covered by \textsc{TextRank} (\{Boson\}, \{Stringari's result\}).
Upon closer inspection, we found that groups of extracted sentences are never more than two sentences apart.

Finally, \textsc{GraphKvD} exhibits a decrease in the spreading of lexical chains, showing instead a clear and smooth transition across the summary.
This translated into an increase in cohesion, as indicated by a higher CCL score, which also impacts the redundancy score.
Similarly to \textsc{MemSum}, the higher redundancy score can be explained by the longer common n-grams between sentences.


\begin{figure}[t]
\tiny
\centering
{
\setlength{\tabcolsep}{6pt} 
\renewcommand{\arraystretch}{1.3} 
\begin{tabular}{p{9.5cm}ccc}
\toprule
\multicolumn{1}{l}{\textbf{System}}       & \multicolumn{1}{c}{\textbf{Avg. ROUGE}} & \multicolumn{1}{c}{\textbf{RdRL}}  & \textbf{CCL}  \\ \midrule
\textbf{Gold Summary}         & \multicolumn{1}{c}{\textbf{-}}          & \multicolumn{1}{c}{\textbf{12.6}}  & \textbf{0.86} \\
\multicolumn{4}{p{14.5cm}}{ \textcolor{red}{Coronary artery disease (CAD)} is the largest killer of males and females in the United States.
There is a need to develop innovative diagnostic markers for \textcolor{red}{this disease}. \textcolor{blue}{MicroRNAs (miRNAs)} are a class of noncoding RNAs that posttranscriptionally regulate the expression of genes involved in important cellular processes, and we hypothesized that the \textcolor{orange}{miRNA expression profile} would be altered \textcolor{orange}{in whole blood} samples of \textcolor{red}{patients with CAD}.
We performed a \textcolor{teal}{microarray analysis} on \textcolor{blue}{RNA} from the blood of 5 male subjects with \textcolor{red}{CAD} and 5 \textcolor{olive}{healthy subjects} (mean age 53 years).
Subsequently, we performed \textcolor{teal}{qRT-PCR analysis} of \textcolor{orange}{miRNA expression in whole blood} of \textcolor{red}{another 10 patients with CAD} and 15 \textcolor{olive}{healthy subjects}. We identified \textcolor{blue}{11 miRNAs} that were significantly downregulated in \textcolor{red}{CAD subjects} ($p < .05$).
Furthermore, we found an association between ACEI/ARB use and downregulation of several \textcolor{blue}{miRNAs} that was independent of the presence of significant \textcolor{red}{CAD}. In conclusion, we have identified a distinct \textcolor{orange}{miRNA signature in whole blood} that discriminates \textcolor{red}{CAD patients} from \textcolor{olive}{healthy subjects}.
Importantly, medication use may significantly alter \textcolor{orange}{miRNA expression}.
These findings may have significant implications for identifying and managing individuals that either have \textcolor{red}{CAD} or are at risk of developing the \textcolor{red}{disease}. }                                                                                                              \\ \midrule

\textbf{E.LG-MMRSel+} & \multicolumn{1}{c}{\textbf{31.60}}       & \multicolumn{1}{c}{\textbf{16.13}} & \textbf{0.20} \\
\multicolumn{4}{p{14.5cm}}{We sought to compare \textcolor{orange}{miRNA expression in whole blood} of patients with \textcolor{red}{angiographically significant CAD} to that of \textcolor{olive}{healthy aged-matched controls}.
We performed an initial exploratory \textcolor{teal}{microarray analysis} in 5 cases and \textcolor{olive}{controls} and then further examined the most highly \textcolor{orange}{expressed miRNAs} in an additional 15 cases and \textcolor{olive}{controls}.
The present study provides insight into \textcolor{orange}{whole blood levels of miRNAs} in \textcolor{red}{patients with CAD} compared to \textcolor{olive}{healthy subjects} and demonstrates their potential utility as \textcolor{purple}{biomarkers for vascular disease}.
Thus, \textcolor{orange}{miRNA expression signatures in tissues and blood} have a potential role in the diagnosis, prognosis, and assessment of therapy.
Study participants were recruited as part of the Emory Cardiology Biobank, consisting of 3492 consecutive patients enrolled prior to undergoing elective or emergent cardiac catheterization across three Emory Healthcare sites, between 2003 and 2008.
Validation of the changes in \textcolor{orange}{miRNA expression} observed here in larger studies will be a necessary step to confirm their candidacy as \textcolor{purple}{biomarkers} and therapeutic targets. Although we were able to detect some differences in \textcolor{orange}{whole blood miRNA levels} between \textcolor{olive}{healthy subjects} and \textcolor{red}{CAD patients} (\textcolor{blue}{miR-584}, in particular), our \textcolor{teal}{microarray data} suggest that, similar to other reports, levels of \textcolor{orange}{miRNAs in the blood} are low and microarrays may lack the sensitivity to adequately identify \textcolor{blue}{miRNAs} that might serve as \textcolor{purple}{vascular disease biomarkers}.}                                                                                                              \\ \midrule
\textbf{E.LG-CCL}     & \multicolumn{1}{c}{\textbf{31.82}}       & \multicolumn{1}{c}{\textbf{18.90}} & \textbf{0.60} \\
\multicolumn{4}{p{14.5cm}}{Thus, \textcolor{orange}{miRNA expression signatures in tissues and blood} have a potential role in the diagnosis, prognosis, and assessment of therapy.
In this study, we sought to compare \textcolor{orange}{miRNA expression in whole blood} of patients with \textcolor{red}{angiographically significant CAD} to that of \textcolor{olive}{healthy aged-matched controls}.
We performed an initial exploratory \textcolor{teal}{microarray analysis} in 5 cases and \textcolor{olive}{controls} and then further examined the most highly \textcolor{orange}{expressed miRNAs} in an additional 15 cases and \textcolor{olive}{controls}.
Study participants were recruited as part of the Emory Cardiology Biobank, consisting of 3492 consecutive patients enrolled prior to undergoing elective or emergent cardiac catheterization across three Emory Healthcare sites, between 2003 and 2008. 
Although we were able to detect some differences in \textcolor{orange}{whole blood miRNA levels} between \textcolor{olive}{healthy subjects} and \textcolor{red}{CAD patients} (\textcolor{blue}{miR-584}, in particular), our \textcolor{teal}{microarray data} suggest that, similar to other reports, \textcolor{orange}{levels of miRNAs in the blood} are low and microarrays may lack the sensitivity to adequately identify \textcolor{blue}{miRNAs} that might serve as \textcolor{purple}{vascular disease biomarkers}.
The present study provides insight into \textcolor{orange}{whole blood levels of miRNAs} in \textcolor{red}{patients with CAD} compared to \textcolor{olive}{healthy subjects} and demonstrates their potential utility as \textcolor{purple}{biomarkers for vascular disease}.
Validation of the changes in \textcolor{blue}{miRNA expression} observed here in larger studies will be a necessary step to confirm their candidacy as \textcolor{purple}{biomarkers and therapeutic targets}.}                                                                                                              \\ \midrule
\textbf{MemSum}       & \multicolumn{1}{c}{\textbf{34.16}}       & \multicolumn{1}{c}{\textbf{22.34}} & \textbf{0.33} \\
\multicolumn{4}{p{14.5cm}}{We sought to compare \textcolor{orange}{miRNA expression in whole blood} of \textcolor{red}{patients with angiographically significant CAD} to that of \textcolor{olive}{healthy age-matched controls}.
This analysis included \textcolor{blue}{miR-150, miR-584, miR-21, miR-24, miR-126, miR-92a, miR-34a, miR-19a, miR-145, miR-155, miR-222, miR-378, miR-29a, miR-30e-5p, miR-342, and miR-181d}.
Among these, we found that \textcolor{blue}{miR-19a, miR-584, miR-155, miR-222, miR-145, miR-29a, miR-378, miR-342, miR-181d, miR-150, and miR-30e-5p} were significantly downregulated in the blood of \textcolor{red}{patients with CAD} compared to \textcolor{olive}{healthy subjects} (Figure 2).
Several recent studies have indicated that there is a potential role for \textcolor{orange}{circulating miRNA levels} as \textcolor{purple}{valuable biomarkers} for different disease processes, including cancer, cardiomyopathy, and acute myocardial infarction.
In this study, we wanted to address the hypothesis that \textcolor{orange}{miRNA expression levels in blood} could predict the presence of significant \textcolor{red}{coronary artery disease in human subjects}.
We identified \textcolor{orange}{11 miRNAs whose expression} was significantly downregulated in patients with \textcolor{red}{angiographic evidence of significant atherosclerosis} compared to \textcolor{olive}{healthy subjects} that were matched for age and gender.
The present study provides insight into \textcolor{orange}{whole blood levels of miRNAs} in \textcolor{red}{patients with CAD} compared to \textcolor{olive}{healthy subjects} and demonstrates their potential utility as \textcolor{purple}{biomarkers for vascular disease}.} \\ \bottomrule
\end{tabular}
}
\caption{Summaries extracted by reinforcement learning-based systems for a \textsc{PubMed} sample with informativeness (average ROUGE score), redundancy (RdRL), and local coherence (CCL) scores.
Text is annotated with color-coded lexical chains, and was detokenized and truecased for ease of reading.}
\label{fig:ex-rl-pubmed-output}
\end{figure}

\begin{figure}[t]
\tiny
\centering
{
\setlength{\tabcolsep}{6pt} 
\renewcommand{\arraystretch}{1.3} 
\begin{tabular}{p{9.5cm}ccc}
\toprule
\multicolumn{1}{l}{\textbf{System}}       & \multicolumn{1}{c}{\textbf{Avg. ROUGE}} & \multicolumn{1}{c}{\textbf{RdRL}}  & \textbf{CCL}  \\ \midrule
\textbf{Gold Summary}         & \multicolumn{1}{c}{\textbf{-}}          & \multicolumn{1}{c}{\textbf{21.49}}  & \textbf{0.91} \\
\multicolumn{4}{p{14.5cm}}{ We study the collective excitations of a \textcolor{purple}{neutral atomic Bose-Einstein condensate} with \textcolor{olive}{gravity-like interatomic attraction} induced by \textcolor{teal}{electromagnetic wave}. Using the \textcolor{teal}{time-dependent variational approach}, we derive an \textcolor{red}{analytical spectrum for monopole and quadrupole mode frequencies} of a \textcolor{purple}{gravity-like self-bound Bose condensed state} at zero temperature.
We also analyze the \textcolor{red}{excitation frequencies} of the \textcolor{teal}{Thomas-Fermi gravity (tf-g) and gravity (g) regimes}.
Our result agrees excellently with \textcolor{orange}{that of Giovanazzi et al.}, which is obtained within the \textcolor{teal}{sum-rule approach}.
We also consider the \textcolor{blue}{vortex state}. We estimate the \textcolor{red}{superfluid coherence length} and \textcolor{red}{the critical angular frequencies} to create a \textcolor{purple}{vortex} around the X axis. We find that the \textcolor{teal}{tf-g regime} can exhibit the \textcolor{blue}{superfluid properties} more prominently than the \textcolor{teal}{g regime}. We find that \textcolor{red}{the monopole mode frequency} of the condensate decreases due to the presence of a \textcolor{purple}{vortex}. }                                                                                                              \\ \midrule

\textbf{TextRank} & \multicolumn{1}{c}{\textbf{38.99}}       & \multicolumn{1}{c}{\textbf{45.02}} & \textbf{0.23} \\
\multicolumn{4}{p{14.5cm}}{ The \textcolor{olive}{gravity-like potential} is balanced by the wave interaction strength. \textcolor{blue}{The ground state energy} per particle varies as @xmath.
\textcolor{red}{The monopole and quadrupole frequencies} obtained from the variational approach are similar to the exact numerical values.
The trap potential and wave interaction can be neglected.
\textcolor{blue}{The total ground state energy} is @xmath.
\textcolor{blue}{The ground state energy} per particle varies as @xmath.
One can use the \textcolor{teal}{time-dependent variational approach} to describe the \textcolor{blue}{vortex state}.
\textcolor{red}{The critical angular frequency} vs. the dimensionless scattering parameter is shown in Fig.4.
\textcolor{teal}{Tf-g regime}: for large \textcolor{teal}{wave scattering length}, \textcolor{blue}{kinetic energy} can be neglected.
\textcolor{red}{The critical angular frequencies} for @xmath and @xmath are @xmath and @xmath respectively.
\textcolor{red}{The monopole mode frequency} for an ordinary atomic bec in the tf regime is independent of the \textcolor{purple}{vortex}.
\textcolor{red}{The monopole mode frequency} for @xmath is @xmath.
The @xmath is also less than the \textcolor{red}{monopole mode frequency} in the \textcolor{blue}{vortex free condensate}.
In the \textcolor{teal}{tf regime} of an ordinary atomic bec, \textcolor{red}{the monopole and quadrupole mode frequencies} are independent of the \textcolor{teal}{scattering length}. }                                                                                                              \\ \midrule
\textbf{TreeKvD}     & \multicolumn{1}{c}{\textbf{39.87}}       & \multicolumn{1}{c}{\textbf{14.62}} & \textbf{0.36} \\
\multicolumn{4}{p{14.5cm}}{ In this system, the \textcolor{olive}{gravity-like attraction} balances the pressure due to the \textcolor{blue}{zero point kinetic energy} and the short range interaction potential.
The bec of \textcolor{purple}{charged Bosons} confined in an ion trap can be described by the above mentioned \textcolor{teal}{Lagrangian} if we set @xmath, where @xmath is the electronic charge. 
To calculate the \textcolor{red}{excitations spectrum of an atomic bec} with \textcolor{olive}{gravity-like interaction}, we will use the \textcolor{teal}{time-dependent variational method}.
\textcolor{teal}{This technique} has been first used to calculate the \textcolor{red}{low-lying excitations spectrum of a harmonically trapped atomic bec} in @xref.
The result obtained from the \textcolor{teal}{variational method} matches with \textcolor{orange}{Stringari's result} within the \textcolor{teal}{sum-rule approach}.
In @xref, it is shown that the \textcolor{red}{oscillation frequencies} obtained from the \textcolor{blue}{exact ground state} and a Gaussian Ansatz are in good agreement.
One can use the \textcolor{teal}{time-dependent variational approach} to describe the \textcolor{blue}{vortex state}.
In \textcolor{teal}{these regimes}, we have calculated the lower bound of the \textcolor{blue}{ground state energy}, \textcolor{red}{sound velocity, monopole and quadrupole mode frequencies}. }                                                                                                              \\ \midrule
\textbf{GraphKvD}       & \multicolumn{1}{c}{\textbf{39.73}}       & \multicolumn{1}{c}{\textbf{21.65}} & \textbf{0.51} \\
\multicolumn{4}{p{14.5cm}}{ Most of the properties of these \textcolor{purple}{dilute gas} can be explained by considering only \textcolor{olive}{two-body short range interaction} which is characterized by the \textcolor{teal}{S-wave scattering length}.
Therefore, we expand around the \textcolor{teal}{time dependent variational parameters around the equilibrium widths} in the following way, and @xmath.
The \textcolor{teal}{time evolution of the widths around the equilibrium points} are @xmath is the first order fluctuations around the equilibrium points of @xmath.
One can use the \textcolor{teal}{time-dependent variational approach} to describe the \textcolor{blue}{vortex state}.
The \textcolor{blue}{vortex state} play an important role in characterizing the \textcolor{blue}{superfluid properties} of \textcolor{purple}{Bose system}.
\textcolor{red}{The critical angular frequency} required to produce a \textcolor{blue}{vortex state} is where is the \textcolor{blue}{energy} of a \textcolor{blue}{vortex states} with vortex quantum number and is the \textcolor{blue}{energy} with no \textcolor{purple}{vortex}.
In \textcolor{teal}{these regimes}, we have calculated the lower bound of the \textcolor{blue}{ground state energy}, \textcolor{red}{sound velocity, monopole and quadrupole mode frequencies}. } \\ \bottomrule
\end{tabular}
}
\caption{Summaries extracted by unsupervised systems for an \textsc{arXiv} sample with informativeness (average ROUGE score), redundancy (RdRL), and local coherence (CCL) scores.
Text is annotated with color-coded lexical chains, and was detokenized and truecased for ease of reading.}
\label{fig:ex-unsup-arxiv-output}
\end{figure}


\subsection{How Simulated Cognitive Processes Affect Final Summaries}

The KvD theory describes cognitive processes involved in short-term memory manipulation and constraints over memory structures.
While it is well-understood how these processes and constraints would influence reading comprehension in a simulated environment, it is less intuitive to 
establish how they influence summary properties through sentence scoring.
In this section, we shed light on how final summaries are affected by the following KvD processes.
First, we investigate the impact of capacity in working memory and the impact of the strategy of proposition scoring used.
Then, the mechanisms in charge of recall and memory replacement (tree persistence) are discussed.
Finally, we investigate what kind of argument overlap strategy is best leveraged by our KvD systems.

\begin{figure}[t]
     \centering
     \begin{subfigure}[t]{\textwidth}
         \centering
         \includegraphics[width=\textwidth]{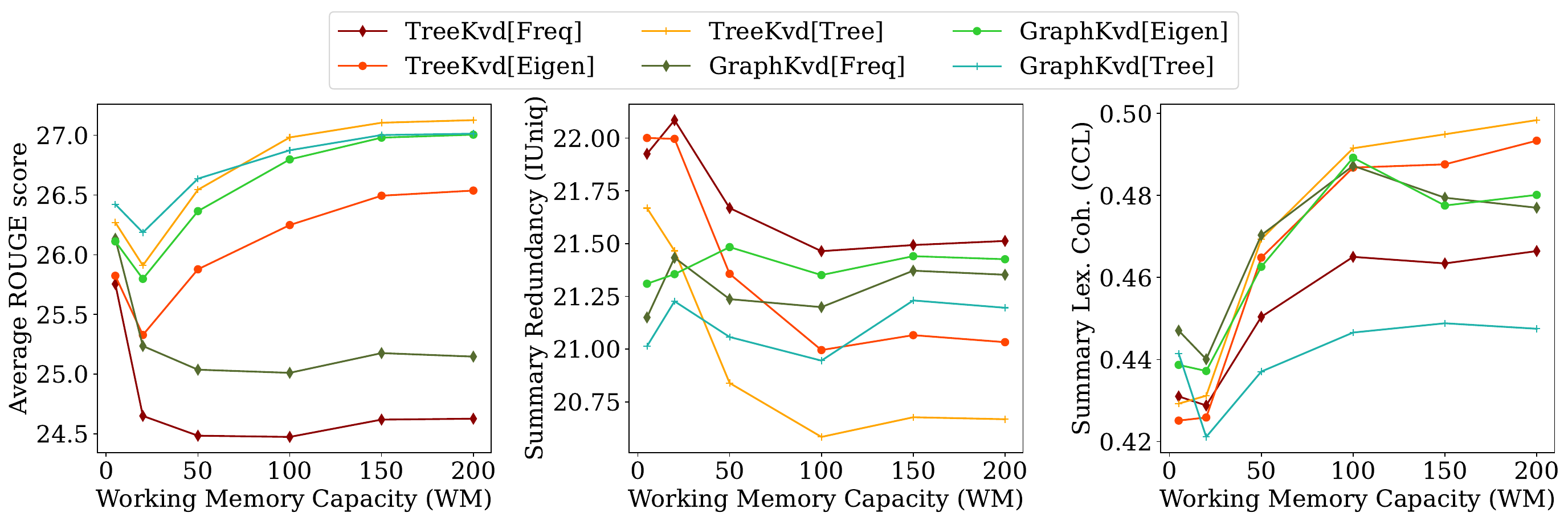}
         \caption{PubMed}
     \end{subfigure}
     \\ \vspace{1em}
     \begin{subfigure}[b]{\textwidth}
         \centering
         \includegraphics[width=\textwidth]{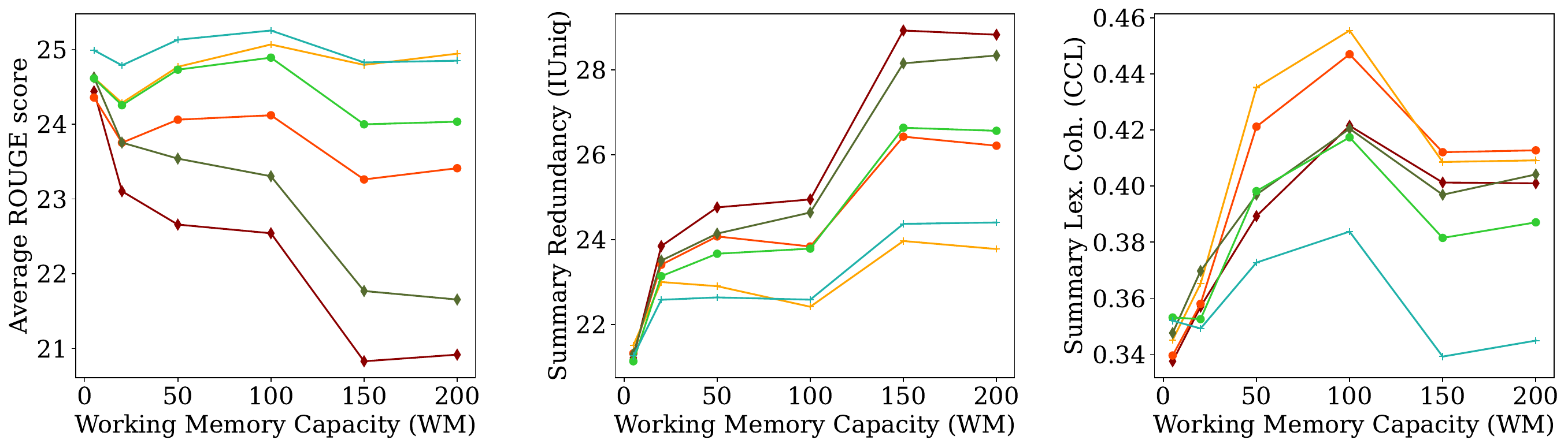}
         \caption{arXiv}
     \end{subfigure}
     
    \caption{Effect of proposition scoring strategy (\textsc{Tree}, \textsc{Eigen}, and \textsc{Freq}) and working memory capacity (\texttt{WM}) on summary informativeness (average ROUGE scores; left), redundancy (IUniq; middle), and local coherence (CCL).}
    \label{fig:pscore}
\end{figure}

\paragraph{Working Memory Capacity.}
Intuitively, the more memory capacity a KvD system has, the more propositions it will be able to retain in memory, increasing the chances that relevant propositions
are scored higher and are eventually selected for the final summary.
This is evidenced by the consistent increase in ROUGE scores for increasing memory capacity, \texttt{WM}, as shown in 
Figure~\ref{fig:pscore}.
However, we did observe an optimal capacity for redundancy and cohesion levels.
This indicates that, as the memory capacity increases, maintaining non-redundant information in the memory tree becomes more challenging.

Moreover, as seen in Table~\ref{c2-table:relv}, KvD systems with $\texttt{WM}=100$ obtain consistently higher relevancy scores than \textsc{FullGraph}, a system that does not simulate working memory and which scoring strategy has access to all the propositions in a document at all times.
This indicates that constraining the size of the memory tree in each iteration encourages KvD systems to retain only information relevant to the current local context.

Another aspect greatly influenced by working memory capacity is that of how much information in the source document can be covered.
As noted in Section~\ref{section:tkvd-lim}, it is possible that some propositions are pruned away and never recalled again, in which case their final score will be zero.
We say that a proposition is \textit{covered} by a KvD system if such a proposition appears at least once in a pruned memory tree during simulation.
Furthermore, we define document coverage as the ratio of covered propositions over the total number of propositions in a document.
Not surprisingly, we found that increasing working memory capacity increased document coverage in both \textsc{TreeKvD} and \textsc{GraphKvD}.
When $\texttt{WM}=5$, \textsc{TreeKvD} is able to cover $62\%$ of all document propositions in the \textsc{arXiv} test set,
and up to $96\%$ when $\texttt{WM}=100$.
\textsc{GraphKvD} further improves coverage to $78\%$ at $\texttt{WM}=5$ and $97\%$ at $\texttt{WM}=100$.
However, we found that \textsc{FangKvD} exhibits a much lower coverage: $22\%$ when $\texttt{WM}=5$ and up to $44\%$ when $\texttt{WM}=100$.
We hypothesize that the drastic improvement in \textsc{GraphKvD} is due to the diffusion mechanism that updates scores of direct neighbours of memory tree nodes.
Similar trends were observed in the \textsc{PubMed} dataset.
These results lay down evidence that the proposed computational implementations of KvD theory are effective at covering most --if not all- content units in a document during simulation.

So far in our analysis we have considered memory capacity as a hyper-parameter of a KvD system, 
expected to remain fixed throughout the entire simulation and fixed for all documents in an evaluation set.
The following question then arises when looking at each sample individually: what is the \textit{right} capacity of working memory in order to produce a summary with the most relevant content?
We attempt to answer this question by selecting for each sample in the validation set, the working memory size \texttt{WM} that yields the highest sum of ROUGE-1 and ROUGE-2 scores.
The results are encouraging: when using the best possible \texttt{WM} per sample in \textsc{arXiv}, \textsc{TreeKvD} exhibits an increase in absolute points of \num{3.19} in ROUGE-1, \num{2.36} in ROUGE-2, and \num{2.86} in ROUGE-L. 
This is compared to the best performing configuration, i.e.\ when using $\texttt{WM}=100$ for all samples.
Most surprisingly, the distribution of best \texttt{WM} per sample is rather balanced, with $26.5\%$ of samples preferring a $\texttt{WM}=100$, $26.7\%$ a $\texttt{WM}=50$, $24.11\%$ a $\texttt{WM}=20$, and $22.5\%$ a $\texttt{WM}=5$.
\textsc{GraphKvD} exhibits a similar increase of \num{3.06}, \num{2.33}, \num{2.75} in ROUGE-1, ROUGE-2, and ROUGE-L, respectively.
A similar trend was observed on the validation set of \textsc{PubMed}.
However, it should be noted that we did not find any strong correlation between working memory capacity and ROUGE or BertScore scores,
which indicates that the ability of a KvD system to produce relevant summaries is not influenced by its working memory capacity.
Instead, we suspect that memory capacity might be an indicator of text difficulty or cognitive easiness, however the exploration of this hypothesis falls out of the scope of this work and we leave it to future investigations.


\paragraph{Working Memory as a Tree.}
Next, we investigated the impact of leveraging the position of a node in the memory tree structure during proposition scoring.
We compared scoring function $c(\cdot)$ in Eq.~\refeq{eq:tkvd-ct}, labeled as \textsc{Tree}, against two other strategies.
The first one, denoted \textsc{Freq}, consists of a frequency heuristic, $c(t,T)=1, \forall t \in T$, which only counts how many memory cycles a proposition participates in.
The second strategy, denoted \textsc{Eigen}, scores nodes based on their eigen-vector centrality as:
\begin{equation*}
c(t,T)=\frac{1}{\lambda} \sum_{ \substack{v \\ s.t.\ (t,v)\in E[T]} } c(v,T)
\end{equation*}
\noindent where $\lambda$ is the largest eigen-value of the adjacency matrix of $T$.\footnote{We use the eigen-vector centrality implementation in the NetworkX Python library.}

Figure~\ref{fig:pscore} shows the performance of our KvD systems over the validation set of \textsc{PubMed} and \textsc{arXiv}.
Systems using scoring function $c(t,T)$ in Eq.~\refeq{eq:pscore} are labeled with \textsc{Tree}, e.g.\ \textsc{TreeKvD[Tree]}.
First, we observe that \textsc{Tree} scoring significantly outperforms \textsc{Eigen} and \textsc{Freq} scoring, for all values of working memory capacity in both datasets.
This results demonstrates the advantage of modeling memory as a tree structure and leveraging the position of a node for scoring,
compared to just considering memory as a bag of content units (as \textsc{Freq} does) or even using node centrality strategies, as done by \textsc{Eigen}.
However, it is worth noticing that for \textsc{GraphKvD}, the gap between \textsc{Tree} and \textsc{Eigen} diminishes as \texttt{WM} increases, even performing comparably in \textsc{PubMed}.
This might indicate that \textsc{GraphKvD} is superior than \textsc{TreeKvD} at placing highly influential (i.e.\ relevant) nodes closer to the root, in which case the proposition ranking given by \textsc{Tree} and \textsc{Eigen} is highly similar.

In conclusion, \textsc{Tree} scoring enables our implementations of KvD not only to better keep track of relevant information but also to better model cohesion in the memory tree, 
which translates to lower redundancy scores and higher cohesion scores in final summaries.

\paragraph{Recall Mechanism and Tree Persistence.}
Additionally, we investigated the effect of allowing our KvD systems to retrieve longer node paths during recalls, 
as well as the effect of allowing systems to persist memory trees for more cycles.
Whilst \cite{kintsch1978toward} do not define a limit for how many propositions can be recalled, \cite{fang2019proposition} limits recall to only one proposition for computational efficiency.
In this experiment, we test \textsc{TreeKvD} and \textsc{GraphKvD} with $\texttt{WM}=100$ and \textsc{Tree} scoring, and set the maximum allowed number of recalled nodes to $R=[2,5,8,10]$
and the maximum persistence parameter to $\Phi=[2,5,8,10]$.
When compared in the validation set of both datasets, no statistical difference was found within \textsc{TreeKvD} and \textsc{GraphKvD} varieties.
Absolute differences in average ROUGE scores were at most $0.1$, whereas differences in IUniq redundancy were at most $0.2$ percentual points.
These results indicates that our implementations of the KvD theory are robust to recall and memory replacement parameters, an encouraging result when planning to use these systems in other domains.

Lastly, it is worth pointing out an additional benefit of the tree persistence mechanism, observed empirically in Figure~\ref{fig:pscore}.
Tree persistence can be seen as a mechanism that guarantees that the content in WM changes periodically, providing the model with robustness to the length of an article section in a scientific article, and adding evidence to its applicability to other domains.
As mentioned in the previous chapter, sections in \textsc{PubMed} articles are shorter than those in \textsc{arXiv} ($16.8$ vs $28.8$ on average).
In \textsc{PubMed}, performance converges at $\texttt{WM}=150$, at which point there is enough capacity to keep all propositions read in the section so far. However, contrary to the behavior of \textsc{FangKvD} in the previous chapter, performance is not hurt at high capacity regimes,
with the persistence mechanism refreshing WM periodically.
In \textsc{arXiv}, sections are long enough for high WM capacity to be a problem, at which point WM starts storing noisy information which eventually hurts performance.

\begin{table}[t]
\centering
\scriptsize
\begin{tabular}{lrrrrl|rrrrl}
\toprule
\textbf{System}          & \multicolumn{5}{c}{\textbf{PubMed}}                                                                                                         & \multicolumn{5}{c}{\textbf{arXiv}}                                                                                                          \\ 
                & \textbf{R1} & \textbf{R2} & \multicolumn{1}{c}{\textbf{RL}}    & \textbf{IUniq} & \multicolumn{1}{c|}{\textbf{CCL}}                       & \textbf{R1} & \textbf{R2} & \multicolumn{1}{c}{\textbf{RL}}    & \textbf{IUniq} & \textbf{CCL}                       \\ \midrule
TreeKvD         &                        &                        & \multicolumn{1}{r}{}      &                          & \multicolumn{1}{r}{}     &                        &                        & \multicolumn{1}{r}{}      &                          & \multicolumn{1}{r}{}     \\
w/ Lex. Overlap & 35.93                  & 12.63                  & \multicolumn{1}{r}{31.53} & 19.05                    & 0.53                      & 35.40                  & 9.84                   & \multicolumn{1}{r}{30.08} & 22.36                    & 0.46                      \\
w/ XLNet        & 35.60                  & 13.53                  & \multicolumn{1}{r}{31.39} & 18.79                    & 0.57                      & 34.47                  & 9.74                   & \multicolumn{1}{r}{29.29} & 21.94                    & 0.46                      \\ \midrule
GraphKvD        &                        &                        & \multicolumn{1}{r}{}      &                          &                           &                        &                        & \multicolumn{1}{r}{}      &                          &                           \\
w/ Lex. Overlap & 36.11                  & 12.97                  & \multicolumn{1}{r}{31.65} & 19.49                    & 0.49                      & 35.60                  & 10.12                  & \multicolumn{1}{r}{30.14} & 22.56                    & 0.38                      \\
w/ XLNet        & 35.75                  & 12.66                  & \multicolumn{1}{r}{31.34} & 19.02                    & 0.51                      & 34.77                  & 9.37                   & \multicolumn{1}{r}{29.32} & 22.55                    & 0.38                      \\ \midrule
Gold            & \multicolumn{1}{c}{-}  & \multicolumn{1}{c}{-}  & \multicolumn{1}{c}{-}     & 18.94                    & 0.92 & \multicolumn{1}{c}{-}  & \multicolumn{1}{c}{-}  & \multicolumn{1}{c}{-}     & 17.15                    & 0.89 \\ \bottomrule
\end{tabular}
\caption{Effect of using lexical overlap and semantic similarity in argument overlap calculation, as measured by ROUGE F$_1$ scores, redundancy (IUniq), and
local coherence (CCL), over the validation sets of \textsc{PubMed} and \textsc{arXiv}. }
\label{table:bertkvd}
\end{table}

\paragraph{Effect of Argument Overlap Strategy.}
Finally, we investigated the effect of employing more sophisticated strategies to calculate argument overlap in propositions.
We compared our proposed strategy --based in lexical overlap-- against a strategy using a pretrained Transformer-based encoder \shortcite{vaswani2017attention} to calculate semantic similarity.
We replace the Jaccard similarity between two arguments in Eq.~\refeq{eq:prop-overlap} by the maximum pairwise cosine similarity between wordpiece embeddings of said 
arguments.
Each sentence is encoded independently using XLNet \shortcite{yang2019xlnet} with the previous three sentences as context.
Recent work \cite{jeon2020centering,jeon2022entity} shown the advantage of using XLNet against other Transformer-based architectures 
when modeling local coherence in contexts a few sentences long.\footnote{Indeed, preliminary experiments using SciBERT \cite{beltagy2019scibert} shown poor results.}

Table~\ref{table:bertkvd} presents the results for \textsc{TreeKvD} and \textsc{GraphKvD}.
In both cases, we observe a reduction of relevancy and redundancy scores when using embedding-based similarity in argument overlap.
In \textsc{PubMed}, both KvD systems obtain higher cohesion scores with XLNet, whilst cohesion remains unchanged in \textsc{arXiv}.
These results indicate that employing semantic similarity in argument overlap hurts informativeness in greedily selected summaries, 
in line with similar findings by \citeA{fang2019proposition}.

We hypothesize that employing embedding-based similarity allows to connect arguments that are not semantically related but might be
close in embedding space, hence resulting in spurious proposition connections during attachment.
Naturally, with memory trees polluted with irrelevant propositions, KvD systems struggle to keep track on truly relevant information 
and informativeness will be impacted.

In conclusion, this section laid evidence as to how simulated cognitive processes impact the properties (informativeness, redundancy, and cohesion) of the final summary.
First, we pointed out the importance of constraining memory capacity in covering relevant content and dealing with redundant information.
Then, we highlighted the benefits of modeling working memory as a tree and how this affects the cohesion-redundancy trade-off.
We demonstrated the robustness of the proposed systems to parameters controlling recall from long-term memory.
Finally, the sensitivity of the systems to spurious connections between propositions was assessed 
and demonstrated that limiting connections through selective lexical overlap provides the best conditions for our systems to better balance informativeness, redundancy, and lexical cohesion in summaries.



\section{Conclusions}
In this paper, we studied the trade-off between redundancy and lexical cohesion in summaries produced by extractive systems,
and how this trade-off impacts informativeness.
We focused on the case when the input is a long document that exhibits information redundancy among the parts it is divided into.
As a case study, we experimented with scientific articles for which the main body --divided into sections-- is considered as the input document and the abstract is used as the reference summary.

Two optimization scenarios were investigated and compared, (i) when a summary property is optimized with a tailored reward in a reinforcement learning setup, and (ii) when a summary property is optimized through proxies inspired by a psycholinguistic model in an unsupervised setup.
In the first scenario, the trade-off between informativeness and cohesion was modeled as a linear combination
between a reward optimizing for ROUGE score w.r.t.\ the reference summary and a classifier-based reward optimizing for cohesion.
We found that models that optimize cohesion are capable of better organizing content in summaries compared to systems that optimize redundancy,
whilst maintaining --if not improving-- informativeness and coverage.

In the second scenario, we introduced two unsupervised summarization systems that implement explicit proxies that capture relevancy, non-redundancy, and lexical cohesion.
The proposed systems closely simulate how information is discretized into semantic propositions and organized
in human working memory, according to the Micro-Macro Structure theory of reading comprehension.
Extensive quantitative and qualitative analysis shown that our systems are able to extract summaries that are highly cohesive and as redundant as reference summaries,
however at the expense of sacrificing informativeness.
Finally, human evaluation campaigns revealed that KvD summaries exhibit a smooth topic transition between
sentences as signaled by proposition chains --an extension to lexical chains--, with chains spanning adjacent or near-adjacent sentences, and each sentence 
being connected to a previous one with at least one chain and to the next sentence with another chain.

\section*{Acknowledgements}

We thank the reviewers for their detailed feedback and Mausam and Julia Hockenmaier for handling the paper as associate editors. We also appreciate feedback from members of the Cohort, Ivan Titov, Arman Cohan, and Mark Steedman. We are grateful for a grant from NAVER Labs Europe, which funded this project. We appreciate the computing resources provided by the University of Birmingham and EPCC at the University of Edinburgh.


\vskip 0.2in
\bibliography{main}
\bibliographystyle{apacite}

\end{document}